\documentclass{article}

\usepackage[numbers,sort&compress]{natbib}
\usepackage[preprint]{neurips_2020}
\usepackage[breaklinks,colorlinks,
linkcolor=red,
anchorcolor=blue,
citecolor=blue
]{hyperref}
\usepackage[utf8]{inputenc} % allow utf-8 input
\usepackage[T1]{fontenc}    % use 8-bit T1 fonts
\usepackage{url}            % simple URL typesetting
\usepackage{booktabs}       % professional-quality tables
\usepackage{amsfonts}       % blackboard math symbols
\usepackage{nicefrac}       % compact symbols for 1/2, etc.
\usepackage{microtype}      % microtypography
\usepackage{graphicx}
\graphicspath{{figures/}}
\usepackage{longfbox}
\usepackage{amsmath}
\usepackage{amsfonts}
\usepackage{amssymb}
\usepackage{amsmath}   
\usepackage{mathtools}
\usepackage{amsthm}
\usepackage[algo2e,algoruled,boxed,lined,norelsize]{algorithm2e} 
\usepackage{subcaption}
\usepackage{wrapfig}
\usepackage{multirow}
\usepackage{makecell}
\usepackage{parnotes}
\usepackage{float}

\makeatletter
\newcommand{\mathleft}{\@fleqntrue\@mathmargin\parindent}
\newcommand{\mathcenter}{\@fleqnfalse\@mathmargin\parindent}
\newcommand{\ra}[1]{\renewcommand{\arraystretch}{#1}}
\makeatother
\DeclarePairedDelimiter{\norm}{\lVert}{\rVert}

\DeclareMathOperator{\Tr}{Tr}

\newtheorem{theorem}{Theorem}[section]

\newtheorem{lemma}[theorem]{Lemma}

\title{Pontryagin Differentiable Programming: \\An End-to-End Learning and Control Framework}

\author{
	\hspace{-4.0mm}Wanxin Jin\quad\qquad\qquad Zhaoran Wang\\
	Purdue University\quad\quad Northwestern University\\
	\texttt{\{wanxinjin,zhaoranwang\}@gmail.com}
	\And
	\hspace{-2.4mm}Zhuoran Yang \\
	\hspace{-2.4mm}Princeton University\\
	\hspace{-2.4mm}\texttt{zy6@princeton.edu}
	\And
	\hspace{-1.5mm}Shaoshuai Mou \\
	\hspace{-1.5mm}Purdue University\\
	\hspace{-2.5mm}\texttt{mous@purdue.edu} 
}

\begin{document}

\maketitle

\vspace{-2mm}
\begin{abstract}
	This paper develops a Pontryagin Differentiable Programming (PDP) methodology, which establishes a unified framework to solve a broad class of learning and control tasks. The PDP  distinguishes from existing methods by two novel techniques: first, we  differentiate through  Pontryagin's Maximum Principle, and  this allows  to obtain the analytical derivative of a  trajectory with respect to tunable parameters within an optimal control system,  enabling end-to-end learning of   dynamics, policies, or/and control objective functions; and second, we propose an auxiliary control system in the backward pass of the PDP framework, and  the output of this auxiliary control system is the  analytical derivative of the original system's trajectory with respect to the  parameters, which can be iteratively solved using standard control tools. We investigate three learning modes of the PDP: inverse reinforcement learning,  system identification, and  control/planning. We demonstrate the capability of the PDP in each learning mode on different high-dimensional systems, including multi-link robot arm,  6-DoF maneuvering quadrotor, and 6-DoF rocket powered landing.
\end{abstract}

\section{Introduction} \label{introduction}
Many learning tasks can find their counterpart problems in control fields. These tasks both seek to obtain  unknown aspects of a decision-making system with different terminologies  compared below.

\vspace{-2mm}

\begin{table*}[h]
	\caption{Topic connections between control and learning (details presented in Section \ref{background})}
	\centering
	\ra{1.1}
	\begin{tabular}{lll}\toprule
		 \small UNKNOWNS IN A SYSTEM & \small LEARNING METHODS & \small CONTROL METHODS\\ \midrule
		 Dynamics $\boldsymbol{x}_{t\text{+}1}{=}\boldsymbol{f}_{\boldsymbol{\theta}}(\boldsymbol{x}_{t},\boldsymbol{u}_{t})$ &Markov decision processes & System identification\\
		 Policy $\boldsymbol{u}_{t}=\boldsymbol{\pi}_{\boldsymbol{\theta}}(t,\boldsymbol{x}_t)$ &Reinforcement learning (RL) & Optimal control (OC)\\
	     Control objective $J{=}\sum_{t}{c_{\boldsymbol{\theta}}(\boldsymbol{x}_t,\boldsymbol{u}_t)}$ & Inverse RL & Inverse OC \\
	\bottomrule
	\end{tabular}\label{connectiontable} \vspace{-2mm}
\end{table*}
\begin{wrapfigure}[12]{r}{0pt}
	\raisebox{0pt}[\dimexpr\height-1.2\baselineskip]{\includegraphics[width=5.8cm, height=3.1cm]{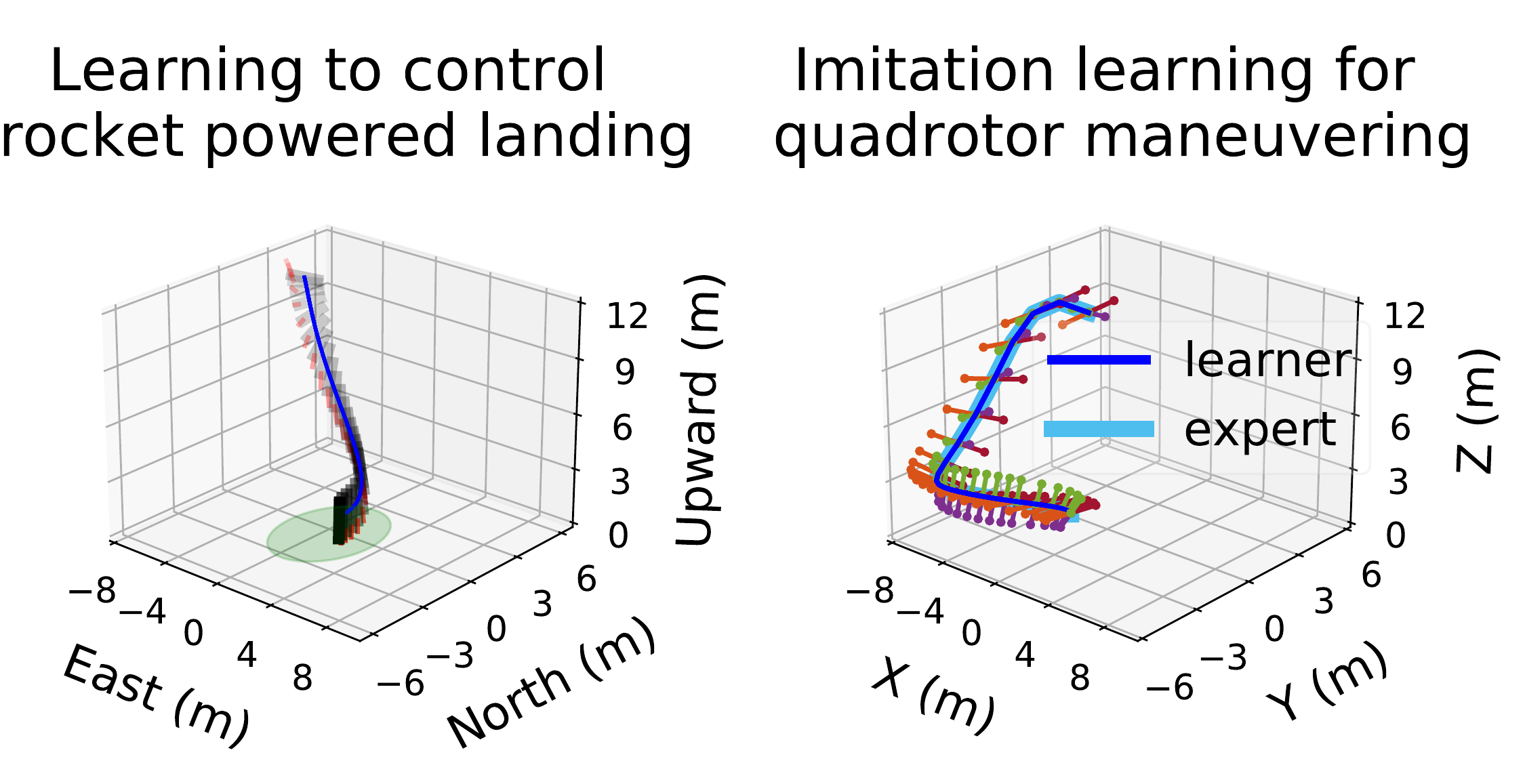}}
	\caption{left:  PDP  learns rocket landing control, right:  PDP  learns quadrotor dynamics and control objective for imitation.}
	\label{fig.intro}
\end{wrapfigure}
With the above connections, learning and control fields have  begun to explore the complementary benefits of each other: control theory may provide abundant models and structures that allow for efficient or certificated algorithms for  high-dimensional tasks, while learning  enables to obtain these models from data, which are otherwise  not readily attainable via classic control tools. Examples that enjoy  both benefits include model-based RL  \citep{gu2016continuous,heess2015learning}, where dynamics models are used for sample efficiency; and Koopman-operator control \citep{proctor2018generalizing,abraham2019active}, where via learning,  nonlinear systems are lifted to a linear observable space to facilitate control design. 
Inspired by those, this paper aims to exploit the advantage of integrating learning and control and develop a unified framework that enables to solve a wide range of learning and control tasks, e.g., the challenging  problems  in Fig.~\ref{fig.intro}.

\section{Background and Related Work}\label{background}

\vspace{-2mm}

\textbf{Learning dynamics.}  This is usually referred as to  as system identification in control fields, which typically consider linear systems represented by  transfer functions \cite{johansson1993system}. For nonlinear systems, the Koopman theory \cite{koopman1931hamiltonian} provides a way  to lift  states to a (infinite-dimensional) linear observable space \cite{williams2015data,proctor2018generalizing}.
In learning,  dynamics is characterized by  Markov decision proceses and implemented using linear regression \cite{haruno2001mosaic}, observation-transition modeling \cite{finn2016unsupervised}, latent-space modeling \cite{watter2015embed}, (deep) neural networks  \cite{fragkiadaki2015recurrent}, Gaussian process \cite{deisenroth2011pilco},  transition graphs \cite{zhang2018composable}, etc. Although  off-the-shelf, most of these methods have to trade off between data efficiency and long-term prediction accuracy. To achieve both, physically-informed learning \cite{raissi2019physics,saemundsson2020variational,lutter2019deep,zhong2019symplectic} injects physics laws into learning models, but they are limited to mechanical systems.   Recently, a trend of work  starts to use dynamical systems to explain (deep) neural networks, and some new algorithms \cite{chen2018neural, han2016deep, li2017maximum, li2018optimal,weinan2019mean, zhang2019you, benning2019deep, liu2019selection} have been established.

This paper focuses on learning general dynamical models, encompassing either physical dynamics with unknown parameters or neural difference equations.  The proposed  learning framework is injected with  inductive knowledge of optimal control theory to achieve  efficiency and explainability.

%Unlike the above work that  requires extensive  data or is limited to specific systems, the proposed PDP  learns general dynamics models, such as  by introducing an inductive bias of the Pontryagin's Maximum Principle of optimal control into the learning process.  The PDP solves for an exact \emph{derivative of a system trajectory with respect to dynamics}, which enables   end-to-end learning of   dynamics models in an explainable and efficient manner. 

\textbf{Learning optimal polices.}   In learning fields, it relates to reinforcement learning (RL).  Model-free RL provides a general-purpose framework to learn policies directly from interacting with environments \cite{oh2016control,mnih2013playing,mnih2015human},  but  usually suffers from significant data complexity. Model-based RL addresses this by first learning a dynamics model from experience  and then integrating it to
policy improvement \citep{schneider1997exploiting,gu2016continuous,abbeel2006using,deisenroth2011pilco,levine2014learning}. The use of a model can assist to augment experience data \cite{sutton1991dyna,tzeng2020adapting}, perform back-propagation through time \cite{deisenroth2011pilco}, or test  policies before deployment.  Model-based RL  also faces some challenges that are not well-addressed. For example, how to efficiently leverage  imperfect models \cite{janner2019trust}, and how to maximize the joint benefit by combining policy learning   and motion planning (trajectory optimization) \cite{levine2014learning,tamar2016value}, where a  policy has the advantage of  execution coherence and fast deployment while the trajectory planning has the competence of adaption to unseen or future situations.

The counterpart topic in control  is optimal control (OC), which is more concerned with characterizing  optimal trajectories in presence of  dynamics models. As in  RL, the main strategy for  OC is based on dynamic programming, and many valued-based methods are available, such as  HJB \cite{yong1999stochastic}, differential dynamical programming (DDP) \cite{jacobson1970differential} (by quadratizing  dynamics  and value function), and iterative linear quadratic regulator (iLQR)  \cite{li2004iterative} (by linearizing dynamics and quadratizing   value function).  The second strategy to solve OC is based on  the Pontryagin's Maximum/Minimal Principle (PMP)  \cite{pontryagin1962mathematical}. 
Derived from calculus of variations,  PMP can be thought of as    optimizing  directly over  trajectories, thus avoiding solving for  value functions.  Popular methods in this vein include  shooting methods \cite{bock1984multiple} and collocation methods \cite{patterson2014gpops}. However, the OC methods based on PMP are essentially \emph{open loop} control and thus susceptible to model errors or  disturbances in deployment. To address these,  model predictive control (MPC) \cite{camacho2013model} generates  controls given the system current  state by repeatedly solving an OC problem over a finite prediction horizon (only the first optimal input is executed),  leading to a \emph{closed-loop} control form. Although MPC has dominated across many industrial applications \cite{qin2000overview}, developing fast MPC implementations is still an active research direction \cite{wang2009fast}.

The proposed learning framework in this work  has a special mode for  model-based control tasks. The method  can be viewed as a complement   to classic open-loop OC methods, because, although derived from PMP (trajectory optimization), the method here is to learn a \emph{feedback/closed-loop} control policy. Depending on the specific policy parameterization, the method here can also be  used for  motion planning. All these  features will provide  new perspectives for  model-based RL or MPC control.

\textbf{Learning control objective functions.} In learning, this relates to  inverse reinforcement learning (IRL), whose goal is to find a control objective function to explain the given optimal demonstrations. The unknown  objective function is typically parameterized as a weighted sum of features \cite{abbeel2004apprenticeship,ziebart2008maximum,ratliff2006maximum}. Strategies to learn the unknown weights include feature matching \cite{abbeel2004apprenticeship} (matching the feature values between demonstrations and reproduced trajectories), maximum entropy \cite{ziebart2008maximum} (finding a trajectory distribution of maximum entropy subject to empirical feature values), and maximum margin \cite{ratliff2006maximum} (maximizing the margin of  objective values between demonstrations and reproduced trajectories). The learning update in the above IRL methods  is  preformed on a selected feature space by taking advantage of  linearity of feature weights, and thus cannot be directly applied to learning  objective functions that are nonlinear in parameters. The counterpart topic in the control field is inverse optimal control (IOC) \cite{keshavarz2011imputing,mombaur2010human,jin2018inverse,jin2019inverse}. With  knowledge of dynamics, IOC focuses on more efficient learning paradigms. For example, by directly minimizing the  violation of  optimality conditions by the observed  demonstration data, \cite{keshavarz2011imputing,englert2017inverse,jin2019inverse,jin2018inverse} directly compute  feature weights without repetitively solving  the OC problems. Despite the efficiency, minimizing optimality violation  does not directly assure the closeness between the  final  reproduced trajectory and  demonstrations or the closeness of their  objective values.

Fundamentally different from existing IRL/IOC methods, this paper will develop a new framework that enables to learn complex control objective functions, e.g., neural  objective functions, by directly minimizing the loss (e.g., the distance) between the reproduced trajectory and demonstrations.

\textbf{A unified perspective on learning dynamics/policy/control objective functions.} Consider a general decision-making system, which typically consists of   aspects of dynamics, control policy, and control objective function. In a unified perspective, learning  dynamics,    policies, or    control objective functions can be viewed as  \emph{instantiations of  the same learning problem} but with (i)  unknown parameters appearing in the system's different aspects and (ii) the different  losses. For example, in learning dynamics, a differential/difference equation is parameterized and the loss function can be defined as the prediction error between the equation's output and  target data; in learning  policies, the unknown parameters are in a feedback policy and the loss function  is just the  control objective function;  and in learning control objective functions, the control objective function is parameterized and the loss function can be the discrepancy between the  reproduced trajectory and  the observed  demonstrations.

\textbf{Claim of contributions.} Motivated by the above, this paper develops a unified learning framework, named as PDP,  that is flexible enough to be customized for different learning and control tasks and capable enough to efficiently solve  high-dimensional and continuous-space  problems. The proposed PDP  framework borrows the idea of `end-to-end' learning \cite{muller2006off} and  chooses to optimize a loss function directly with respect to the tunable parameters in the aspect(s) of a decision-making system, such as the dynamics, policy, or/and control objective function. The key contribution of the PDP is that we  inject the optimal control theory as an inductive bias into the learning process  to expedite  the learning efficiency and explainability. Specifically, the  PDP  framework centers around the  system's  trajectory  and \emph{differentiates through  PMP}, and this allows us to obtain the analytical derivative of the trajectory with respect to the tunable parameters, a  key quantity for  end-to-end learning of (neural) dynamics, (neural) policies, and  (neural) control objective functions. Furthermore, we introduce an \emph{auxiliary control system} in the back pass of the PDP framework, and its output trajectory is  exactly the  derivative of the  trajectory with respect to the  parameters, which can be iteratively solved using standard control tools. In control fields, to our best knowledge, this is  the first work to propose the technique of the \emph{differential PMP},   and more importantly, we show that the \emph{differential PMP} can be easily obtained using the introduced auxiliary control system.

\section{Problem Formulation}\label{problemformulation}
We begin with formulating a base problem and then discuss how to accommodate the base problem to specific applications. Consider  a class of optimal control systems $\boldsymbol{\Sigma}(\boldsymbol{\theta})$, which is parameterized by a tunable $\boldsymbol{\theta}\in\mathbb{R}^r$  in both  dynamics and control (cost) objective function:

{
\begin{longfbox}[padding-top=-2pt,margin-top=-5pt, padding-bottom=1pt, margin-bottom=3pt]
	\mathleft
	\begin{equation}\label{oc}
	\boldsymbol{\Sigma}(\boldsymbol{\theta}):\qquad\quad
	\begin{aligned}
	\text{dynamics:} &\qquad \boldsymbol{x}_{t{+}1}=\boldsymbol{f}(\boldsymbol{x}_{t},\boldsymbol{u}_{t}, {\boldsymbol{\theta}}) \quad  \text{with}  \,\, \text{given}\,\,\boldsymbol{x}_0, \\
	\text{control objective:} &\qquad J(\boldsymbol{\theta})=\sum\nolimits_{t=0}^{T{-}1}c_t(\boldsymbol{x}_t,\boldsymbol{u}_t, {\boldsymbol{\theta}})+h(\boldsymbol{x}_T,{\boldsymbol{\theta}}).
	\end{aligned}
	\end{equation}	\mathcenter
\end{longfbox}
}
Here, $\boldsymbol{x}_t\in\mathbb{R}^n$ is the system state; $\boldsymbol{u}_t\in\mathbb{R}^m$ is the control  input;  $\boldsymbol{f}:\mathbb{R}^n\times\mathbb{R}^m\times\mathbb{R}^r\mapsto\mathbb{R}^n$ is the dynamics model, which is assumed to be twice-differentiable;  $t=0,1,\cdots, T$ is the time step with $T$ being the time  horizon; and $J(\boldsymbol{\theta})$ is the control objective function  with $c_t:\mathbb{R}^n\times\mathbb{R}^m\times\mathbb{R}^r\mapsto\mathbb{R}$ and $h:\mathbb{R}^n\times\mathbb{R}^r\mapsto\mathbb{R}$ denoting the stage/running  and final costs, respectively,  both of which are  twice-differentiable. For a choice of $\boldsymbol{\theta}$,  $\boldsymbol{\Sigma}(\boldsymbol{\theta})$ will produce a trajectory of state-inputs:
\begin{equation}\label{octraj}
\begin{aligned}
 \boldsymbol{\xi}_{\boldsymbol{\theta}}{=}\{\boldsymbol{{x}}_{0:T}^{\boldsymbol{\theta}},\boldsymbol{{u}}_{0:T{-}1}^{\boldsymbol{\theta}}\}\,\,\,\in\,\,\,\arg\quad&\min\nolimits_{\{\boldsymbol{{x}}_{0:T},\boldsymbol{{u}}_{0:T\text{-}1}\}} \,\, J(\boldsymbol{\theta})
 \\ &\,\,\text{subject to}   \qquad  \boldsymbol{x}_{t{+}1}{=}\boldsymbol{f}(\boldsymbol{x}_{t},\boldsymbol{u}_{t}, {\boldsymbol{\theta}}) \,\, \text{for all $t$  given $\boldsymbol{x}_0$}\,\,\,\,
\end{aligned},
\end{equation}
that is, $\boldsymbol{\xi}_{\boldsymbol{\theta}}$ optimizes   $J(\boldsymbol{\theta})$  subject to the dynamics constraint $\boldsymbol{f}(\boldsymbol{\theta})$. For many applications (we will show next), one evaluates the above $\boldsymbol{\xi}_{\boldsymbol{\theta}}$ using a scalar-valued differentiable loss $L(\boldsymbol{\xi}_{\boldsymbol{\theta}},\boldsymbol{\theta})$. Then, the \textbf{problem of interest} is to tune the  parameter $\boldsymbol{\theta}$, such that  $\boldsymbol{\xi}_{\boldsymbol{\theta}}$ has the minimal loss:
\begin{equation}\label{prob}
\min_{\boldsymbol{\theta}}L(\boldsymbol{{\xi}}_{{\boldsymbol{\theta}}},\boldsymbol{\theta})\quad  {\text{subject to}} \quad  \boldsymbol{{\xi}}_{{\boldsymbol{\theta}}} \text{ is in } (\ref{octraj}).
\end{equation}
Under the above base formulation, for a specific learning or control task, one only needs to accordingly  change  precise details of $\boldsymbol{\Sigma}(\boldsymbol{\theta})$ and define a specific loss function $L(\boldsymbol{\xi}_{\boldsymbol{\theta}},\boldsymbol{\theta})$, as we discuss below.

{\textbf{IRL/IOC Mode}.} \quad Suppose that we are given optimal   demonstrations $\boldsymbol{\xi}^{\text{d}}=\{\boldsymbol{{x}}_{0:T}^\text{d},\boldsymbol{{u}}^\text{d}_{0:T-1}\}$ of an  expert optimal control system. We seek to  learn the expert's dynamics and control objective function from $\boldsymbol{\xi}^\text{d}$. To this end, we use $\boldsymbol{\Sigma}(\boldsymbol{\theta})$ in (\ref{oc}) to represent the expert, and define the loss in (\ref{problemformulation})  as
\begin{equation}\label{lossioc}
L(\boldsymbol{\xi}_{\boldsymbol{\theta}},\boldsymbol{\theta})=l(\boldsymbol{\xi}_{\boldsymbol{\theta}}, \boldsymbol{\xi}^\text{d}), 
\end{equation}
where $l$ is a scalar function that penalizes the inconsistency of $\boldsymbol{\xi}_{\boldsymbol{\theta}}$  with $\boldsymbol{\xi}^\text{d}$, e.g.,  $l(\boldsymbol{\xi}_{\boldsymbol{\theta}}, \boldsymbol{\xi}^\text{d})=\norm{\boldsymbol{\xi}_{\boldsymbol{\theta}}-\boldsymbol{\xi}^\text{d}}^2$. By solving (\ref{prob}) with (\ref{lossioc}), we can obtain a  $\boldsymbol{\Sigma}(\boldsymbol{{\theta}}^*)$ whose trajectory is consistent with the observed demonstrations. It should be noted that  even if the demonstrations $\small\boldsymbol{\xi}^\text{d}$ significantly deviate from the optimal ones, the above formulation still finds the `best' control objective function (and dynamics) within the parameterized  set $\boldsymbol{\Sigma}(\boldsymbol{{\theta}})$ such that its reproduced $\boldsymbol{\xi}_{\boldsymbol{\theta}}$ in (\ref{octraj}) has the \emph{minimal distance} to $\boldsymbol{\xi}^\text{d}$.

{\textbf{SysID Mode}.} \quad  Suppose that we are given data $\boldsymbol{\xi}^\text{o}=\{\boldsymbol{{x}}_{0:T}^\text{o}, \boldsymbol{u}_{0:T-1}\}$ collected from, say, a physical system (here, unlike $\small\boldsymbol{\xi}^\text{d}$, $\small\boldsymbol{\xi}^\text{o}$ is not necessarily optimal), and we wish to identify the system's dynamics.  Here,  $\boldsymbol{u}_{0:T-1}$ are usually   externally supplied to ensure the physical system is of persistent excitation \cite{green1986persistence}. In order for
  $\boldsymbol{\Sigma}(\boldsymbol{\theta})$ in (\ref{oc})  to only represent  dynamics (as we do not care about its internal control law), we set  $J(\boldsymbol{{\theta}})=0$. Then, $\boldsymbol{\xi}_{\boldsymbol{\theta}}$ in (\ref{octraj}) accepts any  $\boldsymbol{{u}}_{0:T-1}^{\boldsymbol{\theta}}=\boldsymbol{{u}}_{0:T-1}$ as it always optimizes $J(\boldsymbol{{\theta}}){=}0$. In other words, by setting $J(\boldsymbol{\theta})=0$,  $\boldsymbol{\Sigma}(\boldsymbol{\theta})$ in (\ref{oc})  now only represent a class of  dynamics models:
  
\begin{longfbox}[padding-top=-4pt,margin-top=-6pt, padding-bottom=-1pt, margin-bottom=3pt]
	\mathleft
	\begin{equation}\label{ocmodeid}
	\boldsymbol{\Sigma}(\boldsymbol{\theta}):\qquad\qquad
	\text{dynamics:} \quad
	\boldsymbol{x}_{t+1}=\boldsymbol{f}(\boldsymbol{x}_{t},\boldsymbol{u}_{t}, {\boldsymbol{\theta}}) \qquad  \text{with }  \,\boldsymbol{x}_0 \,\ \text{and \,} \,\boldsymbol{{u}}_{0:T-1}^{\boldsymbol{\theta}}=\boldsymbol{{u}}_{0:T-1}.
	\end{equation}
	\mathcenter
\end{longfbox}
Now, $\boldsymbol{\Sigma}(\boldsymbol{\theta})$   produces  $\boldsymbol{\xi}_{\boldsymbol{\theta}}=\{\boldsymbol{x}^{\boldsymbol{\theta}}_{0:T},\boldsymbol{u}^{\boldsymbol{\theta}}_{0:T-1}\}$ subject to (\ref{ocmodeid}).  To use (\ref{prob}) for  identifying $\boldsymbol{\theta}$, we define
\begin{equation}\label{lossid}
L(\boldsymbol{\xi}_{\boldsymbol{\theta}},\boldsymbol{\theta})=l(\boldsymbol{\xi}_{\boldsymbol{\theta}}, \boldsymbol{\xi}^\text{o}), 
\end{equation}
where $l$ is to quantify the prediction error between  $\boldsymbol{\xi}^{\text{o}}$  and $\boldsymbol{\xi}_{\boldsymbol{\theta}}$ under the same  inputs $\boldsymbol{u}_{0:T-1}$.

\vspace{-1mm}
\subsubsubsection{\textbf{Control/Planning Mode.}} \quad Consider a  system with its dynamics  learned in the above {SysID}. We  want to obtain a  \emph{feedback controller} or \emph{trajectory} such that the system achieves  a  performance of minimizing a given cost function. To that end, we specialize $\boldsymbol{\Sigma}(\boldsymbol{\theta})$ in (\ref{oc}) as follows: first,  set $\boldsymbol{f}$ as the learned dynamics and  $J(\boldsymbol{\theta})=0$; and second, through a \emph{close-loop link}, we connect the input $\boldsymbol{u}_t$ and  state  $\boldsymbol{x}_t$  via a parameterized  policy block $\boldsymbol{u}_t=\boldsymbol{u}(t,\boldsymbol{x}_t,\boldsymbol{\theta})$ (reminder: unlike SysID Mode with $\boldsymbol{u}_t$ supplied externally, the inputs here are from a  policy via a feedback loop).  $\boldsymbol{\Sigma}(\boldsymbol{\theta})$ now becomes

{
\begin{longfbox}[padding-top=-4pt,margin-top=-5pt,padding-bottom=-2pt,margin-bottom=-1pt,]
	\mathleft
	\begin{equation}\label{ocmodeplan}
	\boldsymbol{\Sigma}(\boldsymbol{\theta}):\qquad\qquad\qquad\qquad
	\begin{aligned}
	\text{dynamics:} &\quad \boldsymbol{x}_{t+1}=\boldsymbol{f}(\boldsymbol{x}_{t},\boldsymbol{u}_{t}) \quad  \text{with}  \quad\,\,\boldsymbol{x}_0, \\
	\text{control policy:} &\quad \boldsymbol{u}_t=\boldsymbol{u}(t,\boldsymbol{x}_t,\boldsymbol{\theta}).
	\end{aligned}
	\end{equation}
	\mathcenter
\end{longfbox}
}\normalsize
Now, $\boldsymbol{\Sigma}(\boldsymbol{\theta})$  produces a trajectory  $\boldsymbol{\xi}_{\boldsymbol{\theta}}=\{\boldsymbol{x}^{\boldsymbol{\theta}}_{0:T},\boldsymbol{u}_{0:T-1}^{\boldsymbol{\theta}}\}$ subject to (\ref{ocmodeplan}). We set  the loss  in (\ref{prob}) as 
\begin{equation}\label{lossmodeplan}
L(\boldsymbol{\xi}_{\boldsymbol{\theta}},\boldsymbol{\theta})=\sum\nolimits_{t=0}^{T-1}l(\boldsymbol{x}^{\boldsymbol{\theta}}_t,\boldsymbol{u}^{\boldsymbol{\theta}}_t) +l_f(\boldsymbol{x}^{\boldsymbol{\theta}}_{T}),
\end{equation}
where $l$ and $l_f$ are the stage and final  costs, respectively. Then, (\ref{prob})  is  an optimal control or planning problem: if  $\boldsymbol{u}_t{=}\boldsymbol{u}(t,\boldsymbol{x}_t,\boldsymbol{\theta})$ (i.e., feedback policy explicitly depends on  $\boldsymbol{x}_t$), (\ref{prob})  is a \emph{close-loop optimal control} problem; otherwise if    $\boldsymbol{u}_t{=}\boldsymbol{u}(t,\boldsymbol{\theta})$ (e.g., polynomial parameterization), (\ref{prob})  is an \emph{open-loop motion planning} problem.  This  mode  can also be used as a component to solve (\ref{oc}) in IRL/IOC Mode.

\section{An End-to-End Learning Framework} \label{scheme}
\begin{figure}[h]
	\centering
	\includegraphics[width=0.66\columnwidth]{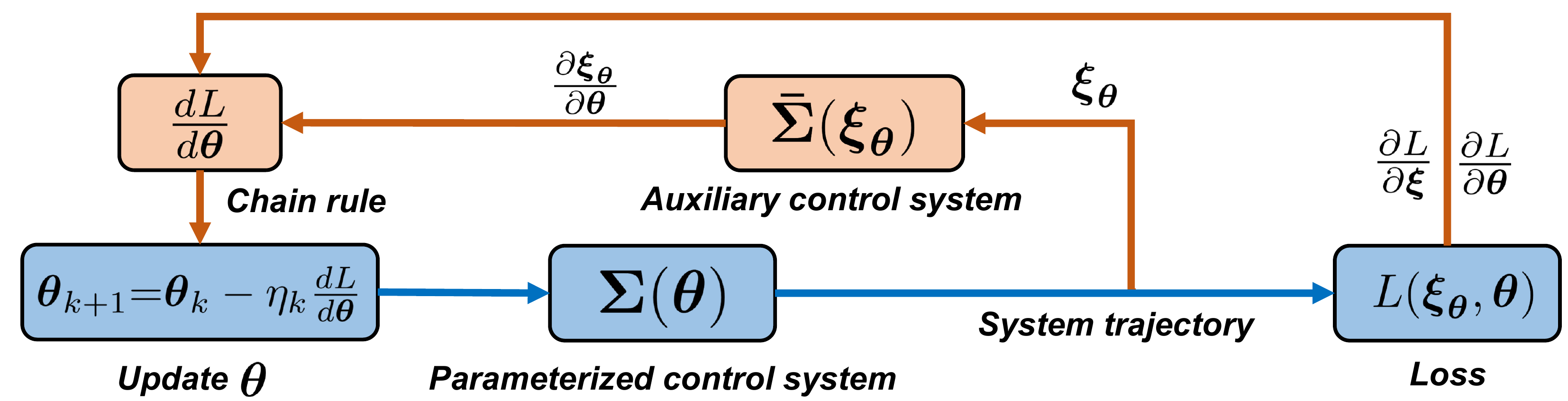}
	\caption{PDP end-to-end learning framework.}
	\label{overview}
\end{figure}

To solve the generic problem in (\ref{prob}), the  idea of end-to-end learning \cite{muller2006off}  seeks to optimize the loss $L(\boldsymbol{{\xi}}_{{\boldsymbol{\theta}}},\boldsymbol{\theta})$ \emph{directly} with respect to the tunable parameter $\boldsymbol{\theta}$, by  applying
the gradient descent
\begin{small}
	\begin{equation}\label{GD}
	\boldsymbol{\theta}_{k+1}=\boldsymbol{\theta}_{k}-\eta_{k}\frac{d L}{d \boldsymbol{\theta}}\Bigr|_{\boldsymbol{\theta}_k} \quad \text{with}\quad \frac{d L}{d \boldsymbol{\theta}}\Bigr|_{\boldsymbol{\theta}_k}=\frac{\partial L}{\partial \boldsymbol{{\xi}}}\Bigr|_{\boldsymbol{{\xi}}_{\boldsymbol{\theta}_k}}\frac{\partial  \boldsymbol{{\xi}}_{\boldsymbol{\theta}}}{\partial \boldsymbol{{\theta}}}\Bigr|_{\boldsymbol{\theta}_k}+\frac{\partial L}{\partial \boldsymbol{{\theta}}}\Bigr|_{\boldsymbol{\theta}_k}.
	\end{equation}
\end{small}%
Here, $k=0,1,\cdots$ is the iteration index;  $\frac{d L}{d \boldsymbol{\theta}}\bigr|_{\boldsymbol{\theta}_k}$ is the gradient of the loss with respect to  $\boldsymbol{\theta}$ evaluated at $\boldsymbol{{\theta}}_{k}$; and $\eta_{k}$ is the learning rate.
From  (\ref{GD}), we can draw a learning architecture in Fig. \ref{overview}. Each update of $\boldsymbol{\theta}$ consists of a \emph{forward pass}, where at $\boldsymbol{{\theta}}_{k}$,  the corresponding trajectory  $\boldsymbol{{\xi}}_{\boldsymbol{\theta}_k}$ is solved from $\boldsymbol{\Sigma}(\boldsymbol{\theta}_k)$ and the loss is computed, and a \emph{backward pass}, where $\frac{\partial L}{\partial \boldsymbol{{\xi}}}\bigr|_{\boldsymbol{{\xi}}_{\boldsymbol{\theta}_k}}$,  $\frac{\partial  \boldsymbol{{\xi}}_{\boldsymbol{\theta}}}{\partial \boldsymbol{{\theta}}}\bigr|_{\boldsymbol{\theta}_k}$, and $\frac{\partial  L}{\partial \boldsymbol{{\theta}}}\bigr|_{\boldsymbol{\theta}_k}$ are computed.

In the forward pass,   $\boldsymbol{{\xi}}_{\boldsymbol{\theta}}$ is obtained by solving an optimal control problem in  $\boldsymbol{\Sigma}(\boldsymbol{\theta})$ using any available OC methods, such as iLQR or Control/Planning Mode, (note that in SysID or Control/Planning modes, it is reduced to  integrating  difference equations (\ref{ocmodeid}) or (\ref{ocmodeplan})). In  backward pass, $\frac{\partial L}{\partial \boldsymbol{{\xi}}}$ and $\frac{\partial  L}{\partial \boldsymbol{{\theta}}}$ are  easily obtained  from the  loss function  $L(\boldsymbol{{\xi}}_{\boldsymbol{\theta}},\boldsymbol{{\theta}})$. The main challenge, however, is to solve $\frac{\partial  \boldsymbol{{\xi}}_{\boldsymbol{{\theta}}}}{\partial \boldsymbol{{\theta}}}$, i.e., \emph{the derivative of a trajectory with respect to the parameters in the system}. Next, we will analytically solve  $\frac{\partial  \boldsymbol{{\xi}}_{\boldsymbol{{\theta}}}}{\partial \boldsymbol{{\theta}}}$ by proposing two techniques:  \emph{differential PMP} and  \emph{auxiliary control system}. 

\vspace*{-2mm}

\section{Key Contributions: Differential PMP \& Auxiliary Control System} \label{DPMPAUX}

We first recall the discrete-time Pontryagin's Maximum/Minimum Principle (PMP) \citep{pontryagin1962mathematical} (a derivation of  discrete-time PMP is given in Appendix \ref{appendix3}). For the optimal control system $\boldsymbol{\Sigma}(\boldsymbol{\theta})$ in (\ref{oc}) with a fixed $\boldsymbol{\theta}$,  PMP describes a set of optimality conditions which the trajectory $\boldsymbol{\xi}_{\boldsymbol{\theta}}=\{\boldsymbol{{x}}_{0:T}^{\boldsymbol{\theta}},\boldsymbol{{u}}_{0:T-1}^{\boldsymbol{\theta}}\}$ in (\ref{octraj}) must satisfy. To introduce these conditions, we first define the following  \emph{Hamiltonian},
\begin{equation}\label{Hamil}
H_t=c_t(\boldsymbol{x}_t,\boldsymbol{u}_t;\boldsymbol{\theta})+\boldsymbol f(\boldsymbol{x}_t,\boldsymbol{u}_t;\boldsymbol{\theta})^\prime\boldsymbol{\lambda}_{t\text{+}1},
\end{equation}
where $\boldsymbol{\lambda}_{t}\in \mathbb{R}^{n}$ ($t=1,2, \cdots, T$) is called the \emph{costate variable}, which can be also thought of as the Lagrange multipliers for the  dynamics constraints.
According to  PMP,  there exists a sequence of costates $\boldsymbol{{\lambda}}^{\boldsymbol{\theta}}_{1:T}$, which together with the  optimal trajectory  $\boldsymbol{{\xi}}_{\boldsymbol{{\theta}}}=\{\boldsymbol{{x}}^{\boldsymbol{\theta}}_{0:T},\boldsymbol{{u}}^{\boldsymbol{\theta}}_{0:T-1}\}$ satisfy
\begin{subequations}\label{mp}
		\begin{align}
		\text{\normalsize dynamics equation:}\quad\quad\,\, \small\boldsymbol{{x}}^{\boldsymbol{\theta}}_{t\text{+}1}&=\small\dfrac{\partial\small H_t}{\partial \boldsymbol{{\lambda}}^{\boldsymbol{\theta}}_{t+1}}=\boldsymbol{f}(\boldsymbol{{x}}^{\boldsymbol{\theta}}_t,\boldsymbol{{u}}^{\boldsymbol{\theta}}_t;\boldsymbol{\theta}), \label{mp.1} \hspace{2.0cm}\\[-2pt]
		\text{\normalsize costate equation:}\quad\quad \quad \small\boldsymbol{\lambda}^{\boldsymbol{\theta}}_{t}&=\small\dfrac{\partial H_t}{\partial \boldsymbol{{x}}^{\boldsymbol{\theta}}_{t}}=\dfrac{\partial c_t}{\partial \boldsymbol{{x}}^{\boldsymbol{\theta}}_{t}}+\dfrac{\partial \boldsymbol{f}^\prime}{\partial \boldsymbol{{x}}^{\boldsymbol{\theta}}_{t}}\boldsymbol{{\lambda}}^{\boldsymbol{\theta}}_{t\text{+}1},\label{mp.2}\hspace{2.0cm}\\[-2pt]
		\text{\normalsize input equation:} \quad\quad\quad\,\,\,\,
		\small\boldsymbol{0}&=\small\dfrac{\partial H_t}{\partial \boldsymbol{{u}}^{\boldsymbol{\theta}}_{t}}=\dfrac{\partial c_t}{\partial \boldsymbol{{u}}^{\boldsymbol{\theta}}_{t}}+\dfrac{\partial \boldsymbol{f}^\prime}{\partial \boldsymbol{{u}}^{\boldsymbol{\theta}}_{t}}\boldsymbol{{\lambda}}^{\boldsymbol{\theta}}_{t\text{+}1},\label{mp.3}
		\\[-2pt]
		\text{\normalsize boundary conditions:}\quad\quad\quad \small\boldsymbol{{\lambda}}^{\boldsymbol{\theta}}_{T}&=\small\frac{\partial h}{\partial\boldsymbol{{x}}^{\boldsymbol{\theta}}_T},  \quad\quad \boldsymbol{x}_0^{\boldsymbol{\theta}}=\boldsymbol{x}_0 \label{mp.4}. \hspace{2.6cm}
		\end{align}
\end{subequations}%
For notation simplicity,  $\frac{\partial\boldsymbol{g}}{\partial \boldsymbol{x}_t}$  means the derivative of  function $\boldsymbol{g}(\boldsymbol{x})$ with respect to $\boldsymbol{x}$  evaluated at $\boldsymbol{x}_t$.

\subsection{Differential PMP}
To begin, recall that our goal (in Section \ref{scheme}) is to obtain $\frac{\partial  \boldsymbol{{\xi}}_{\boldsymbol{{\theta}}}}{\partial \boldsymbol{{\theta}}}$, that is,
\begin{equation}\label{unknown}
		{\frac{\partial \boldsymbol{{\xi}}_{\boldsymbol{\theta}}}{\partial \boldsymbol{\theta}}=\small\left\{\frac{\partial \boldsymbol{{x}}^{\boldsymbol{\theta}}_{0:T}}{\partial \boldsymbol{\theta}},\frac{\partial \boldsymbol{{u}}^{\boldsymbol{\theta}}_{0:T-1}}{\partial \boldsymbol{\theta}}\right\} .}
\end{equation}
To this end,  we are motivated to differentiate the PMP conditions in (\ref{mp}) on both sides  with respect to  $\boldsymbol{\theta}$. This leads to the following \emph{differential PMP}:
\begin{subequations}\label{diffpontryagin}
	\begin{align}
	\text{differential dynamics equation:}\quad
	\small\dfrac{\partial \boldsymbol{{x}}^{\boldsymbol{\theta}}_{t\text{+}1}}{\partial \boldsymbol{\theta}}&=\small
	F_t\dfrac{\partial \boldsymbol{{x}}^{\boldsymbol{\theta}}_{t}}{\partial \boldsymbol{\theta}}+G_t\dfrac{\partial \boldsymbol{{u}}^{\boldsymbol{\theta}}_{t}}{\partial \boldsymbol{\theta}}+E_t,\label{diffpontryagin.1} \hspace{0.5cm}
	\\[-1pt]
	\text{differential costate equation:}\quad\,\,\,\,
	\small\dfrac{\partial \boldsymbol{{\lambda}}^{\boldsymbol{\theta}}_{t}}{\partial \boldsymbol{\theta}}&
	=\small
	H_t^{xx}\dfrac{\partial \boldsymbol{{x}}^{\boldsymbol{\theta}}_{t}}{\partial \boldsymbol{\theta}}+ H_t^{xu}\dfrac{\partial \boldsymbol{{u}}^{\boldsymbol{\theta}}_{t}}{\partial \boldsymbol{\theta}}+F_t^\prime\dfrac{\partial \boldsymbol{{\lambda}}^{\boldsymbol{\theta}}_{t+1}}{\partial \boldsymbol{\theta}}+H_t^{xe},\label{diffpontryagin.2} \hspace{0.5cm}
	\\[-1pt]
	\text{differential input equation:}\quad\quad\quad
	\small\boldsymbol{0}&=
	\small H_t^{ux}\dfrac{\partial \boldsymbol{{x}}^{\boldsymbol{\theta}}_{t}}{\partial \boldsymbol{\theta}}+H_t^{uu}\dfrac{\partial \boldsymbol{{u}}^{\boldsymbol{\theta}}_{t}}{\partial \boldsymbol{\theta}}+G_t^\prime\dfrac{\partial \boldsymbol{{\lambda}}^{\boldsymbol{\theta}}_{t+1}}{\partial \boldsymbol{\theta}}+H_t^{ue} \label{diffpontryagin.3},\hspace{0.5cm}
	\\[-1pt]
	\text{differential boundary conditions:}\quad\,\,\,
	\small\dfrac{\partial \boldsymbol{{\lambda}}^{\boldsymbol{\theta}}_{T}}{\partial \boldsymbol{\theta}}&=\small H_T^{xx}\dfrac{\partial \boldsymbol{{x}}^{\boldsymbol{\theta}}_T}{\partial\boldsymbol{\theta}}+H_T^{xe},\quad\quad \small\frac{\partial\boldsymbol{x}_0^{\boldsymbol{\theta}}}{\partial \boldsymbol{\theta}}\small=\frac{\partial\boldsymbol{x}_0}{\partial \boldsymbol{\theta}}=\small\boldsymbol{0} \label{diffpontryagin.4}.
	\end{align}
\end{subequations}
Here, to simplify notations and distinguish knowns and unknowns, the coefficient matrices  in the above differential PMP (\ref{diffpontryagin}) are defined as follows:
\begin{small}
	\begin{subequations}\label{mats}
		\begin{align}
		\small F_t& {=}\small \dfrac{\partial \boldsymbol{f}}{\partial \boldsymbol{{x}}^{\boldsymbol{\theta}}_t}, 
		&   G_t&{=} \dfrac{\partial \boldsymbol{f}}{\partial \boldsymbol{{u}}^{\boldsymbol{\theta}}_t}, 
		&   H_t^{xx}& {=}\dfrac{\partial^2 H_t}{\partial \boldsymbol{{x}}^{\boldsymbol{\theta}}_{t}\partial \boldsymbol{{x}}^{\boldsymbol{\theta}}_{t}}, 
		&   H_t^{xe}& {=}\dfrac{\partial^2 H_t}{\partial \boldsymbol{{x}}^{\boldsymbol{\theta}}_{t}\partial \boldsymbol{\theta}}, 
		&   H_t^{xu}& {=}\dfrac{\partial^2 H_t}{\partial \boldsymbol{{x}}^{\boldsymbol{\theta}}_{t}\partial \boldsymbol{{u}}^{\boldsymbol{\theta}}_{t}}{=}{(H_t^{ux})}^\prime, \label{mats.1}
		\\[-1pt]
		E_t&{=}  \dfrac{\partial \boldsymbol{f}}{\partial \boldsymbol{\theta}}, 
		&    H_t^{uu}& {=}\dfrac{\partial^2 H_t}{\partial \boldsymbol{{u}}^{\boldsymbol{\theta}}_{t}\partial \boldsymbol{{u}}^{\boldsymbol{\theta}}_{t}}, 
		&    H_t^{ue}&  {=}\dfrac{\partial^2 H_t}{\partial \boldsymbol{{u}}^{\boldsymbol{\theta}}_{t}\partial \boldsymbol{\theta}},
		&   H_T^{xx}& {=}\dfrac{\partial^2 h}{\partial \boldsymbol{{x}}^{\boldsymbol{\theta}}_{T}\partial \boldsymbol{{x}}^{\boldsymbol{\theta}}_{T}},
		& H_T^{xe}& {=}\small\dfrac{\partial^2 h}{\partial \boldsymbol{{x}}^{\boldsymbol{\theta}}_{T}\partial \boldsymbol{\theta}}
		\label{mats.2},
		\end{align}
	\end{subequations}
\end{small}%
where we use  $\frac{\partial^2 \boldsymbol{g}}{\partial \boldsymbol{{x}}_{t}\partial \boldsymbol{{u}}_t}$ to denote the second-order derivative of a function $\boldsymbol{g}(\boldsymbol{x},\boldsymbol{u})$  evaluated at ($\boldsymbol{x}_t, \boldsymbol{u}_t$). Since the  trajectory $\boldsymbol{{\xi}}_{\boldsymbol{\theta}}=\{\boldsymbol{{x}}^{\boldsymbol{\theta}}_{0:T},\boldsymbol{{u}}^{\boldsymbol{\theta}}_{0:T-1}\}$ is obtained in the forward pass (recall Fig.~\ref{overview}),  all   matrices in (\ref{mats})  are thus known (note that the computation of these matrices also requires $\boldsymbol{{\lambda}}^{\boldsymbol{\theta}}_{1:T}$, which can be obtained by iteratively solving (\ref{mp.2}) and (\ref{mp.4}) given $\boldsymbol{{\xi}}_{\boldsymbol{\theta}}$). 
From the  differential PMP in (\ref{diffpontryagin}),  we note that to obtain $\frac{\partial  \boldsymbol{{\xi}}_{\boldsymbol{{\theta}}}}{\partial \boldsymbol{{\theta}}}$ in (\ref{unknown}), it is sufficient to compute the unknowns $\left\{\frac{\partial \boldsymbol{{x}}^{\boldsymbol{\theta}}_{0:T}}{\partial \boldsymbol{\theta}},  \frac{\partial \boldsymbol{{x}}^{\boldsymbol{\theta}}_{0:T-1}}{\partial \boldsymbol{\theta}},\frac{\partial \boldsymbol{{\lambda}}^{\boldsymbol{\theta}}_{1:T}}{\partial \boldsymbol{\theta}}   \right\}$ in (\ref{diffpontryagin}).  Next we will show that how these unknowns are elegantly solved by introducing a new system.

\subsection{Auxiliary Control System}
One important observation to the differential PMP in (\ref{diffpontryagin}) is that it shares a similar structure to the original PMP in (\ref{mp}); so it can be viewed as a new set of PMP equations corresponding to an `oracle control optimal system' whose the `optimal trajectory' is exactly (\ref{unknown}). This motivates us to `unearth' this oracle optimal control system,  because by doing so,  (\ref{unknown}) can be obtained from this oracle system by an OC solver. To this end, we first define the new `state' and 'control' (matrix) variables:
\begin{align}\label{vairablemat}
\small{{X}}_t=\small\frac{\partial \boldsymbol{{x}}_{t}}{\partial \boldsymbol{\theta}}\in\mathbb{R}^{n\times r},\quad\quad \small{{U}}_t=\small\frac{\partial \boldsymbol{{u}}_{t}}{\partial \boldsymbol{\theta}}\in\mathbb{R}^{m\times r},
\end{align}  
respectively. Then, we `artificially' define the following \emph{auxiliary control system}  $\boldsymbol{\overline\Sigma}(\boldsymbol{{\xi}}_{\boldsymbol{\theta}})$:

\begin{longfbox}[padding-top=-3pt,margin-top=-5pt,padding-bottom=-6pt,margin-bottom=-6pt,]
\mathleft
\small{
	\begin{equation}
\boldsymbol{\overline{\Sigma}}(\boldsymbol{{\xi}}_{\boldsymbol{\theta}}):\quad\quad
\begin{aligned}
\text{dynamics:} &\quad {{X}}_{t+1}=
F_t{{X}}_{t}+G_t{{U}}_{t}+E_t \quad \text{with} \quad {{X}}_0=\boldsymbol{0}, 
\\
\text{control objective:}& \quad \bar{J}=\Tr\sum_{t=0}^{T-1}\Bigg(\frac{1}{2}\small\begin{bmatrix}
{{X}}_{t}\\[3pt]
{{U}}_{t}
\end{bmatrix}^\prime\begin{bmatrix}
H_t^{xx} & H_t^{xu}\\[3pt]
H_t^{ux}& H_t^{uu}
\end{bmatrix}\begin{bmatrix}
{{X}}_{t}\\[3pt]
{{U}}_{t}
\end{bmatrix}+\begin{bmatrix}
H_t^{xe}\\[3pt]
H_t^{ue}
\end{bmatrix}^\prime\begin{bmatrix}
{{X}}_{t}\\[3pt]
{{U}}_{t}
\end{bmatrix}\Bigg)\\
&\quad\quad\quad+\Tr\left(\frac{1}{2}{{X}}_{T}^\prime \, H_T^{xx} \,{{U}}_{T}+ (H_T^{xe})^\prime\,{{X}}_{T}\right).\label{backoc}
\end{aligned}
\end{equation}
}
	\mathcenter
\end{longfbox}
Here, $X_0=\frac{\partial\boldsymbol{x}_0}{\partial \boldsymbol{\theta}}=\boldsymbol{0}$ because $\boldsymbol{x}_0$ in (\ref{oc}) is given;  $\bar{J}$ is the defined control objective function which needs to be optimized in the auxiliary control system; and $\Tr$ denotes matrix trace. Before presenting the key results, we make some comments on the above  auxiliary control system $\boldsymbol{\overline\Sigma}(\boldsymbol{{\xi}}_{\boldsymbol{\theta}})$. First, its state and control variables  are both matrix variables  defined in (\ref{vairablemat}). Second, its dynamics is linear and control objective function $\bar{J}$ is quadratic, for which the coefficient matrices  are given in (\ref{mats}). Third, its dynamics and objective function are determined by the trajectory $\boldsymbol{{\xi}}_{\boldsymbol{{\theta}}}$ of the system $\boldsymbol{\Sigma}(\boldsymbol{\theta})$ in forward pass, and this is why we denote it as $\boldsymbol{\overline\Sigma}(\boldsymbol{{\xi}}_{\boldsymbol{\theta}})$. Finally, we have the following important result.
\begin{lemma}\label{theorem0}
	Let $\small\{{{X}}_{0:T}^{\boldsymbol{\theta}},{{U}}_{0:T-1}^{\boldsymbol{\theta}}\}$  be  a stationary solution to the   auxiliary control system $\boldsymbol{\overline\Sigma}(\boldsymbol{{\xi}}_{\boldsymbol{\theta}})$ in (\ref{backoc}). Then,  $\small\{{{X}}_{0:T}^{\boldsymbol{\theta}},{{U}}_{0:T-1}^{\boldsymbol{\theta}}\}$  satisfies  Pontryagin's Maximum Principle of $\boldsymbol{\overline\Sigma}(\boldsymbol{{\xi}}_{\boldsymbol{\theta}})$, which is  (\ref{diffpontryagin}), and 
		\begin{equation}\label{uknown2}
		\{{{X}}_{0:T}^{\boldsymbol{\theta}},{{U}}_{0:T-1}^{\boldsymbol{\theta}}\}=\small\left\{\frac{\partial \boldsymbol{{x}}^{\boldsymbol{\theta}}_{0:T}}{\partial \boldsymbol{\theta}},\frac{\partial \boldsymbol{{u}}^{\boldsymbol{\theta}}_{0:T-1}}{\partial \boldsymbol{\theta}}\right\}= \frac{\partial \boldsymbol{{\xi}}_{\boldsymbol{\theta}}}{\partial \boldsymbol{\theta}}.
		\end{equation}
\end{lemma}
A proof of Lemma \ref{theorem0} is in Appendix \ref{appendix1}.  Lemma \ref{theorem0} states two assertions. First, the PMP condition for the auxiliary control system $\boldsymbol{\overline\Sigma}(\boldsymbol{{\xi}}_{\boldsymbol{\theta}})$ is exactly the differential PMP in (\ref{diffpontryagin}) for the original system $\boldsymbol{\Sigma}(\boldsymbol{\theta})$; and second, importantly, the trajectory $\small\{{{X}}_{0:T}^{\boldsymbol{\theta}},{{U}}_{0:T-1}^{\boldsymbol{\theta}}\}$ produced by the auxiliary control system $\boldsymbol{\overline\Sigma}(\boldsymbol{{\xi}}_{\boldsymbol{\theta}})$ is exactly the derivative of trajectory of the original system $\boldsymbol{\Sigma}(\boldsymbol{\theta})$ with respect to the parameter $\boldsymbol{\theta}$. Based on Lemma \ref{theorem0},  we can obtain  $\frac{\partial  \boldsymbol{{\xi}}_{\boldsymbol{{\theta}}}}{\partial \boldsymbol{{\theta}}}$ from $\boldsymbol{\overline\Sigma}(\boldsymbol{{\xi}}_{\boldsymbol{\theta}})$ efficiently by the lemma below.

\begin{lemma}\label{theorem1}
	If ${H_t^{uu}}$  in (\ref{backoc}) is invertible for all $t=0,1\cdots,T-1$,   define the following  recursions
\begin{normalsize}
		\begin{subequations}\label{ricc}
		\begin{align}
		{{P}}_t&={{Q}}_t+{{A}}_t^\prime({I}+{{P}}_{t+1}{{R}}_t)^{-1}{{P}}_{t+1}{{A}}_t,\label{ricc.1}\\
		{{W}}_t&={{A}}_t^\prime({I}+{{P}}_{t+1}{{R}}_t)^{-1}({{W}}_{t+1}{+{P}}_{t+1}{{M}}_{t})+{{N}}_t, \label{ricc.2}
		\end{align}
	\end{subequations}
\end{normalsize}%
	with   $\small{{P}}_T=H_T^{xx}$ and $\small{{W}}_T=H_T^{xe}.$
	Here, $I$ is identity matrix, $\small{{A}}_t{=}F_t-G_t(H_t^{uu})^{\text{-}1}H_t^{ux},
		{{R}}_t=G_t(H_t^{uu})^{\text{-}1}G_t^\prime, 
		{{M}}_t{=}E_t{-}G_t(H_t^{uu})^{\text{-}1}H_t^{ue},
		{{Q}}_t{=}H_t^{xx}{-}H_t^{xu}(H_t^{uu})^{\text{-}1}H_t^{ux}, 
		{{N}}_t{=}H_t^{xe}{-}H_t^{xu}(H_t^{uu})^{\text{-}1}H_t^{ue}$
	are all known given  (\ref{mats}). Then, the stationary solution  $\small\{{{X}}_{0:T}^{\boldsymbol{\theta}},{{U}}_{0:T-1}^{\boldsymbol{\theta}}\}$ in (\ref{uknown2}) can be  obtained by iteratively solving the following equations from $t=0$ to $T-1$ \text{with} $  {{X}}^{\boldsymbol{\theta}}_0={{X}}_0=\boldsymbol{0}$:
\begin{small}
	\begin{subequations}\label{iter}
		\begin{align}
		{{U}}^{\boldsymbol{\theta}}_t&=-(H_t^{uu})^{\text{-}1}\left(
		H_t^{ux}{{X}}^{\boldsymbol{\theta}}_t+H_t^{ue} +
		{G_t}^\prime({I}+{{P}}_{t+1}{{R}}_t)^{-1}\Big({{P}}_{t+1}{{A}}_{t}{{X}}^{\boldsymbol{\theta}}_t+{{P}}_{t+1}{{M}}_{t}+{{W}}_{t+1}\Big)
		\right),
		\label{iter.2}\\[3pt]
		{{X}}^{\boldsymbol{\theta}}_{t+1}&=
		F_t{{X}}^{\boldsymbol{\theta}}_{t}+G_t{{U}}^{\boldsymbol{\theta}}_{t}+E_t    \label{iter.3}.
		\end{align}
	\end{subequations}
\end{small}
\end{lemma}
A proof of Lemma \ref{theorem1} is in Appendix \ref{appendix2}. Lemma \ref{theorem1} states that the trajectory  of the above auxiliary  control system $\boldsymbol{\overline\Sigma}(\boldsymbol{{\xi}}_{\boldsymbol{\theta}})$ can be obtained by two steps: first, iteratively solve  (\ref{ricc}) backward in time to obtain  matrices ${P}_t$ and ${{W}}_t$ (all other coefficient matrices are known given $\boldsymbol{\overline\Sigma}(\boldsymbol{{\xi}}_{\boldsymbol{\theta}})$); second, calculate $\small\{{{X}}_{0:T}^{\boldsymbol{\theta}},{{U}}_{0:T-1}^{\boldsymbol{\theta}}\}$ by iteratively integrating a feedback-control system (\ref{iter}) forward in time. In fact, these two steps constitute the standard procedure to solve general finite-time LQR problems \cite{kwakernaak1972linear}.

As  a conclusion to the techniques developed in Section \ref{DPMPAUX}, in Algorithm \ref{ag_auxsys_solver} we summarize the procedure of computing $\small\frac{\partial  \boldsymbol{{\xi}}_{\boldsymbol{{\theta}}}}{\partial \boldsymbol{{\theta}}}$  via the introduced auxiliary control system. Algorithm \ref{ag_auxsys_solver}  serves as a key component in the backward pass of  the PDP learning framework, as shown in Fig. \ref{overview}. 

\vspace{-4mm}
\begin{algorithm2e}[h]
	\small
	\SetKwComment{Comment}{$\triangleright$\ }{}
	\SetKwInput{given}{Input}
	\given{The trajectory $\boldsymbol{{\xi}}_{\boldsymbol{\theta}}$ in (\ref{octraj}) produced by the  system  $\boldsymbol{\Sigma}(\boldsymbol{\theta})$ in (\ref{oc}) in the forward pass. %\Comment*[r]{solve (\ref{oc})} 
	}
	
	\Indp\Indp\Indp Compute the coefficient matrices (\ref{mats}) to obtain the auxiliary control system  $\boldsymbol{\overline\Sigma}(\boldsymbol{{\xi}}_{\boldsymbol{\theta}})$ in (\ref{backoc})\;
	 Solve the auxiliary control system  $\boldsymbol{\overline\Sigma}(\boldsymbol{{\xi}}_{\boldsymbol{\theta}})$ to obtain $\{{{X}}_{0:T}^{\boldsymbol{\theta}},{{U}}_{0:T-1}^{\boldsymbol{\theta}}\}$ using Lemma \ref{theorem1}\;

	\Indm\Indm\Indm
	\SetKwInput{return}{Return}
	\return{$\frac{\partial  \boldsymbol{{\xi}}_{\boldsymbol{{\theta}}}}{\partial \boldsymbol{{\theta}}}=\{{{X}}_{0:T}^{\boldsymbol{\theta}},{{U}}_{0:T-1}^{\boldsymbol{\theta}}\}$}

	\caption{Solving $\frac{\partial  \boldsymbol{{\xi}}_{\boldsymbol{{\theta}}}}{\partial \boldsymbol{{\theta}}}$ using Auxiliary Control System \quad (\emph{See details in Appendix \ref{appendixalgorithem}})}\label{ag_auxsys_solver}
\end{algorithm2e}
\setlength{\textfloatsep}{5pt}

\vspace{-4mm}

\section{Applications to Different Learning Modes and Experiments} \label{applicationandexperiments}

\vspace{-3mm}

We investigate three learning modes of  PDP, as described in Section \ref{problemformulation}. For each mode, we demonstrate its capability in four environments listed in Table \ref{experimenttable}, and a baseline and a state-of-the-art method are compared. Both PDP and environment codes are available at \url{https://github.com/wanxinjin}.

\vspace*{-4.5mm}
\begin{table}[h]
	\captionsetup[table]{skip=-10pt}  
	\caption{Experimental environments (results for 6-DoF rocket  landing is in Appendix \ref{rocketexperiment})}
	\label{experimenttable}
	\centering
	\begin{tabular}{lll}
		\toprule
		\small Systems     & \small Dynamics parameter $\small\boldsymbol{\theta}_\text{dyn}$    & \small Control objective parameter $\small\boldsymbol{\theta}_\text{obj}$  \\
		\midrule
		\small Cartpole     & \small cart mass, pole mass and length & \small \multirow{4}{*}{
			\makecell{$c(\boldsymbol{x},\boldsymbol{u}){=}\norm{\boldsymbol{\theta}_\text{obj}^\prime(\boldsymbol{x}-\boldsymbol{x}_{\text{g}})}^2{+}\norm{\boldsymbol{u}}^2$\\ $h(\boldsymbol{x},\boldsymbol{u})=\norm{\boldsymbol{\theta}_\text{obj}^\prime(\boldsymbol{x}-\boldsymbol{x}_{\text{g}})}^2$}
		}   \\
		\small Two-link robot arm     & \small length and mass for each link     &  \\
		\small 6-DoF  quadrotor maneuvering & \small mass, wing length, inertia matrix & \\
		\small 6-DoF rocket powered landing & \small mass, rocket length, inertia matrix & \\
		\bottomrule
	\end{tabular}
	
	{\raggedright \small{We fix the unit weight to $\norm{\boldsymbol{u}}^2$, because estimating all weights will incur ambiguity \cite{keshavarz2011imputing}}; $\boldsymbol{x}_{\text{g}}$ is the goal state. \par}
\end{table}

\vspace*{-2.5mm}
\textbf{IRL/IOC Mode.}
The parameterized  $\boldsymbol{\Sigma}(\boldsymbol{\theta})$ is in (\ref{oc}) and the loss  in (\ref{lossioc}). In the forward pass of  PDP,  $\boldsymbol{\xi}_{\boldsymbol{\theta}}$ is solved from  $\boldsymbol{\Sigma}(\boldsymbol{\theta})$ by any OC solver. In the backward pass, $\small\frac{\partial \boldsymbol{\xi}_{\boldsymbol{\theta}}}{\partial \boldsymbol{\theta}}$ is computed from the auxiliary control system $\boldsymbol{\overline\Sigma}(\boldsymbol{{\xi}}_{\boldsymbol{\theta}})$ in (\ref{backoc})  using Algorithm \ref{ag_auxsys_solver}. The full algorithm is in Appendix \ref{appendixalgorithem}.

\begin{figure}[h]
	\begin{subfigure}{.245\textwidth}
		\centering
		\includegraphics[width=\linewidth]{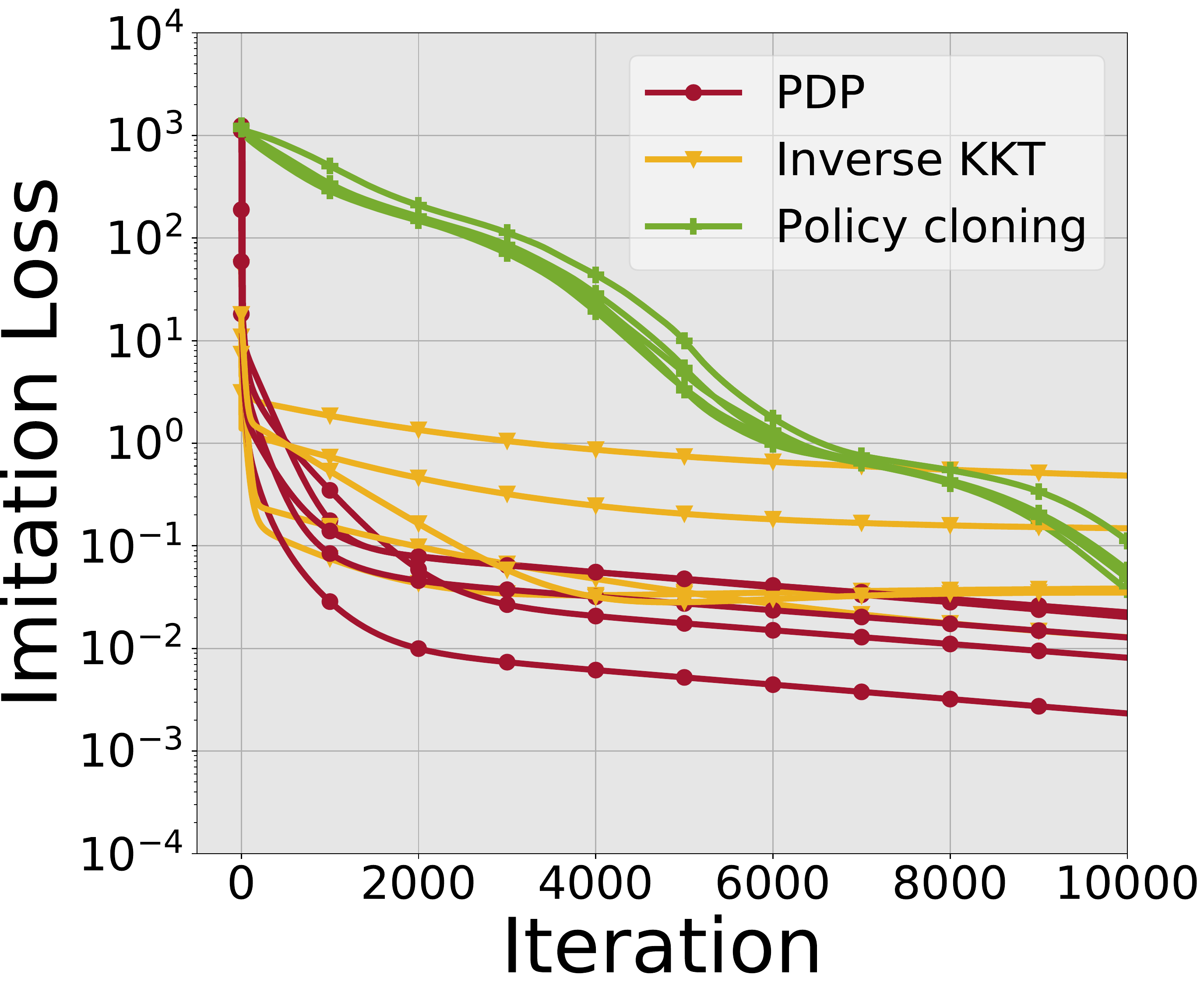}
		\caption{Cart-pole}
		\label{figioc.1}
	\end{subfigure}
	\begin{subfigure}{.245\textwidth}
		\centering
		\includegraphics[width=\linewidth]{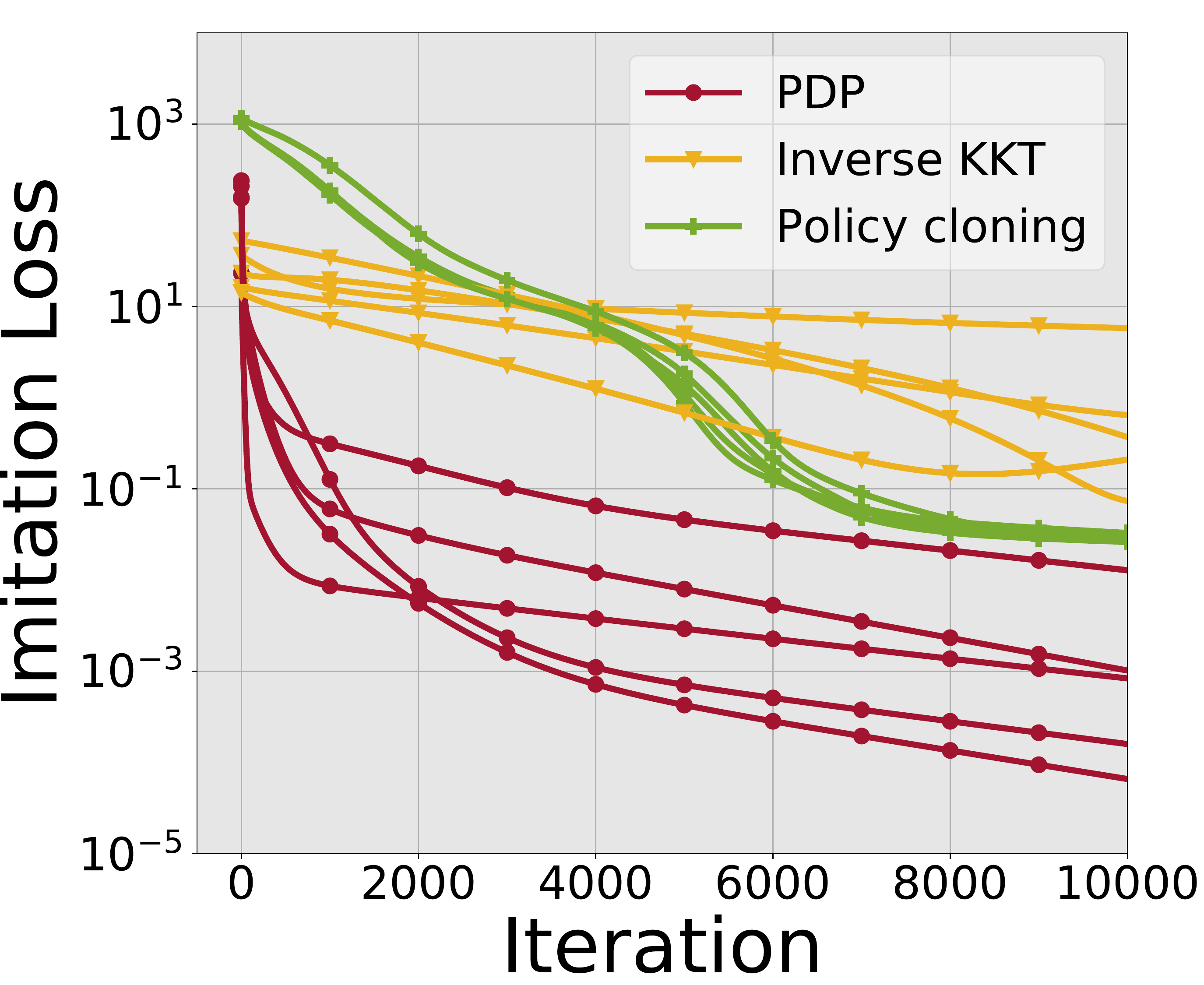}
		\caption{Robot arm}
		\label{figioc.2}
	\end{subfigure}%
	\begin{subfigure}{.245\textwidth}
		\centering
		\includegraphics[width=\linewidth]{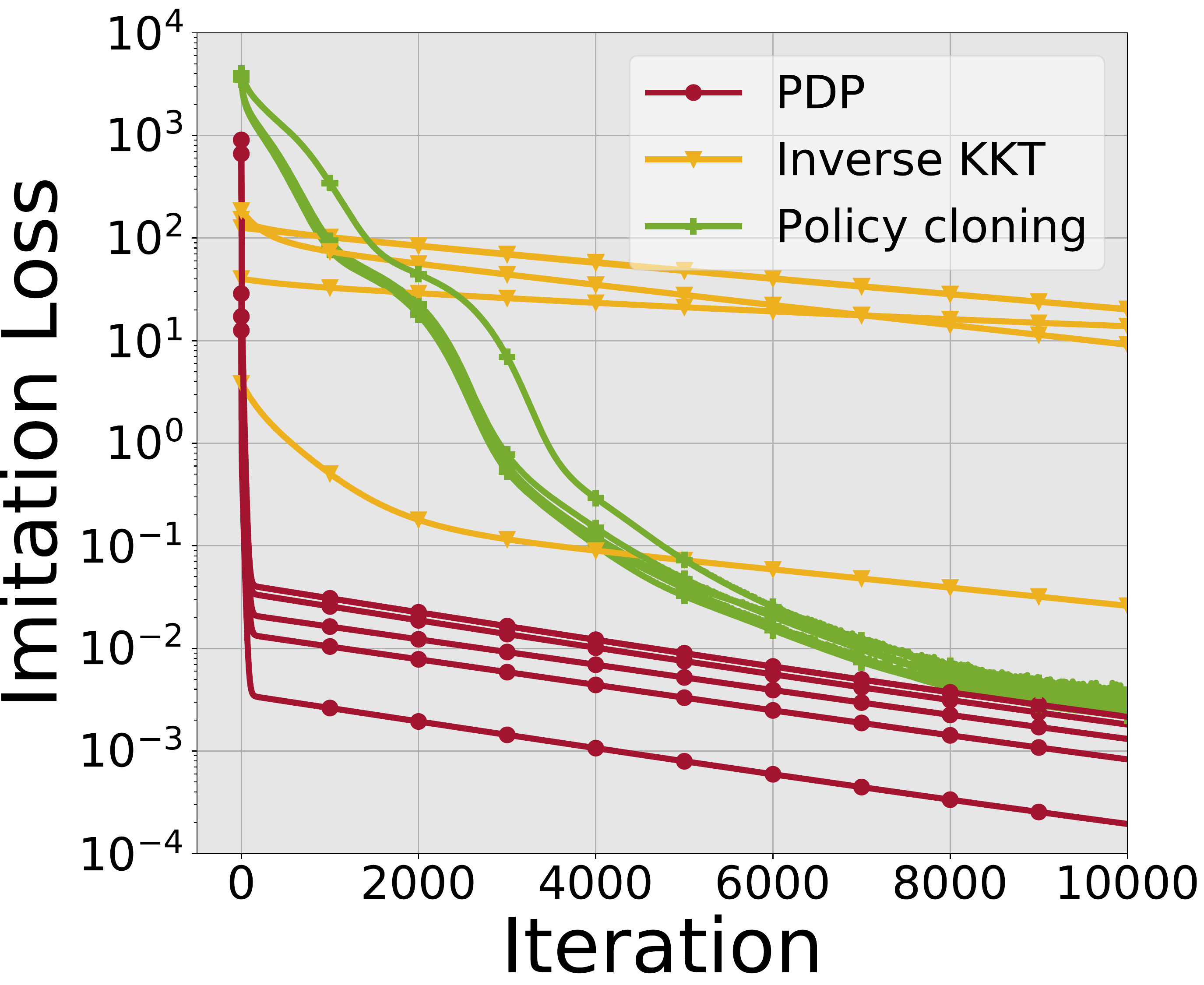}
		\caption{Quadrotor}
		\label{figioc.3}
	\end{subfigure}
	\begin{subfigure}{.245\textwidth}
		\centering
		\includegraphics[width=\linewidth]{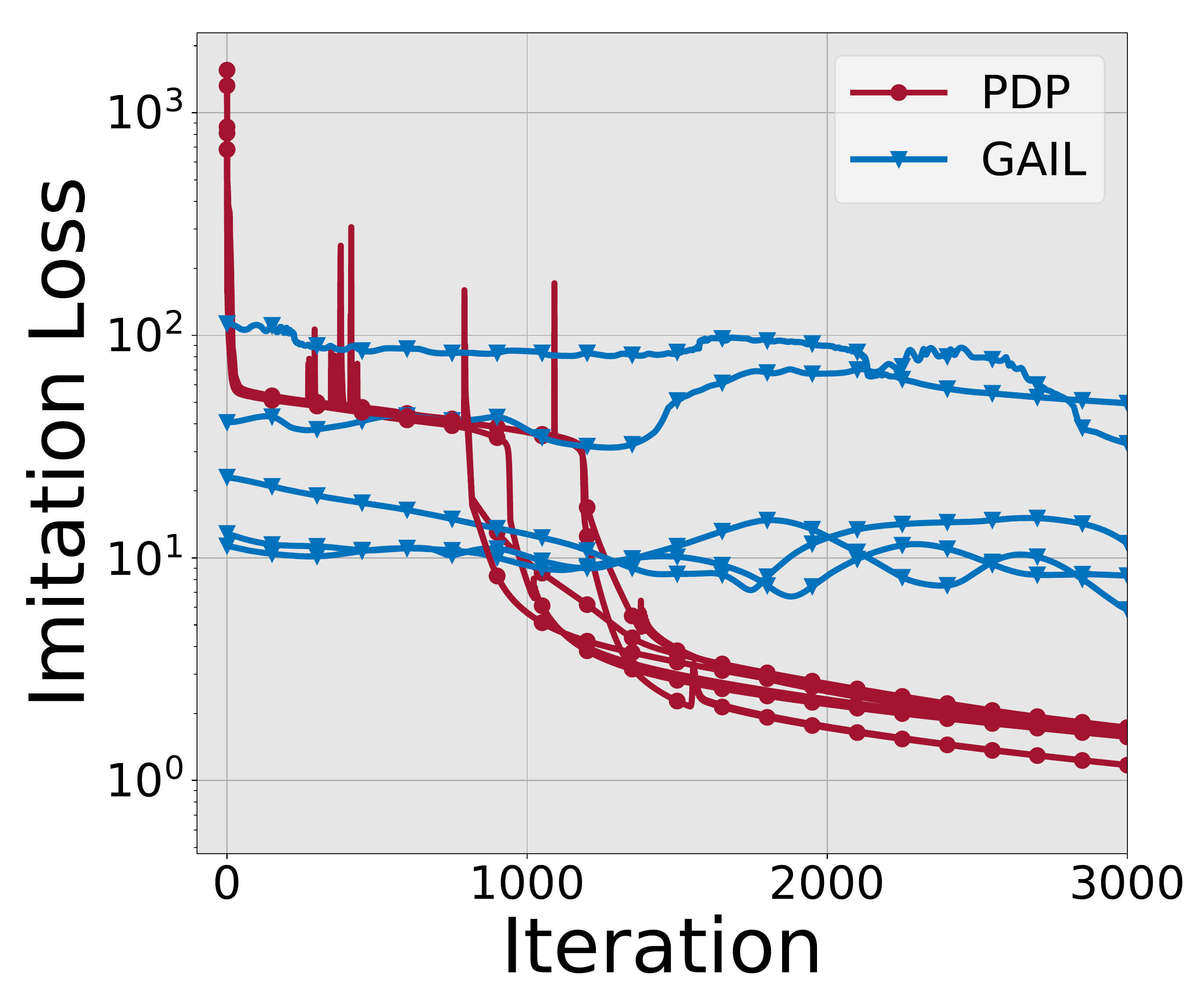}
		\caption{Comparison}
		\label{figioc.4}
	\end{subfigure}\vspace*{-1mm}
	\caption{(a-c) imitation loss v.s. iteration, (d) PDP learns a neural  objective function and comparison.}
	\label{figioc}
\end{figure}

\textbf{Experiment: imitation learning.} We use IRL/IOC Mode to solve imitation learning in environments in  Table \ref{experimenttable}. The true dynamics is parameterized, and   control objective is parameterized as a weighted distance to the goal, $\small\boldsymbol{\theta}=\{\boldsymbol{\theta}_{\text{dyn}},\boldsymbol{\theta}_{\text{obj}}\}$.
Set imitation loss $\small L(\boldsymbol{\xi}_{\boldsymbol{\theta}},\boldsymbol{\theta}){=} {\norm{\boldsymbol{{\xi}}^{\text{d}}-\boldsymbol{{\xi}}_{\boldsymbol{\theta}}}^2}.$ Two other methods are compared: (i) neural policy cloning, and (ii) inverse KKT  \cite{englert2017inverse}. We set learning rate $\eta=10^{-4}$  and run five trials given random initial  $\boldsymbol{\theta}_0$. The results in
Fig. \ref{figioc.1}-\ref{figioc.3} show that   PDP significantly outperforms the  policy cloning and inverse-KKT for a much lower training loss and   faster convergence. 
In Fig. \ref{figioc.4}, we apply the PDP to learn a neural control objective function for the robot arm using the same demonstration data  in Fig. \ref{figioc.2}, and we also compare with  the GAIL   \cite{ho2016generative}. Results in Fig. \ref{figioc.4} show that the PDP  successfully learns a neural objective function and the imitation loss of PDP is  much lower than that of GAIL. It should note that because the demonstrations are not strictly realizable (optimal)  under the parameterized neural objective function, the final loss for the  PDP is small but not zero. This indicates that given sub-optimal demonstrations,  PDP can still find the `best' control objective function within the  function set $J(\boldsymbol{\theta})$ such that its  reproduced $\boldsymbol{\xi}_{\boldsymbol{\theta}}$ has the \emph{minimal distance} to   the demonstrations. 
Please refer to Appendix \ref{apendix-exp-imitation} for more experiment details and additional validations.

\vspace*{-3mm}

\vspace*{-2mm}

\textbf{SysID Mode.} In this mode, $\boldsymbol{\Sigma}(\boldsymbol{\theta})$ is   (\ref{ocmodeid}) and loss is   (\ref{lossid}). PDP  is greatly simplified: in  forward pass,  $\boldsymbol{\xi}_{\boldsymbol{\theta}}$ is solved by integrating the difference equation (\ref{ocmodeid}). In the backward pass,  $\boldsymbol{\overline\Sigma}(\boldsymbol{\xi}_{\boldsymbol{\theta}})$  is reduced to

\vspace*{-0.2mm}

\begin{longfbox}[padding-top=-5pt, margin-top=-5pt, padding-bottom=1pt,  margin-bottom=-1pt]
	\mathleft
	\begin{equation}\label{iddiff}
	\boldsymbol{\overline\Sigma}(\boldsymbol{\xi}_{\boldsymbol{\theta}}):\qquad\qquad\qquad\text{dynamics:}\quad \small{{X}}_{t+1}^{\boldsymbol{{\theta}}}=
	F_t{{X}}^{\boldsymbol{{\theta}}}_{t}+E_t \quad \text{with} \quad {{X}}_{0}=\boldsymbol{0}.
	\end{equation} 
	\mathcenter
\end{longfbox}

This is because  $\boldsymbol{\Sigma}(\boldsymbol{\theta})$  in (\ref{ocmodeid})  results from letting $J(\boldsymbol{\theta})=0$,    (\ref{diffpontryagin.2}-\ref{diffpontryagin.4}) and $\bar{J}$ in (\ref{backoc}) are then trivialized, and due to $\boldsymbol{u}_{0:T-1}$  given,  $\small U_t^{\boldsymbol{\theta}}=\boldsymbol{0}$ in (\ref{diffpontryagin.1}). The algorithm is in Appendix \ref{appendixalgorithem}.

\begin{figure} [h]
	\begin{subfigure}{.245\textwidth}
		\centering
		\includegraphics[width=\linewidth]{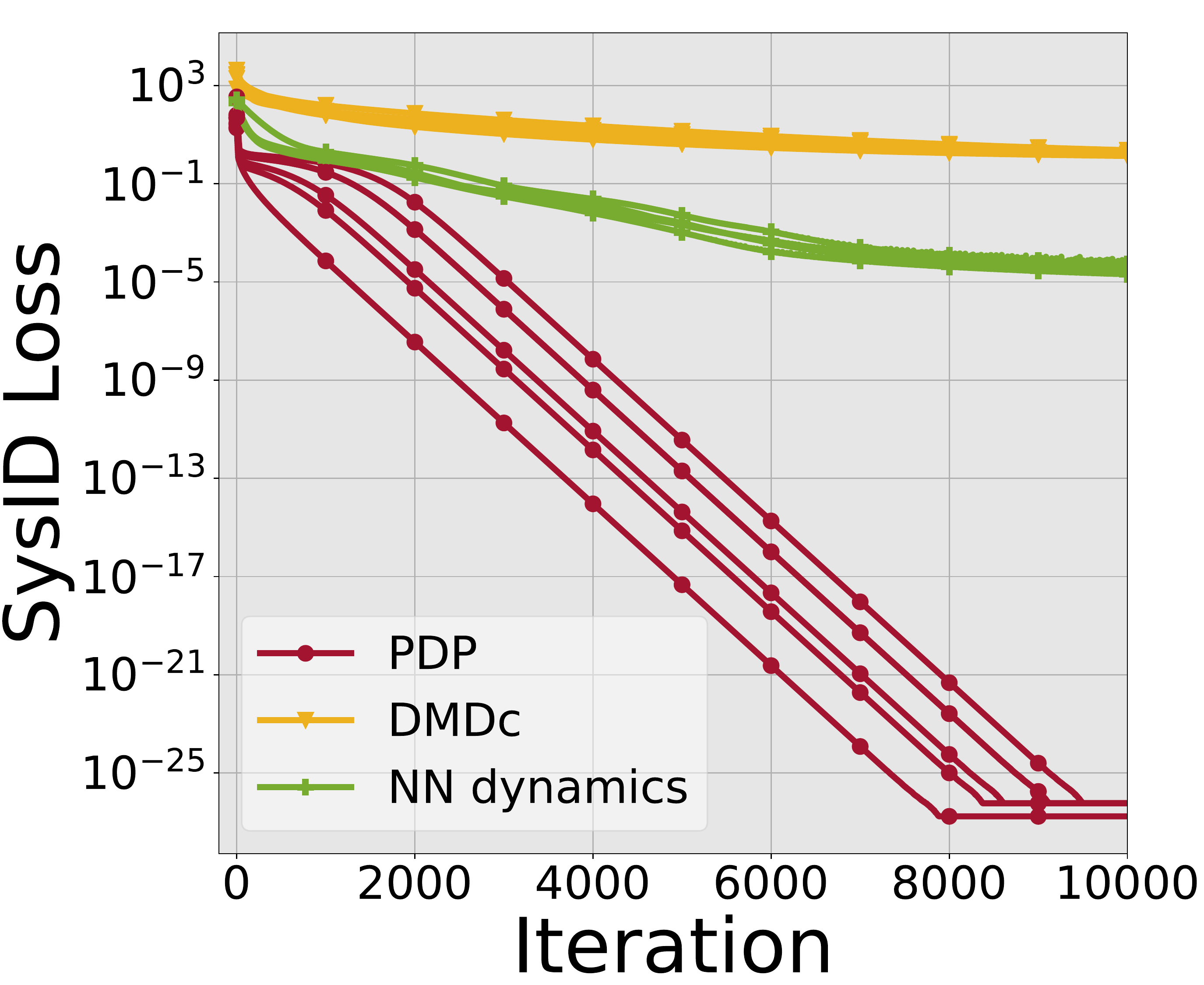}
		\caption{Cart-pole}
		\label{figid.1}
	\end{subfigure}
	\begin{subfigure}{.245\textwidth}
		\centering
		\includegraphics[width=\linewidth]{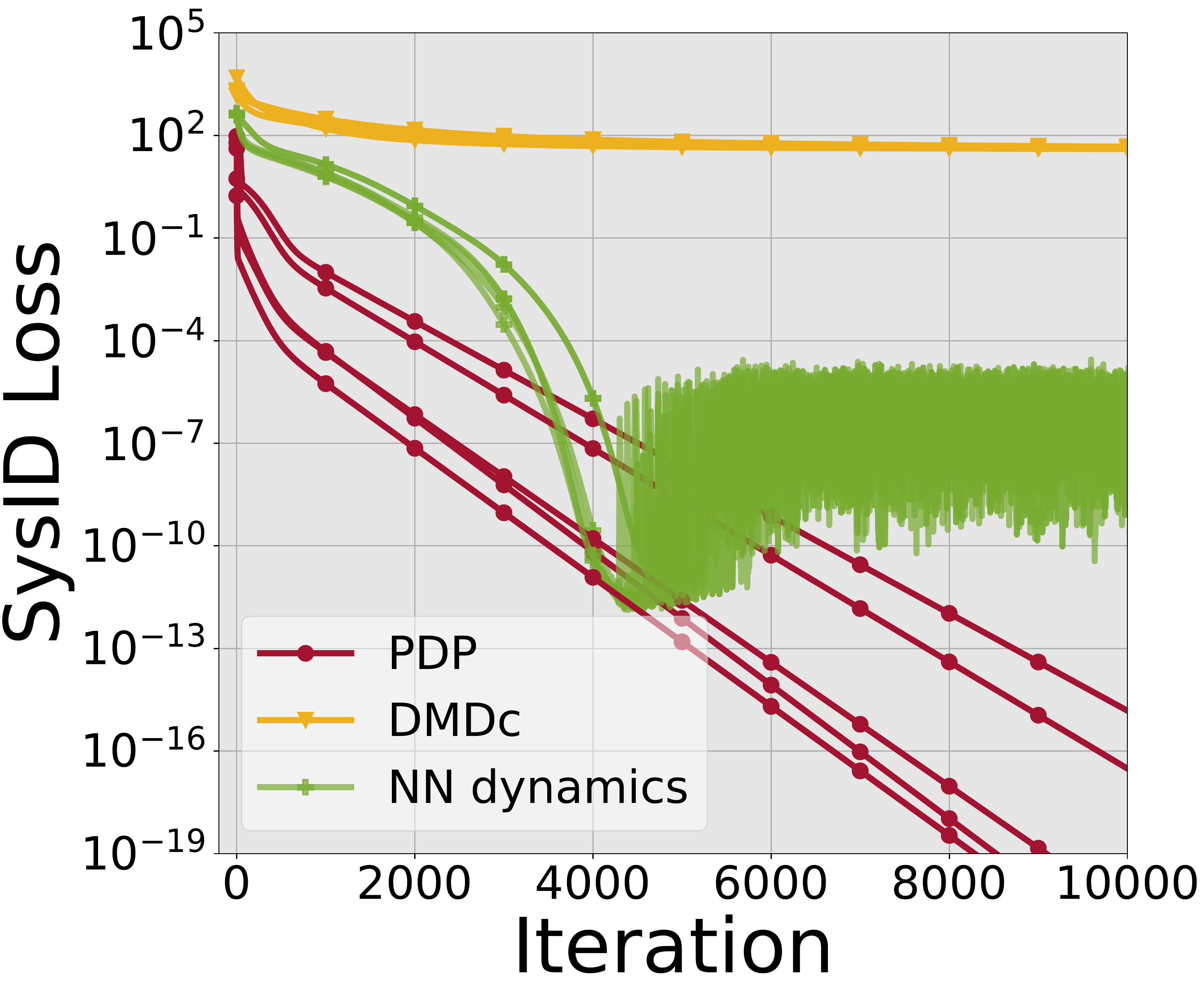}
		\caption{Robot arm}
		\label{figid.2}
	\end{subfigure}%
	\begin{subfigure}{.245\textwidth}
		\centering
		\includegraphics[width=\linewidth]{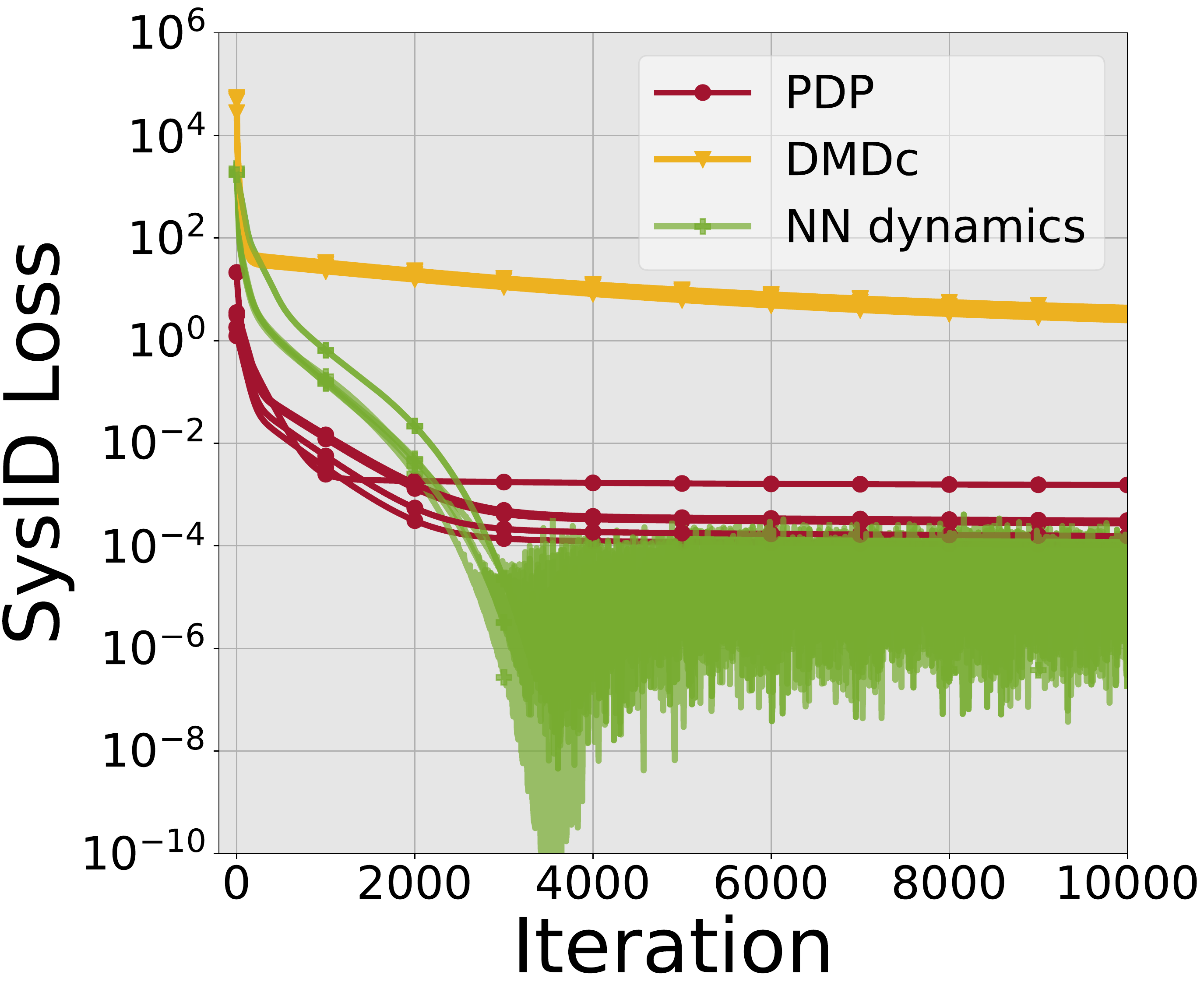}
		\caption{Quadrotor}
		\label{figid.3}
	\end{subfigure}
	\begin{subfigure}{.245\textwidth}
		\centering
		\includegraphics[width=\linewidth]{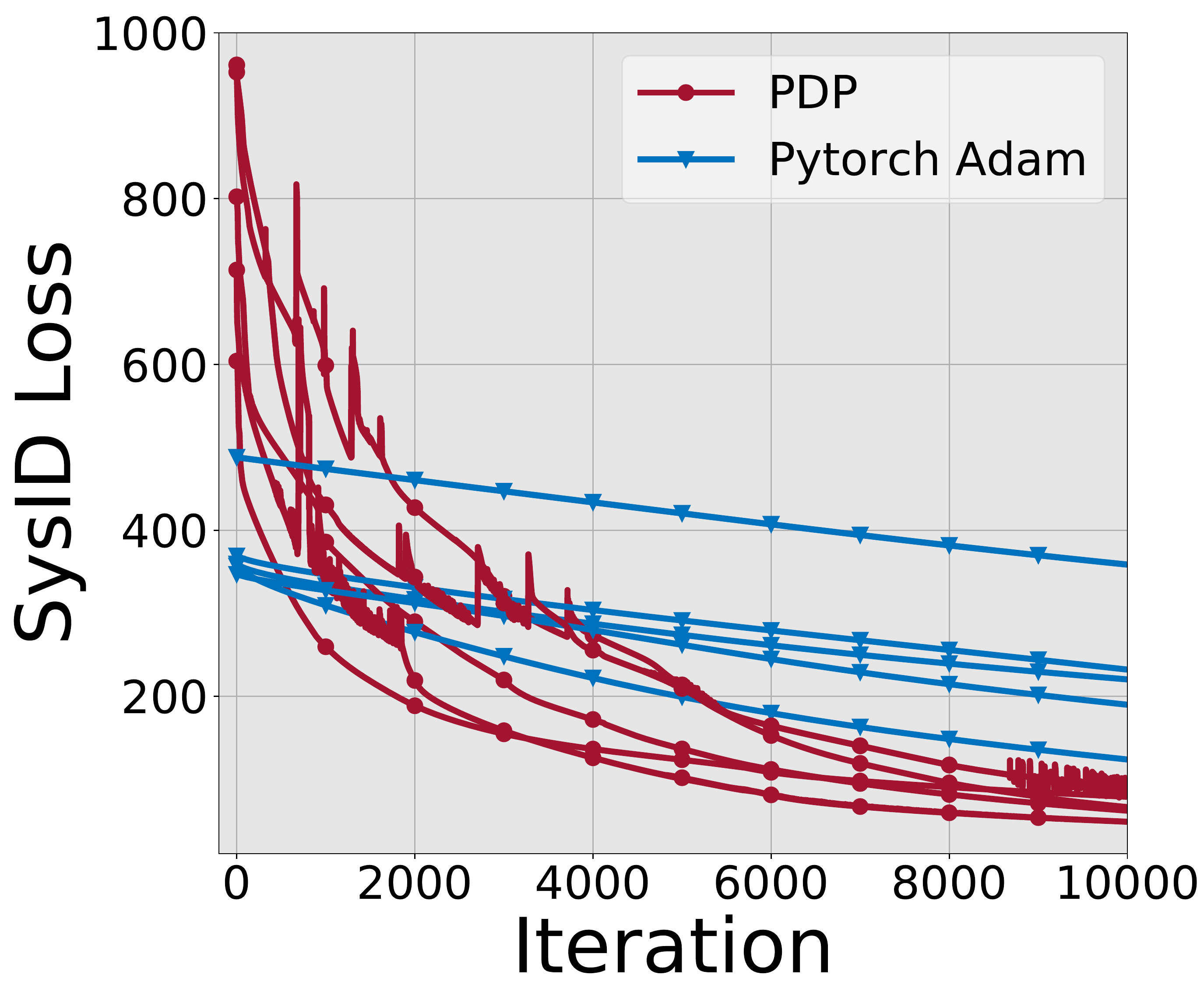}
		\caption{Learn neural dynamics}
		\label{figid.4}
	\end{subfigure}%
	\caption{(a-c) SysID loss v.s. iteration, (d) PDP learns a neural dynamics model.}
	\label{figid}
\end{figure}

\textbf{Experiment: system identification.} 
We use the SysID Mode to identify the  dynamics parameter $\boldsymbol{\theta}_{\text{dyn}}$ for the systems in Table~\ref{experimenttable}. 
Set the SysID loss  $L(\boldsymbol{\xi}_{\boldsymbol{\theta}},\boldsymbol{\theta})= {\norm{\boldsymbol{{\xi}}^{\text{o}}-\boldsymbol{{\xi}}_{\boldsymbol{\theta}}}^2}.$  Two other methods are compared: (i)  learning a neural network (NN) dynamics model, and (ii)  DMDc \cite{proctor2016dynamic}. For all methods, we set  learning rate $\eta=10^{-4}$, and run five trials with random $\boldsymbol{\theta}_0$. The results are  in Fig. \ref{figid}.
Fig. \ref{figid.1}-\ref{figid.3}  show an obvious advantage of  PDP over the NN baseline and DMDc in terms of lower training loss and faster convergence speed. In Fig. \ref{figid.4}, we  compare PDP and Adam \cite{kingma2015adam} (here both with $\eta=10^{-5}$) for training the same neural dynamics model  for the robot arm. The results again show that PDP outperforms  Adam for  faster learning speed and lower training loss. Such advantages are due to  that  PDP has injected an inductive bias of optimal control into  learning, making it more efficient for handling dynamical systems. More experiments and validations are in Appendix \ref{appendix-exp-sysid}.

\vspace*{-3mm}

\vspace*{-2mm}
\textbf{Control/Planning Mode.} The parameterized system  $\boldsymbol{\Sigma}(\boldsymbol{\theta})$ is (\ref{ocmodeplan}) and loss is (\ref{lossmodeplan}).
PDP for this mode is also  simplified. In forward pass,  $\boldsymbol{\xi}_{\boldsymbol{\theta}}$ is solved   by integrating a (controlled) difference equation (\ref{ocmodeplan}). In backward pass, $\bar{J}$ in the auxiliary control system (\ref{backoc}) is trivialized because we have considered $J(\boldsymbol{\theta})=0$ in (\ref{ocmodeplan}). Since the control is now given by  $\boldsymbol{u}_t=\boldsymbol{u}(t, \boldsymbol{x}_t, \boldsymbol{\theta})$,  $\small{{U}}_t^{\boldsymbol{\theta}}$ is obtained by differentiating the policy on both side with respect to $\boldsymbol{\theta}$, that is, $\small{{U}}^{\boldsymbol{\theta}}_{t}=U_t^{{x}}X_{t}^{\boldsymbol{\theta}}+U_t^{{e}}$ with $
\small{{U}}^x_t=\frac{\partial \boldsymbol{u}_t}{\partial \boldsymbol{x}_t}$ and $\small{{U}}_t^e=\frac{\partial \boldsymbol{u}_t}{\partial \boldsymbol{\theta}}$. Thus,  
\begin{longfbox}[padding-top=-3pt, margin-top=0pt, padding-bottom=-1pt]
	\mathleft
	\begin{equation}\label{mode3controlback}
	\boldsymbol{\overline\Sigma}(\boldsymbol{\xi}_{\boldsymbol{\theta}}):\qquad\qquad
	\begin{aligned}
	\text{dynamics:}&\quad \small{{X}}_{t+1}^{\boldsymbol{{\theta}}}=
	F_t{{X}}^{\boldsymbol{{\theta}}}_{t}+G_t{{U}}^{\boldsymbol{\theta}}_{t} \quad \text{with} \quad {{X}}_{0}=\boldsymbol{0},\\
	\text{control policy:} &\quad \quad\small{{U}}^{\boldsymbol{\theta}}_{t}=U_t^{{x}}X_{t}^{\boldsymbol{\theta}}+U_t^{{e}}.
	\end{aligned}
	\end{equation} 
	\mathcenter
\end{longfbox}  Integrating  (\ref{mode3controlback}) from $t=0$ to $T$ leads to
{$\small
	\{{{X}}_{0:T}^{\boldsymbol{\theta}},{{U}}_{0:T-1}^{\boldsymbol{\theta}}\}= \frac{\partial \boldsymbol{{\xi}}_{\boldsymbol{\theta}}}{\partial \boldsymbol{\theta}}.
	$
}\normalsize The algorithm is in Appendix \ref{appendixalgorithem}.

\begin{figure}[h]
	\begin{subfigure}{.245\textwidth}
		\centering
		\includegraphics[width=\linewidth]{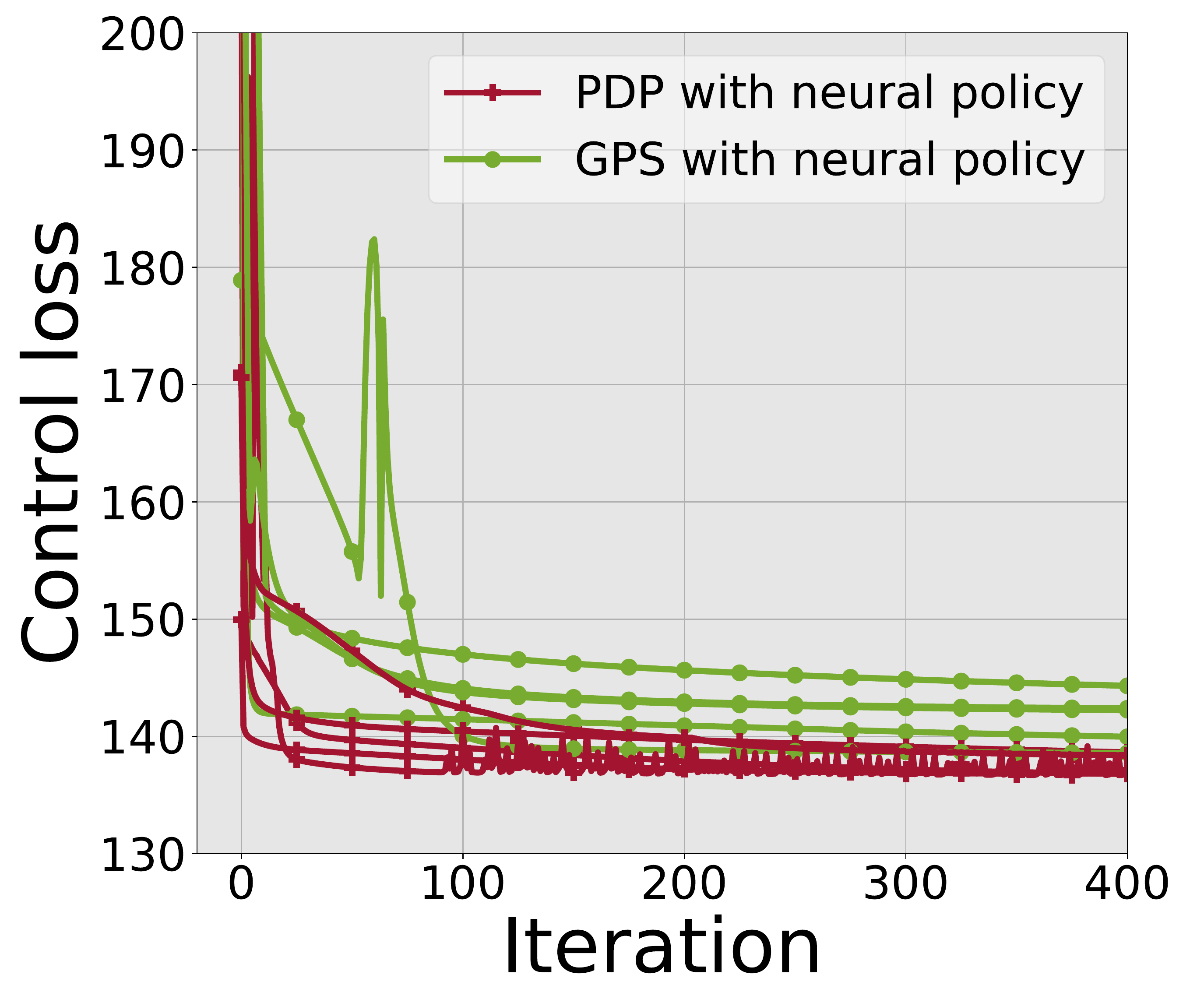}
		\caption{Cart-pole control}
		\label{figoc.1}
	\end{subfigure}
	\begin{subfigure}{.245\textwidth}
		\centering
		\includegraphics[width=\linewidth]{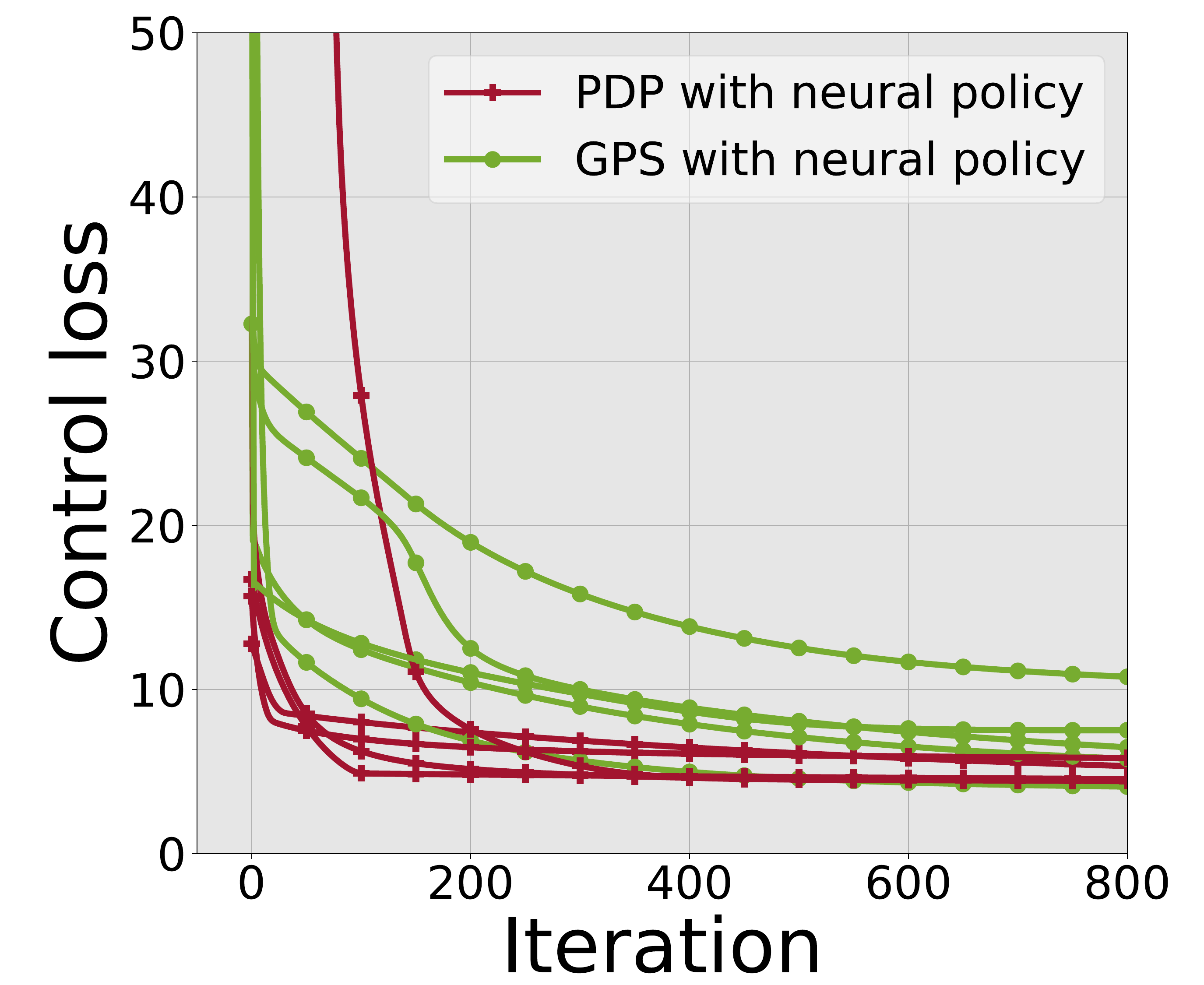}
		\caption{Robot arm control}
		\label{figoc.2}
	\end{subfigure}%
	\begin{subfigure}{.245\textwidth}
		\centering
		\includegraphics[width=\linewidth]{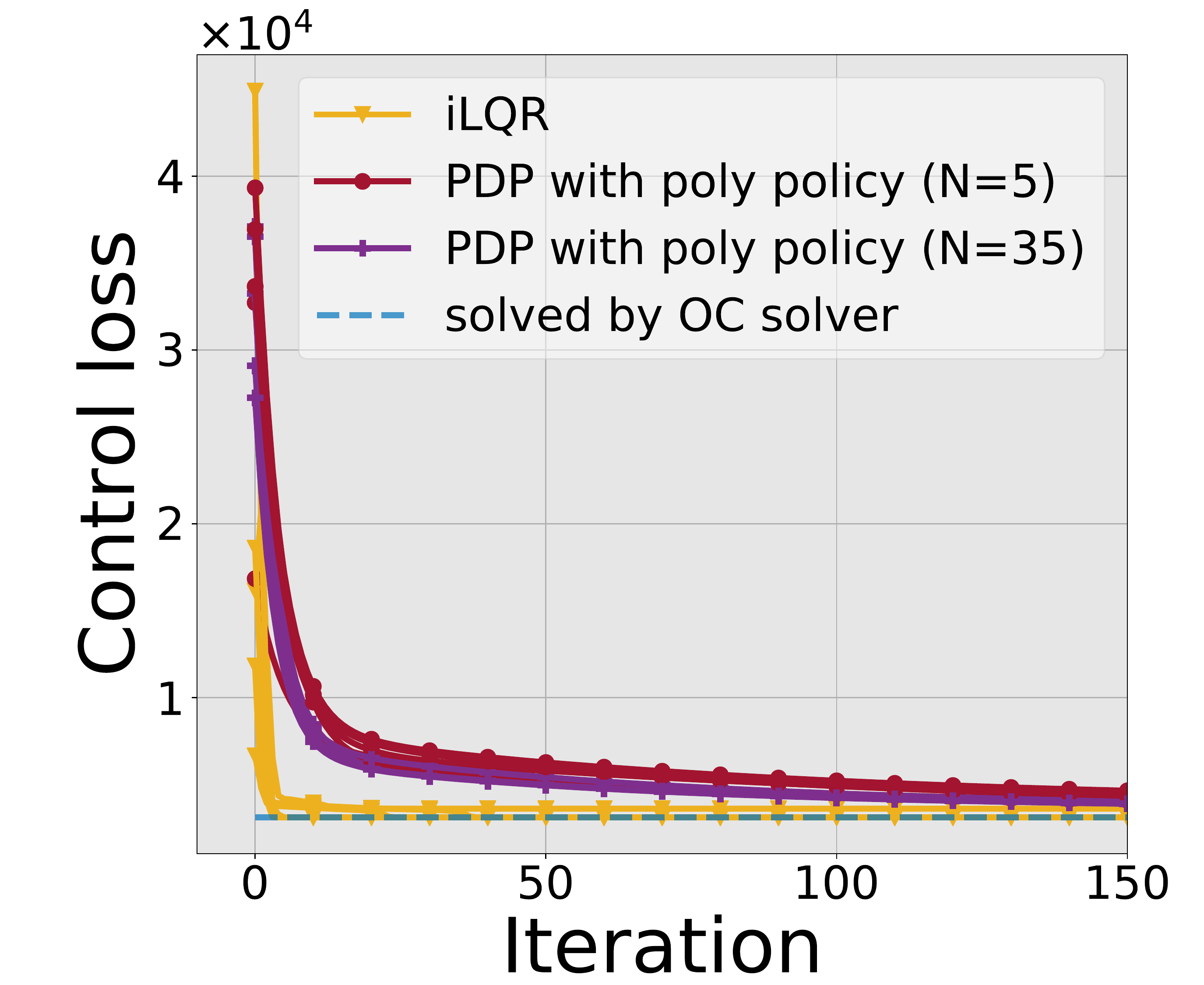}
		\caption{Quadrotor  planning}
		\label{figoc.3}
	\end{subfigure}
	\begin{subfigure}{.245\textwidth}
		\centering
		\includegraphics[width=\linewidth]{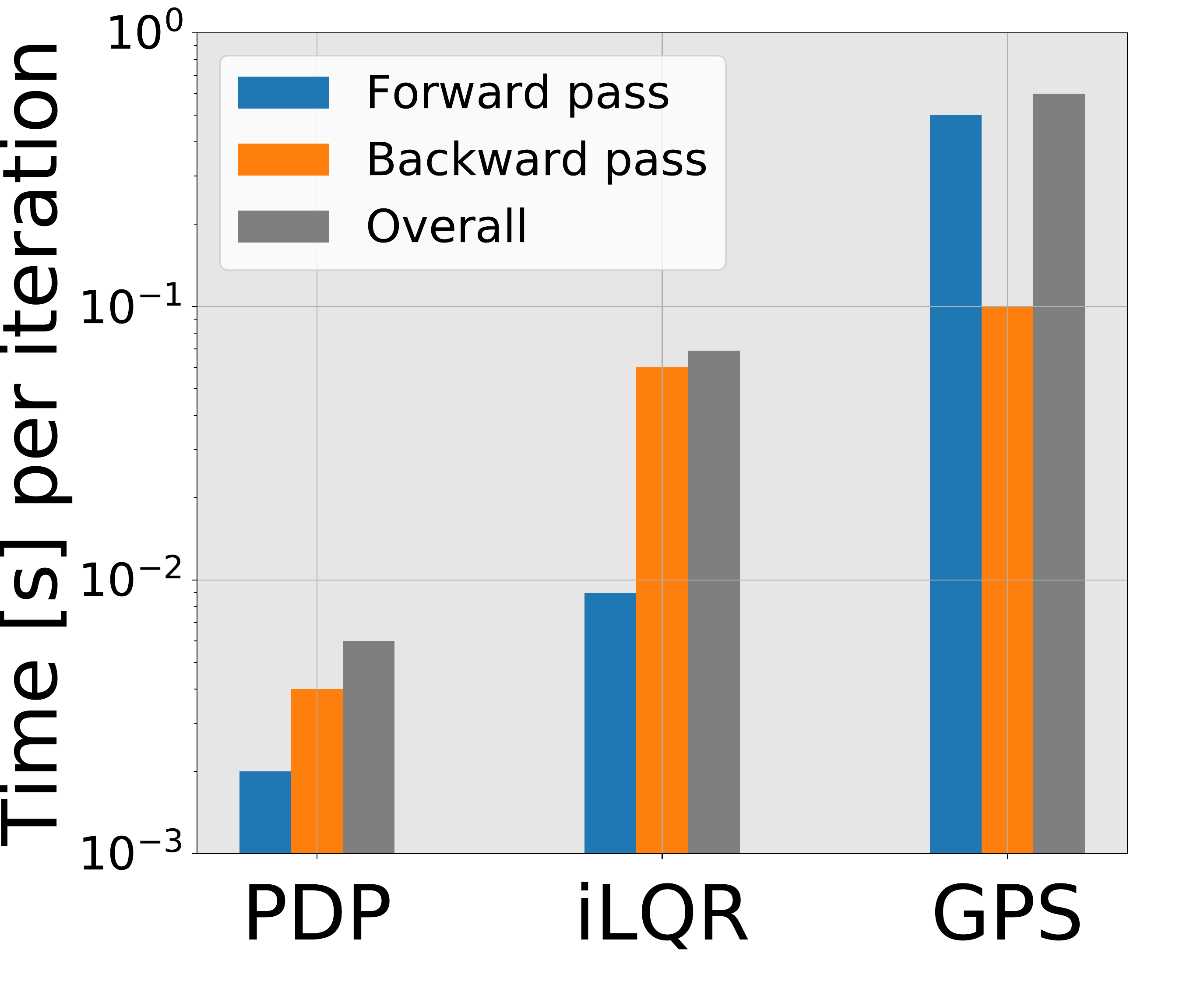}
		\caption{Timing results}
		\label{figoc.4}
	\end{subfigure}%
	\caption{(a-c) control loss v.s. iteration, (d) comparison for   running time per iteration.}
	\label{figoc}
\end{figure}

\textbf{Experiment: control and planning.} Based on  identified dynamics,  we learn policies of each system to optimize a control objective with given $\boldsymbol{\theta}_{\text{obj}}$. We set  loss  (\ref{lossmodeplan})  as the control objective (below called control loss). To  parameterize  policy   (\ref{ocmodeplan}), we use a Lagrange polynomial  of  degree $N$ (for planning) or  neural network (for feedback control). iLQR \cite{li2004iterative} and guided policy search (GPS) \cite{levine2013guided} are compared.  We set  learning rate  $\eta{=}10^{-4}$ or $10^{-6}$ and run five trials for each system.  Fig. \ref{figoc.1}-\ref{figoc.2} are learning  neural network feedback policies for the cart-pole and robot arm, respectively. The results show that  PDP outperforms  GPS for having lower   control loss. Fig.~\ref{figoc.3} is  motion   planning for quadrotor using a polynomial policy. It shows that  PDP  achieves a competitive performance with iLQR. Compared to iLQR,   PDP minimizes over  polynomial policies instead of input sequences, and thus has a higher final loss which depends on the expressiveness of the polynomial: e.g., the polynomial  of degree $N{=}35$ has a lower loss than that of $N{=}5$. Since iLQR can be viewed as `$1.5$-order'  method (discussed in Section \ref{background}), it has  faster converging speed than PDP which is only first-order, as shown in Fig.~\ref{figoc.3}. But iLQR is computationally extensive,  PDP, instead, has a huge advantage of  running time, as illustrated in Fig. \ref{figoc.4}. Due to space constraint, we put detailed analysis between GPS and PDP in Appendix \ref{appendix_exp_pdp_oc}.

\vspace*{-3mm}

\section{Discussion} \label{secton-comaprison-end-to-end-framework}

\textbf{The related end-to-end learning frameworks.} Two  lines of recent work are related to PDP. One is the recent work \cite{amos2017optnet,wang2019satnet,wilder2019melding,de2018end,donti2017task} that seeks to replace  a layer within a deep neural network  by an \emph{argmin layer},  in order to  capture the information flow characterized by a  solution of an optimization. Similar to PDP, these methods differentiate the argmin layer through KKT conditions.  They mainly focus on static optimization problems, which can not directly be applied to dynamical  systems. The second line is the recent RL development \cite{okada2017path,pereira2018mpc,amos2018differentiable,srinivas2018universal} that  embeds an implicit planner within a policy.  The idea is analogous to MPC, because using a predictive OC system (i.e., embedded planner) to generate controls leads to better adaption to unseen situations. The  key problem in these  methods is  to learn a
planner (i.e., OC system), which is similar to our formulation.  \cite{okada2017path,pereira2018mpc}  learn a path-integral OC system \cite{kappen2005path}, which is  a special class of OC systems.   \cite{srinivas2018universal} learns an OC system in a latent space. However, all these methods adopt the `unrolling' strategy to facilitate differentiation. Specifically, they treat the forward pass of solving an OC problem as an `unrolled' computational graph of multiple steps of applying gradient descent, because by this  computational graph,  automatic differentiation  tool \cite{abadi2016tensorflow} can be  immediately applied.  The drawbacks of this `unrolling' strategy are apparent:  (i) they need to store all intermediate results over the entire computational graph, thus are memory-expensive; and (ii)
 the accuracy of  gradient depends on   the length of the `unrolled' graph, thus  facing trade-off between  complexity and accuracy.
To address these, \cite{amos2018differentiable}  develops a  differentiable MPC framework, where in  forward pass, a LQR approximation of the OC system  is  obtained, and  in  backward pass, the gradient is solved by differentiating such LQR approximation.  Although promising, this framework   has one main  weakness: differentiating  LQR requires to solve a large linear equation, which involves  the inverse of a matrix of  size ${\scriptsize{(2n{+}m)T\times(2n{+}m)T}}$, thus can incur huge  cost when  handling  systems of longer   horizons $T$. Detailed  descriptions for all these methods is  in  Appendix \ref{comparewithotherend2endframeworks}.

\begin{wrapfigure}[11]{r}{0pt}
	\raisebox{0pt}[\dimexpr\height-1.2\baselineskip]{\includegraphics[width=6cm, height=2.8cm]{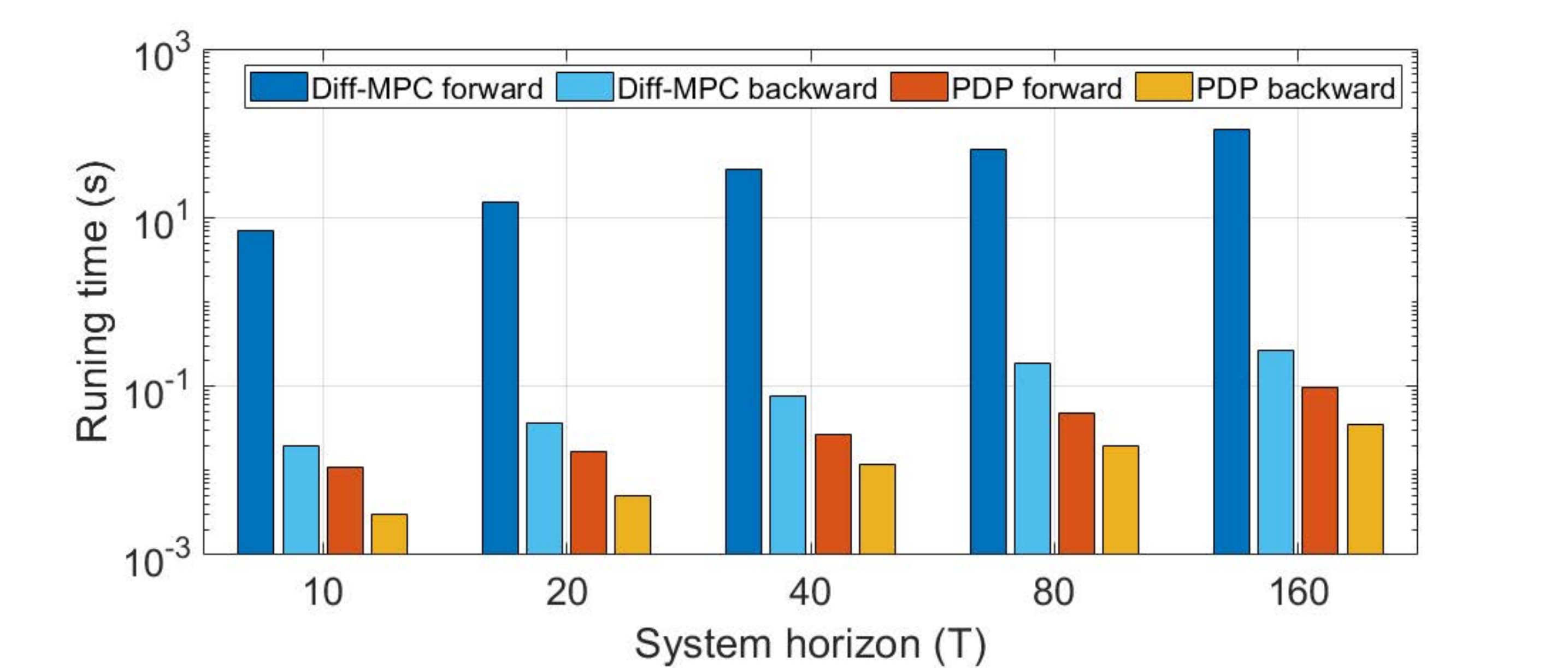}}
	\caption{Runtime (per iteration) comparison between  PDP and differentiable MPC for varying horizons of a pendulum system.}
	\label{section_comare}
\end{wrapfigure}

Compared to \cite{tamar2016value,okada2017path,pereira2018mpc,amos2018differentiable,srinivas2018universal}, the efficiency of PDP stems from the following novel aspects. First, in  forward pass,  without needing an unrolled computational graph, PDP only computes and stores the resulting trajectory of the OC system, $\boldsymbol{{\xi}}_{\boldsymbol{{\theta}}}$, (does not care about
how  $\boldsymbol{{\xi}}_{\boldsymbol{{\theta}}}$ is solved). Second, without obtaining intermediate (LQR) approximations, PDP  differentiates through  PMP of the OC system  to directly obtain  the {exact analytical gradient}. Third, in the backward pass, unlike differentiable MPC which  costs  at least  a complexity of $\mathcal{O}\left((m{+}2n)^2T^2\right)$ to differentiate a LQR approximation,   PDP  explicitly solves   $\frac{\partial  \boldsymbol{{\xi}}_{\boldsymbol{{\theta}}}}{\partial \boldsymbol{{\theta}}}$ by an auxiliary control system, where thanks to the  recursion structure, the memory and comptuation complexity of  PDP is only  $\mathcal{O}\left((m{+}2n)T\right)$.  In Fig. \ref{section_comare}, we have compared  the running time of PDP with that of differentiable MPC. The results show PDP is 1000x faster than differentiable MPC. Due to space constraint, we put the detailed  complexity analysis of PDP in Appendix \ref{pdpcomplexity}.

\textbf{Convergence and limitation of PDP.} Since all  gradient quantities in PDP are analytical and exact, and the   development of PDP does not involves any second-order derivative of functions or models,  PDP essentially is a \emph{first-order gradient-descent framework to solve non-convex bi-level optimization}. Therefore, in general, \emph{PDP can only achieve local minima}. As explored by  \cite{ghadimi2018approximation}, if we pose  further assumptions  
such as convexity and smoothness on all functions (dynamics,  policy, loss, and control objective function),  the global convergence of the bi-level programming could be established. But we do think these conditions are too restrictive for  dynamical control systems.  As a direction of  future work, we will investigate the mild  conditions for good convergence  by taking advantage of  control theory, e.g., Lyapunov theory.  Due to space constraint,  limitation of  PDP  is detailed in Appendix \ref{pdplimitation}.

\vspace{-2mm}

\section{Conclusions}
\vspace{-2mm}

This paper proposes a Pontryagin differentiable programming (PDP)  methodology to establish an end-to-end learning framework for solving a range of learning and control tasks. The key contribution in  PDP is that we incorporate the knowledge of optimal control theory as an inductive bias into the learning  framework. Such combination enables PDP to achieve higher efficiency and capability than existing learning and control methods in solving many tasks including  inverse reinforcement learning,  system identification, and  control/planning. We envision  the proposed PDP could benefit to  both  learning and control fields for solving many high-dimensional continuous-space problems.

\section*{Broader Impact}
This work is expected to have the impacts on both learning and control fields. 

\begin{itemize}
	\item To the learning field, this work connects some fundamental  topics in machine learning  to their counterparts in the control field, and unifies some concepts from  reinforcement learning, backpropagation/deep learning, and  control theory in one generic learning framework. The contribution of this framework is a deep integration of optimal control theory into  end-to-end learning process, leading to an optimal-control-informed end-to-end learning framework that is  flexible  enough to solve a broad range of  learning and control tasks and efficient enough to handle  high-dimensional and continuous-space problems. In a broad perspective, we hope that this paper could motivate more future work that integrates the benefits of both control  and learning to promote  efficiency and explainability of artificial intelligence.

	\item To the control field, this work proposes a generic paradigm, which shows how a challenging control task can be converted  into a learning formulation and  solved using readily-available learning techniques, such as (deep) neural networks and backpropagation. For example, the proposed framework, equipped with (deep) neural networks, shows significant advantage for handling non-linear system identification and optimal control over state-of-the-art control methods. Since classic control theory typically requires knowledge of models,  we expect that this work could pave a new way to extend classic control with data-driven techniques.

\end{itemize}

Since the formulation of this paper does not consider the boundness or constraints of a decision-making system, the real-world use of this work on physical systems might possibly  raise safety issues during the training process; e.g., the state or input of the physical system at some time instance might exceeds the  safety bounds that are physically required. One option to address this is to include these safety boundness as soft constraints added to the control objective or loss that is optimized. In future work, we will formally discuss  PDP within a safety framework.

\begin{ack}
We acknowledge support for this research from Northrop Grumman Mission Systems’ University Research Program.
\end{ack}

\bibliographystyle{unsrt}
\bibliography{diffcontrol} 

\newpage
\appendix
Supplementary materials for the Pontryagin Differentiable Programming paper

\setcounter{equation}{0}
\setcounter{figure}{0}   
\setcounter{table}{0}
\renewcommand\theequation{S.\arabic{equation}}
\renewcommand{\thefigure}{S\arabic{figure}}
\renewcommand{\thetable}{S\arabic{table}}

\setcounter{page}{1}

\section{Proof of Lemma \ref{theorem0} \label{appendix1}}
To prove Lemma \ref{theorem0}, we just need to show that the Pontryagin's Maximum Principle for the auxiliary control system $\boldsymbol{\overline\Sigma}(\boldsymbol{{\xi}}_{\boldsymbol{\theta}})$ in (\ref{backoc}) is exactly the differential PMP  in (\ref{diffpontryagin}). To this end, we define the following Hamiltonian for the auxiliary control system $\boldsymbol{\overline\Sigma}(\boldsymbol{{\xi}}_{\boldsymbol{\theta}})$:
\begin{equation}\label{auxHamil}
\bar{H}_t=\Tr\Bigg(\frac{1}{2}\small\begin{bmatrix}
{{X}}_{t}\\[5pt]
{{U}}_{t}
\end{bmatrix}^\prime\begin{bmatrix}
H_t^{xx} & H_t^{xu}\\[5pt]
H_t^{ux}& H_t^{uu}
\end{bmatrix}\begin{bmatrix}
{{X}}_{t}\\[5pt]
{{U}}_{t}
\end{bmatrix}+\begin{bmatrix}
H_t^{xe}\\[5pt]
H_t^{ue}
\end{bmatrix}^\prime\begin{bmatrix}
{{X}}_{t}\\[5pt]
{{U}}_{t}
\end{bmatrix}\Bigg)+\Tr\big({\Lambda}_{t+1}^\prime(F_t{{X}}_{t}+G_t{{U}}_{t}+E_t)\big),
\end{equation}
with $t=0,1,\cdots,T-1$. Here ${\Lambda}_{t+1}\in\mathbb{R}^{n\times r}$ denotes the costate (matrix) variables for the auxiliary control system. Based on Section 3 in \cite{athans1967matrix}, there exists a sequence of costates ${\Lambda}_{1:T}^{\boldsymbol{{\theta}}}$, which together 
the stationary solution $\{{{X}}_{0:T}^{\boldsymbol{\theta}},{{U}}_{0:T-1}^{\boldsymbol{\theta}}\}$ to the auxiliary control system  must satisfy the following the matrix version of PMP (we here follow the notation style used in (\ref{mp})).

\begin{subequations}\label{mpmp}
The dynamics equation:
	\begin{align}
	\frac{\partial \bar{H}_t}{\partial {\Lambda}_{t+1}^{\boldsymbol{\theta}}}&=\frac{\partial \Tr\big({\Lambda}_{t+1}^\prime(F_t{{X}}_{t}+G_t{{U}}_{t}+E_t)\big)}{\partial {\Lambda}_{t+1}}\bigg\rvert{\substack{{\Lambda}_{t+1}={\Lambda}_{t+1}^{\boldsymbol{\theta}}\\
			{X}_{t}={X}_{t}^{\boldsymbol{\theta}}\\{U}_{t}={U}_{t}^{\boldsymbol{\theta}} } }\nonumber\\
	&=F_t{{X}}_{t}^{\boldsymbol{\theta}}+G_t{{U}}_{t}^{\boldsymbol{\theta}}+E_t=\boldsymbol{0} \label{mpmp.1}.
	\end{align}
The costate equation:
	\begin{align}
	\frac{\partial \bar{H}_t}{\partial {X}_{t}^{\boldsymbol{\theta}}}&= \frac{\partial \Tr{(\frac{1}{2}{X}_t^\prime H_{t}^{{xx}}{X}_t)}
		+\partial \Tr{({U}_t^\prime H_{t}^{{ux}}{X}_t)}+\partial \Tr{ (H_{t}^{{ex}}{X}_t)} +\partial\Tr{ ({\Lambda}_{t+1}^\prime F_t{X}_t)}}{\partial {X}_t}
	\bigg\rvert{\substack{{\Lambda}_{t+1}={\Lambda}_{t+1}^{\boldsymbol{\theta}}\\
			{X}_{t}={X}_{t}^{\boldsymbol{\theta}}\\{U}_{t}={U}_{t}^{\boldsymbol{\theta}} } }\nonumber\\
	&=H_t^{xx}{X}_t^{\boldsymbol{\theta}}+H_t^{xu}{U}_t^{\boldsymbol{\theta}}+H_t^{xe}+F_t^{\prime}{\Lambda}_{t+1}^{\boldsymbol{\theta}}={\Lambda}_{t}^{\boldsymbol{\theta}}\label{mpmp.2}.
	\end{align}
Input equation:
	\begin{align}
	\frac{\partial \bar{H}_t}{\partial {U}_{t}^{\boldsymbol{\theta}}}&=\frac{\partial \Tr{(\frac{1}{2}{U}_t^\prime H_{t}^{{uu}}{U}_t)}
		+\partial \Tr{({U}_t^\prime H_{t}^{{ux}}{X}_t)}+\partial \Tr{ (H_{t}^{{eu}}{U}_t)} +\partial \Tr{({\Lambda}_{t+1}^\prime G_t{U}_t)}}{\partial {U}_t}\bigg\rvert{\substack{{\Lambda}_{t+1}={\Lambda}_{t+1}^{\boldsymbol{\theta}}\\
			{X}_{t}={X}_{t}^{\boldsymbol{\theta}}\\{U}_{t}={U}_{t}^{\boldsymbol{\theta}} } }\nonumber\\
	&=H_t^{uu}{U}_t^{\boldsymbol{\theta}}+H_t^{ux}{X}_t^{\boldsymbol{\theta}}+H_t^{ue}+G_t^{\prime}{\Lambda}_{t+1}^{\boldsymbol{\theta}}=\boldsymbol{0}\label{mpmp.3}.
	\end{align}
And boundary conditions:
\begin{equation}
{\Lambda}_T^{\boldsymbol{\theta}}=\frac{\partial \Tr(\frac{1}{2}{X}_T^\prime H^{xx}_T{X}_T)+\partial\Tr((H^{xe}_T)^\prime{X}_T)}{\partial {X}_T}\bigg\rvert_{
	{X}_{T}={X}_{T}^{\boldsymbol{\theta}} }=H^{xx}_T{X}_T^{\boldsymbol{\theta}}+H^{xe}_T,\label{mpmp.4}
\end{equation}
\end{subequations}
and $X_0^{\boldsymbol{\theta}}=\boldsymbol{0}$. Note that in the above derivations, we used the following matrix calculus \cite{athans1967matrix}:
\begin{equation}
\frac{\partial \Tr(AB)}{\partial A}= B^\prime, \quad \frac{\partial f(A)}{\partial A^\prime}= \left[\frac{\partial f(A)}{\partial A}\right]^\prime, \quad \frac{\partial \Tr(X^\prime HX)}{\partial X}=HX+H^\prime X,
\end{equation}
and the following matrix trace properties:
\begin{equation}
\Tr(A)=\Tr(A^\prime),\quad \Tr(ABC)=\Tr(BCA)=\Tr(CAB),\quad \Tr(A+B)=\Tr(A)+\Tr(B).
\end{equation}
Since the above obtained PMP equations  (\ref{mpmp}) are the same with the differential PMP in (\ref{diffpontryagin}), we thus can conclude that the Pontryagin's Maximum Principle of the  auxiliary control system $\boldsymbol{\overline\Sigma}(\boldsymbol{{\xi}}_{\boldsymbol{\theta}})$ in (\ref{backoc}) is exactly  the differential PMP equations (\ref{diffpontryagin}), and thus (\ref{uknown2}) holds. This completes the proof. \qed

\section{Proof of Lemma \ref{theorem1} \label{appendix2}}
Based on  Lemma \ref{theorem0} and its proof, we known that the PMP of the auxiliary control system, (\ref{mpmp}), is exactly the differential PMP equations (\ref{diffpontryagin}). Thus below, we only look at the differential PMP equations in (\ref{mpmp}).
From (\ref{mpmp.3}), we solve for ${{U}}^{\boldsymbol{\theta}}_{t}$ (if $H_t^{uu}$ invertible):
\begin{equation}
{{U}}^{\boldsymbol{\theta}}_{t}=-(H_t^{uu})^{-1}\Big(H_t^{ux}{{X}}^{\boldsymbol{\theta}}_{t}+G_t^\prime{{\Lambda}}_{t+1}^{\boldsymbol{\theta}}+H_t^{ue}\Big). \label{du}
\end{equation}
By substituting (\ref{du}) into (\ref{mpmp.1}) and (\ref{mpmp.2}), respectively, and  considering the definitions of matrices $A_t, R_t, M_t, Q_t$ and $N_t$ in (\ref{ricc}), we have
\begin{align}
{ X}^{\boldsymbol{\theta}}_{t+1}&={{A}}_t{ X}^{\boldsymbol{\theta}}_{t}-{{R}}_t{ \Lambda}^{\boldsymbol{\theta}}_{t+1}+{{M}}_t,\label{relation1}\\
{ \Lambda}^{\boldsymbol{\theta}}_{t}&={{Q}}_t{ X}^{\boldsymbol{\theta}}_{t}+{{A}}_t^\prime{ \Lambda}^{\boldsymbol{\theta}}_{t+1}+{{N}}_t, \label{relation2}
\end{align}
for $t=0, 1, \dots, T-1$, and also the boundary condition in (\ref{mpmp.4})
\begin{equation}
{ \Lambda}^{\boldsymbol{\theta}}_{T}=H_T^{xx}{ X}^{\boldsymbol{\theta}}_{T}+H_T^{xe}, \nonumber
\end{equation}
for $t=T$. Next, we  prove that there exist matrices ${{P}}_{t}$ and ${{W}}_{t}$ such that 
\begin{equation}
{ \Lambda}^{\boldsymbol{\theta}}_{t}={{P}}_{t}{ X}^{\boldsymbol{\theta}}_{t}+{{W}}_{t}. \label{lambdaX}
\end{equation}
Proof by induction: (\ref{mpmp.4}) shows that (\ref{lambdaX}) holds for $t=T$ if ${{P}}_T=H_T^{xx}$ and ${{W}}_T=H_T^{xe}$. Assume (\ref{lambdaX}) holds for $t+1$, then by manipulating (\ref{relation1}) and (\ref{relation2}), we have
\begin{equation}
{ \Lambda}^{\boldsymbol{\theta}}_{t}=\underbrace{\big({{Q}}_t+{{A}}_t^\prime({I}+{{P}}_{t\text{+}1}{{R}}_t)^{-1}{{P}}_{t\text{+}1}{{A}}_t\big)}_{{{P}}_t}{ X}^{\boldsymbol{\theta}}_{t}+\underbrace{{{A}}_t^\prime({I}+{{P}}_{t+1}{{R}}_t)^{-1}({{W}}_{t\text{+}1}{+{P}}_{t\text{+}1}{{M}}_{t})+{{N}}_t}_{{{W}_t}},
\end{equation}
which indicates  (\ref{lambdaX}) holds for $t$, if ${{P}}_{t}$ and ${{W}}_{t}$ satisfy (\ref{ricc.1}) and (\ref{ricc.2}), respectively. Substituting (\ref{lambdaX}) to (\ref{relation2}) and also considering (\ref{du}) will lead to (\ref{iter.2}). (\ref{iter.3}) directly results from (\ref{mpmp.1}). We complete the proof.\qed

\section{Proof of the Discrete-Time Pontryagin's Maximum Principle}\label{appendix3}
We here provide an easy-approach derivation of the discrete-time PMP based on
Karush-Kuhn-Tucker (KKT) conditions in non-linear optimization \cite{avriel2003nonlinear}. The original derivation for continuous optimal control systems uses the calculus of variation theory, which can be found in \cite{pontryagin1962mathematical} and \cite{liberzon2011calculus}.

We view the optimal control system (\ref{oc}) with a fixed  $\boldsymbol{\theta}$ as a constrained optimization problem, where the objective function is given by ${J}(\boldsymbol\theta)$ and the constraints given by  dynamics $\boldsymbol{f}(\boldsymbol{\theta})$. Define the following Lagrangian for this constrained optimization problem:
\begin{equation}\label{lagarange}
\begin{aligned}
L&=J(\boldsymbol{\theta})+\sum\nolimits_{t=0}^{T-1}\boldsymbol{\lambda}_{t+1}^\prime\big(\boldsymbol{f}(\boldsymbol{x}_t,\boldsymbol{u}_t,\boldsymbol{\theta})-\boldsymbol{x}_{t+1}\big) \\
&=\sum\nolimits_{t=0}^{T{-}1}\bigg(c_t(\boldsymbol{x}_t,\boldsymbol{u}_t, {\boldsymbol{\theta}})+\boldsymbol{\lambda}_{t+1}^\prime\big(\boldsymbol{f}(\boldsymbol{x}_t,\boldsymbol{u}_t,\boldsymbol{\theta})-\boldsymbol{x}_{t+1}\big)\bigg)+h(\boldsymbol{x}_T,{\boldsymbol{\theta}}) \\
&=\sum\nolimits_{t=0}^{T{-}1}\bigg(H_t-\boldsymbol{\lambda}_{t+1}^\prime\boldsymbol{x}_{t+1}\bigg)+h(\boldsymbol{x}_T,{\boldsymbol{\theta}}),
\end{aligned}
\end{equation}
where $\boldsymbol{\lambda}_t$ is the Lagrange multiplier for the dynamics constraint for $t=1,2,\cdots,T$, and the third line in (\ref{lagarange}) is due to the definition of Hamiltonian in (\ref{Hamil}). According to the KKT conditions, for the optimal solution $\boldsymbol{\xi}_{\boldsymbol{\theta}}=\{\boldsymbol{{x}}_{0:T}^{\boldsymbol{\theta}},\boldsymbol{{u}}_{0:T-1}^{\boldsymbol{\theta}}\}$, there must exist the multiplers $\boldsymbol{\lambda}_{1:T}^{\boldsymbol{\theta}}$ (in optimal control they are called costates) such that the following first-order conditions are satisfied:
\begin{align}\label{KKT}
\frac{\partial L}{\partial \boldsymbol{\lambda}_{1:T}^{\boldsymbol{{\theta}}}}=\boldsymbol{0}, \quad
\frac{\partial L}{\partial \boldsymbol{x}_{0:T}^{\boldsymbol{{\theta}}}}=\boldsymbol{0}, \quad
\frac{\partial L}{\partial \boldsymbol{u}_{0:T\text{-}1}^{\boldsymbol{{\theta}}}}=\boldsymbol{0}.
\end{align}
By extending the above three conditions in (\ref{KKT}) at each  $\boldsymbol{\lambda}_t$, $\boldsymbol{x}_t$ and $\boldsymbol{u}_t$, respectively, and particularly taking care of $\boldsymbol{x}_T$,  we will obtain

\begin{subequations}\label{appendixmp}
	\begin{align}
	\boldsymbol{0}&=\boldsymbol{f}(\boldsymbol{{x}}^{\boldsymbol{\theta}}_t,\boldsymbol{{u}}^{\boldsymbol{\theta}}_t;\boldsymbol{\theta})-\boldsymbol{{x}}^{\boldsymbol{\theta}}_{t\text{+}1}, \label{appendixmp.1}\\[0pt]
	\boldsymbol{0}&=\dfrac{\partial H_t}{\partial \boldsymbol{{x}}^{\boldsymbol{\theta}}_{t}}-\boldsymbol{\lambda}^{\boldsymbol{\theta}}_{t}=\dfrac{\partial c_t}{\partial \boldsymbol{{x}}^{\boldsymbol{\theta}}_{t}}+\dfrac{\partial \boldsymbol{f}^\prime}{\partial \boldsymbol{{x}}^{\boldsymbol{\theta}}_{t}}\boldsymbol{{\lambda}}^{\boldsymbol{\theta}}_{t\text{+}1}-\boldsymbol{\lambda}^{\boldsymbol{\theta}}_{t},\label{appendixmp.2}\\[0pt]
	\boldsymbol{0}&=\dfrac{\partial H_t}{\partial \boldsymbol{{u}}^{\boldsymbol{\theta}}_{t}}=\dfrac{\partial c_t}{\partial \boldsymbol{{u}}^{\boldsymbol{\theta}}_{t}}+\dfrac{\partial \boldsymbol{f}^\prime}{\partial \boldsymbol{{u}}^{\boldsymbol{\theta}}_{t}}\boldsymbol{{\lambda}}^{\boldsymbol{\theta}}_{t\text{+}1},\label{appendixmp.3}
	\\[0pt]
	\boldsymbol{0}&=\frac{\partial h}{\partial\boldsymbol{{x}}^{\boldsymbol{\theta}}_T}-\boldsymbol{{\lambda}}^{\boldsymbol{\theta}}_{T} \label{appendixmp.4},
	\end{align}
\end{subequations}
respectively, which are exactly the PMP equations in (\ref{mp}). This completes the proof. \qed

\section{Algorithms Details for Different Learning Modes}\label{appendixalgorithem}

\begin{algorithm2e}[H]
	\SetAlgoNoLine
	\SetKwComment{Comment}{$\triangleright$\ }{}

	\SetKwInput{given}{Input}
	\given{The  trajectory $\boldsymbol{{\xi}}_{\boldsymbol{\theta}}$ generated by the  system $\boldsymbol{\Sigma}(\boldsymbol{\theta})$}
	
	\Indp\Indp 
	
	Compute the coefficient matrices (\ref{mats}) to obtain the auxiliary control system  $\boldsymbol{\overline\Sigma}(\boldsymbol{{\xi}}_{\boldsymbol{\theta}})$ in (\ref{backoc})\;
	
	\textbf{def} Auxiliary\_Control\_System\_Solver ( $\boldsymbol{\overline\Sigma}(\boldsymbol{{\xi}}_{\boldsymbol{\theta}})$ ): \Comment*[f]{implementation of Lemma \ref{theorem1}}

	\SetKw{KwBy}{by}
	\Indp
\smallskip
	 Set ${{P}}_T=H_T^{xx}$ and ${{W}}_T=H_T^{xe}$\;
	 \For{$t\gets T$ \KwTo $0$ \KwBy $-1$ }{
	 	Update $P_t$ and $W_t$ using equations (\ref{ricc}); \Comment*[f]{backward in time} 
	 }
 \smallskip
 	Set ${X}_0^{\boldsymbol{\theta}}=\boldsymbol{0}$\;
 	\For{$t\gets 0$ \KwTo $T$ \KwBy $1$ }{
 		Update $X_t^{\boldsymbol{\theta}}$ and $U_t^{\boldsymbol{\theta}}$ using equations (\ref{iter}); \Comment*[f]{forward in time} 
 	}
 \smallskip
	\textbf{Return:} $\{{{X}}_{0:T}^{\boldsymbol{\theta}},{{U}}_{0:T-1}^{\boldsymbol{\theta}}\}$

	\Indm\Indm\Indm
	\SetKwInput{return}{Return}
	\return{$\frac{\partial  \boldsymbol{{\xi}}_{\boldsymbol{{\theta}}}}{\partial \boldsymbol{{\theta}}}=\{{{X}}_{0:T}^{\boldsymbol{\theta}},{{U}}_{0:T-1}^{\boldsymbol{\theta}}\}$}
	
	\caption{Solving $\frac{\partial  \boldsymbol{{\xi}}_{\boldsymbol{{\theta}}}}{\partial \boldsymbol{{\theta}}}$ using Auxiliary Control System}\label{algsolvauxsys}
\end{algorithm2e}

\begin{algorithm2e}
	\SetKwComment{Comment}{$\triangleright$\ }{}

	\SetKwInOut{Input}{Data}
	\Input{Expert demonstrations $\{\boldsymbol{{\xi}}^{\text{d}}\}$}
	
	\SetKwInput{Parameterization}{Parameterization}
	\Parameterization{The parameterized optimal control system $\boldsymbol{\Sigma}(\boldsymbol{\theta})$ in (\ref{oc})}

	\SetKwInput{loss}{Loss}
	\loss{$L(\boldsymbol{\xi}_{\boldsymbol{\theta}},\boldsymbol{\theta})$ in (\ref{lossioc})}

	\SetKwInOut{Initial}{Initialization}
	\Initial{$\boldsymbol{\theta}_{0}$, learning rate $\{\eta_{k}\}_{k=0,1,\cdots}$}
	
	\For{$k=0,1,2,\cdots$}{
		\smallskip
		Solve $\boldsymbol{{\xi}}_{\boldsymbol{\theta}_k}$ from the current optiaml control system $\boldsymbol{\Sigma}(\boldsymbol{\theta}_k)$ \Comment*[r]{using any OC solver}
		\smallskip

		Obtain $\frac{\partial  \boldsymbol{{\xi}}_{\boldsymbol{\theta}}}{\partial \boldsymbol{{\theta}}}\bigr|_{\boldsymbol{\theta}_k}$ using Algorithm \ref{algsolvauxsys} given  $\boldsymbol{{\xi}}_{\boldsymbol{\theta}_k}$ \Comment*[r]{using Algorithm \ref{algsolvauxsys}}
		
		\smallskip
		
		Obtain $\frac{\partial L}{\partial \boldsymbol{{\xi}}}\bigr|_{\boldsymbol{{\xi}}_{\boldsymbol{\theta}_k}}$ from the given loss function 	$L(\boldsymbol{\xi}_{\boldsymbol{\theta}},\boldsymbol{\theta})$
		\;
		Apply the chain rule (\ref{GD}) to obtain $\frac{d L}{d \boldsymbol{\theta}}\bigr|_{\boldsymbol{\theta}_{k}}$ \;
		\smallskip
		Update $\boldsymbol{\theta}_{k+1}\leftarrow\boldsymbol{\theta}_{k}-\eta_{k}\frac{d L}{d \boldsymbol{\theta}}\bigr|_{\boldsymbol{\theta}_{k}}$\;
	}		
	\caption{PDP Algorithm for IRL/IOC Mode}
	\label{agpdpiocmode}
\end{algorithm2e}

\begin{algorithm2e}
	
	\SetKwInput{Input}{Data}
	\Input{Input-state data $\{\boldsymbol{\xi}^{\text{o}}\}$}

	\SetKwInput{parameterization}{Parameterization}
	\parameterization{The parameterized dynamics model $\boldsymbol{\Sigma}(\boldsymbol{\theta})$ in (\ref{ocmodeid})}

	\SetKwInput{loss}{Loss}
	\loss{$L(\boldsymbol{\xi}_{\boldsymbol{\theta}},\boldsymbol{\theta})$ in (\ref{lossid})}

	\SetKwInput{Initial}{Initialization}
	\Initial{$\boldsymbol{\theta}_{0}$, learning rate $\{\eta_{k}\}_{k=0,1,\cdots}$}

	\For{$k=0,1,2,\cdots$}{
		\smallskip
		Obtain $\boldsymbol{{\xi}}_{\boldsymbol{\theta}_k}$ by iteratively integrating $\boldsymbol{\Sigma}(\boldsymbol{\theta}_k)$ in (\ref{ocmodeid})  for $t=0,..., T-1$\;
		\smallskip

		Compute the coefficient matrices (\ref{mats}) to obtain the auxiliary control system  $\boldsymbol{\overline\Sigma}(\boldsymbol{{\xi}}_{\boldsymbol{\theta}})$ in (\ref{iddiff})\;
		\smallskip

		Obtain  $\frac{\partial  \boldsymbol{{\xi}}_{\boldsymbol{\theta}}}{\partial \boldsymbol{{\theta}}}\bigr|_{\boldsymbol{\theta}_k}$ by iteratively integrating $\boldsymbol{\overline\Sigma}(\boldsymbol{{\xi}}_{\boldsymbol{\theta}_k})$ in (\ref{iddiff}) for $t=0,..., T-1$\;
		\smallskip
		Obtain $\frac{\partial L}{\partial \boldsymbol{{\xi}}}\bigr|_{\boldsymbol{{\xi}}_{\boldsymbol{\theta}_k}}$ from the given loss function in (\ref{lossid})\;	
		Apply the chain rule (\ref{GD}) to obtain $\frac{d L}{d \boldsymbol{\theta}}\bigr|_{\boldsymbol{\theta}_{k}}$ \;
		\smallskip
		Update $\boldsymbol{\theta}_{k+1}\leftarrow\boldsymbol{\theta}_{k}-\eta_{k}\frac{d L}{d \boldsymbol{\theta}}\bigr|_{\boldsymbol{\theta}_{k}}$\;
	}	
	\caption{PDP Algorithm for SysID Mode}
	\label{agpdpid}
\end{algorithm2e}

\begin{algorithm2e}[H]\label{c}\label{agpdpplan}

	\SetKwInOut{Input}{Input}
	\SetKwInput{parameterization}{Parameterization}
	\parameterization{The parameterized-policy system $\boldsymbol{\Sigma}(\boldsymbol{\theta})$ in (\ref{ocmodeplan})}

	\SetKwInput{loss}{Loss}
	\loss{$L(\boldsymbol{\xi}_{\boldsymbol{\theta}},\boldsymbol{\theta})$ in (\ref{lossmodeplan})}

	\SetKwInput{Initial}{Initialization}
	\Initial{$\boldsymbol{\theta}_{0}$, learning rate $\{\eta_{k}\}_{k=0,1,\cdots}$}

	\For{$k=0,1,2,\cdots$}{
		Obtain $\boldsymbol{{\xi}}_{\boldsymbol{\theta}_k}$ by iteratively integrating  $\boldsymbol{\Sigma}(\boldsymbol{\theta}_k)$ in (\ref{ocmodeplan}) for $t=0,..., T-1$\;
		\smallskip
		
		Compute the coefficient matrices (\ref{mats}) to obtain the auxiliary control system  $\boldsymbol{\overline\Sigma}(\boldsymbol{{\xi}}_{\boldsymbol{\theta}})$ in (\ref{mode3controlback})\;
		\smallskip
		
		Obtain  $\frac{\partial  \boldsymbol{{\xi}}_{\boldsymbol{\theta}}}{\partial \boldsymbol{{\theta}}}\bigr|_{\boldsymbol{\theta}_k}$ by iteratively integrating  $\boldsymbol{\overline\Sigma}(\boldsymbol{{\xi}}_{\boldsymbol{\theta}_k})$ in (\ref{mode3controlback}) for $t=0,..., T-1$\;	
		\smallskip
		Obtain $\frac{\partial L}{\partial \boldsymbol{{\xi}}}\bigr|_{\boldsymbol{{\xi}}_{\boldsymbol{\theta}_k}}$ from the given loss function $L(\boldsymbol{\xi}_{\boldsymbol{\theta}},\boldsymbol{\theta})$ in (\ref{lossmodeplan})\;	
		Apply the chain rule (\ref{GD}) to obtain $\frac{d L}{d \boldsymbol{\theta}}\bigr|_{\boldsymbol{\theta}_{k}}$ \;
		\smallskip
		Update $\boldsymbol{\theta}_{k+1}\leftarrow\boldsymbol{\theta}_{k}-\eta_{k}\frac{d L}{d \boldsymbol{\theta}}\bigr|_{\boldsymbol{\theta}_{k}}$\;
	}	
	\caption{PDP Algorithm for  Control/Planning Mode}
	
\end{algorithm2e}

\vspace{3mm}
\textbf{Additional comments: combining different learning modes.}\quad
In addition to using different learning modes to solve  different types of problems, one can combine  different  modes  in a single learning task. For example, when solving model-based reinforcement learning, one can call {SysID Mode} to first learn  a dynamics model, then use the learned dynamics in {Control/Planning Mode} to obtain an optimal policy. In problems such as imitation learning, one can first learn a dynamics model using {SysID Mode}, then use the learned dynamics as the initial guess in {IRL/IOC Mode}. In  forward pass of {IOC/IRL Mode}, one  can call {Control/Planning Mode} to  solve  the OC system. For  control and planning problems, the loss required in {Control/Planning Mode} can be learned using {IOC/IRL Mode}. In   MPC-based learning and control  \citep{amos2018differentiable}, one can use the general formulation in (\ref{prob}) to learn a MPC controller, and then execute the   MPC controller by calling {Control/Planning Mode}.

\section{Experiment Details}\label{appendixexperiment}
We have released  the PDP source codes and different simulation environments/systems  in this paper as two standalone packages, both of which are available at \url{https://github.com/wanxinjin/Pontryagin-Differentiable-Programming}. The video demos for some of the experiments are available at \url{https://wanxinjin.github.io/posts/pdp}.

\subsection{System/Environment Setup}\label{appendix_experiment_setup}

\textbf{Quadrotor maneuvering control on \emph{SE}(3).} \quad
We consider a quadrotor system maneuvering on \emph{SE}(3) space (i.e. full position and full attitude space). The equation of motion of a quadrotor  is given by: 
\begin{equation}
\begin{aligned}
\dot{\boldsymbol{p}}_{I}&=\dot{\boldsymbol{v}}_{I},\\
m\dot{\boldsymbol{v}}_{I}&=m\boldsymbol{g}_{I}+\boldsymbol{f}_{I},\\
\dot{\boldsymbol{q}}_{B/I}&=\frac{1}{2}\Omega(\boldsymbol{\omega}_{B}){\boldsymbol{q}}_{B/I},\\
J_{B}\dot{\boldsymbol{\omega}}_{B}&=\boldsymbol{M}_{B}-\boldsymbol{\omega}\times J_{B}\boldsymbol{\omega}_{B}.
\end{aligned}
\end{equation}
Here, the subscriptions $_{B}$ and $_{I}$ denote that a quantity is expressed in the quadrotor's body frame and inertial (world) frame, respectively;  $m$ is the mass of the quadrotor, respectively; $\boldsymbol{p}\in\mathbb{R}^{3}$ and $\boldsymbol{v}\in\mathbb{R}^{3}$ are the position and velocity vector of the quadrotor; $J_{B}\in\mathbb{R}^{3\times3}$ is the moment of inertia of the quadrotor with respect to its body frame; $\boldsymbol{\omega}_{B}\in\mathbb{R}^{3}$ is the angular velocity of the quadrotor; $\boldsymbol{q}_{B/I}\in\mathbb{R}^{4}$ is the unit quaternion \cite{kuipers1999quaternions} describing the attitude of quadrotor  with respect to the inertial frame; $\Omega(\boldsymbol{\omega}_{B})$ is  
\begin{equation}
\Omega(\boldsymbol{\omega}_{B})=\begin{bmatrix}
0&-\omega_x &-\omega_y &-\omega_z\\
\omega_x&0&\omega_z &-\omega_y \\
\omega_y&-\omega_z&0 &\omega_x \\
\omega_z&\omega_y &-\omega_x &0
\end{bmatrix}
\end{equation}
and used for quaternion
multiplication;
$\boldsymbol{M}_{B}\in \mathbb{R}^{3}$ is the torque applied to the quadrotor; and $\boldsymbol{f}_{I}\in \mathbb{R}^{3}$ is the force vector applied to the quadrotor's center of mass (COM). The total force magnitude $\norm{\boldsymbol{f}_{I}}\in\mathbb{R}$ (along the z-axis of the body frame) and torque $\boldsymbol{M}_{B}=[M_x, M_y, M_z]$ are generated by thrusts $[T_1, T_2, T_3, T_4]$ of the four  rotating propellers of the quadrotor, which can be written as
\begin{equation}
\begin{bmatrix}
\norm{\boldsymbol{f}_{I}}\\
M_x\\
M_y\\
M_z
\end{bmatrix}=
\begin{bmatrix}
1&1 &1 &1\\
0&-l_w/2& 0 & l_w/2 \\
-l_w/2& 0 & l_w/2 & 0 \\
c&-c&c&-c
\end{bmatrix}
\begin{bmatrix}
T_1\\
T_2\\
T_3\\
T_4
\end{bmatrix},
\end{equation}
with $l_w$ being  the wing length of the quadrotor and $c$  a fixed constant.

We define the state  and input vectors of the quadrotor system  as
\begin{equation}
\boldsymbol{x}=\begin{bmatrix}
\boldsymbol{p}^\prime & \boldsymbol{v}^\prime  &\boldsymbol{q}^\prime  & \boldsymbol{\omega}^\prime 
\end{bmatrix}^\prime \in\mathbb{R}^{13} \quad
\text{and} \quad \boldsymbol{u}=\begin{bmatrix}
T_1 & T_2  &T_3 & T_4
\end{bmatrix}^\prime \in\mathbb{R}^{4}.
\end{equation}
respectively. In design of the quadrotor's control objective function, to achieve \emph{SE}(3) maneuvering control performance, we need to carefully design the attitude error.  As used in \cite{lee2010geometric}, we define the attitude error between the quadrotor's current attitude $\boldsymbol{q}$ and the goal attitude $\boldsymbol{q}_{\text{g}}$ as
\begin{equation}
e(\boldsymbol{q},\boldsymbol{q}_{\text{g}})=\frac{1}{2}\Tr (I-R^\prime(\boldsymbol{q}_{\text{g}})R(\boldsymbol{q})),
\end{equation}
where $R(\boldsymbol{q})\in\mathbb{R}^{3\times3}$ are the  direction cosine matrix directly corresponding to   $\boldsymbol{q}$ (see \cite{kuipers1999quaternions} for more details). Other error term in the control objective is the distance to the respective goal:
\begin{equation}
e(\boldsymbol{p},\boldsymbol{p}_{\text{g}})=\norm{\boldsymbol{p}-\boldsymbol{p}_{\text{g}}}^2, \quad e(\boldsymbol{v},\boldsymbol{v}_{\text{g}})=\norm{\boldsymbol{v}-\boldsymbol{v}_{\text{g}}}^2, \quad e(\boldsymbol{\omega},\boldsymbol{\omega}_{\text{g}})=\norm{\boldsymbol{\omega}-\boldsymbol{\omega}_{\text{g}}}^2.
\end{equation}

\smallskip

\textbf{Two-link robot arm.} \quad The dynamics of a two-link robot arm  can be found in \cite[p. 171]{spong2008robot}, where the  state vector is $\boldsymbol{x}=[\boldsymbol{q},\dot{\boldsymbol{q}}]^\prime$ with $\boldsymbol{q}\in\mathbb{R}^2$  the vector of joint angles and $\dot{\boldsymbol{q}}\in\mathbb{R}^2$  the vector of joint angular velocities, and the control input $\boldsymbol{u}\in\mathbb{R}^2$ is the vector of torques applied to each joint.

\smallskip

\textbf{Dynamics discretization.} \quad The continuous-time dynamics of all experimental systems in Table \ref{experimenttable} are discretized using the Euler method: $\boldsymbol{x}_{t+1}=\boldsymbol{x}_{t}+\Delta\cdot \boldsymbol{f}(\boldsymbol{x}_t,\boldsymbol{u}_t)$ with the discretization interval $\Delta=0.05$s or $\Delta=0.1$s.

\textbf{Simulation environment source codes.} We have made  different simulation environments/systems in  Table \ref{experimenttable} as a standalone Python package, which is available at \url{https://github.com/wanxinjin/Pontryagin-Differentiable-Programming}. This  environment package is easy to use and has user-friendly interfaces for customization.

\subsection{Experiment of Imitation Learning}\label{apendix-exp-imitation}

\textbf{Data acquisition.} \quad
The dataset of expert demonstrations  $\{\boldsymbol{\xi}^{\text{d}}\}$ is generated by solving an expert optimal control system with the expert's dynamics and control objective parameter $\boldsymbol{\theta}^*=\{\boldsymbol{\theta}^*_{\text{dyn}},\boldsymbol{\theta}^*_{\text{dyn}}\}$ given. We generate a  number of five trajectories, where different trajectories $\boldsymbol{\xi}^{\text{d}}=\{\boldsymbol{{x}}_{0:T}^{\text{d}},\boldsymbol{{u}}^{\text{d}}_{0:T-1}\}$  have different initial conditions $\boldsymbol{x}_0$ and time horizons $T$ ($T$ ranges from $40$ to $50$).

\textbf{Inverse KKT method.} \quad
We choose the inverse KKT method \cite{englert2017inverse} for comparison because it is suitable for learning  objective functions for high-dimensional continuous-space systems. We adapt the inverse KKT method, and define the KKT loss as the norm-2 violation of  the KKT condition (\ref{KKT}) by the demonstration data $\boldsymbol{\xi}^{\text{d}}$, that is,
\begin{equation}\label{inverseKKT}\small
\min_{\boldsymbol{\theta},\boldsymbol{{\lambda}}_{1:T}}\left(\left|\left|
\frac{\partial L}{\partial \boldsymbol{x}_{0:T}^{\boldsymbol{{}}}}(\boldsymbol{{x}}_{0:T}^{\text{d}},\boldsymbol{{u}}^{\text{d}}_{0:T-1})\right|\right|^2+\left|\left|
\frac{\partial L}{\partial \boldsymbol{u}_{0:T\text{-}1}^{\boldsymbol{{}}}}(\boldsymbol{{x}}_{0:T}^{\text{d}},\boldsymbol{{u}}^{\text{d}}_{0:T-1})\right|\right|^2\right),
\end{equation}
where $\frac{\partial L}{\partial \boldsymbol{x}_{0:T}^{\boldsymbol{{}}}}(\cdot)$ and $\frac{\partial L}{\partial \boldsymbol{u}_{0:T-1}^{\boldsymbol{{}}}}(\cdot)$ are defined in (\ref{KKT}) and $\boldsymbol{\theta}{=}\{\boldsymbol{\theta}_{\text{dyn}},\boldsymbol{\theta}_{\text{dyn}}\}$. We minimize the above KKT-loss  with respect to the unknown $\boldsymbol{\theta}$ and the costate variables $\boldsymbol{\lambda}_{1:T}$. 

Note that to illustrate the inverse-KKT learning results  in Fig. \ref{figioc}, we plot the imitation loss  $L(\boldsymbol{\xi}_{\boldsymbol{\theta}},\boldsymbol{\theta})= {\norm{\boldsymbol{{\xi}}^{\text{d}}-\boldsymbol{{\xi}}_{\boldsymbol{\theta}}}^2}$ instead of the  KKT loss (\ref{inverseKKT}), because we want to guarantee that  the comparison criterion is the same across different methods. Thus for each iteration ${k}$ in minimizing the KKT loss (\ref{inverseKKT}), we use the parameter $\boldsymbol{\theta}_k$ to compute the optimal trajectory $\boldsymbol{\xi}_{{{\boldsymbol{\theta}_k}}}$ and obtain the imitation loss.

\textbf{Neural policy cloning.} \quad
For the neural policy cloning (similar to \cite{bojarski2016end}), we directly learn a neural-network  policy $\boldsymbol{u}=\boldsymbol{\pi}_{\boldsymbol{\theta}}(\boldsymbol{x})$ from the dataset using supervised learning, that is
\begin{equation}
\min_{\boldsymbol{\theta}}\sum\nolimits_{t=0}^{T-1}\norm{\boldsymbol{u}_t^{\text{d}}-\boldsymbol{\pi}_{\boldsymbol{\theta}}(\boldsymbol{x}^{\text{d}}_t)}^2.
\end{equation}

\textbf{Learning neural control objective function.} \quad  
In Fig. \ref{figioc.4}, we apply PDP to learn a neural objective function of the robot arm. The neural objective function is constructed as 
\begin{equation}\label{appendix_neural_obj}
J(\boldsymbol{\theta})=V_{\boldsymbol{\theta}}(\boldsymbol{x})+0.0001\norm{\boldsymbol{u}}^2,
\end{equation}
with $V_{\boldsymbol{\theta}}(\boldsymbol{x})$ a fully-connected feed-forward network with \texttt{n-n-1} layers and $\tanh$ activation functions, i.e., an input layer with \texttt{n} neurons equal to the dimension of state, $n$, one hidden layer with \texttt{n} neurons and one output layer with \texttt{1} neuron. $\boldsymbol{\theta}$ is the neural network parameter. We separate the input cost  from the neural network because otherwise it will cause instability when  solving OC problems in the forward pass. Also, in learning the above neural objective function, we fix the dynamics because otherwise it will also lead to instability of solving OC.

In the comparing GAIL method \cite{ho2016generative}, we use the following hyper-parameters: the policy network is  a fully-connected feed-forward network with \texttt{n-400-300-m} layers and \texttt{relu} activation functions; the discriminator network is a \texttt{(n+m)-400-300-1} fully-connected feed-forward network with  \texttt{tanh} and \texttt{sigmoid} activation functions; and the policy regularizer $\lambda$ is set to zero.

\begin{figure}
	\begin{subfigure}{1\textwidth}
		\centering
		\includegraphics[width=\linewidth]{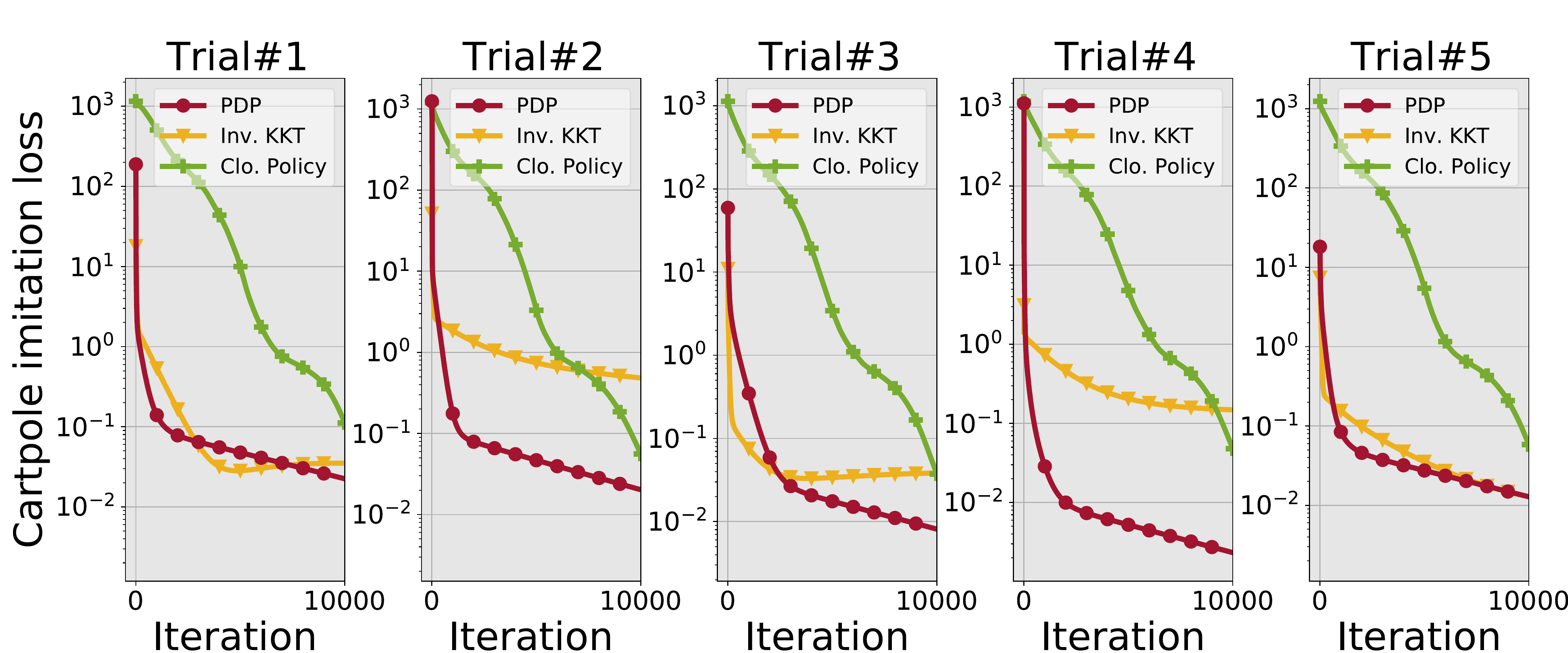}
		\label{appendix-ioc.1}
	\end{subfigure}
	\begin{subfigure}{1\textwidth}
		\centering
		\includegraphics[width=\linewidth]{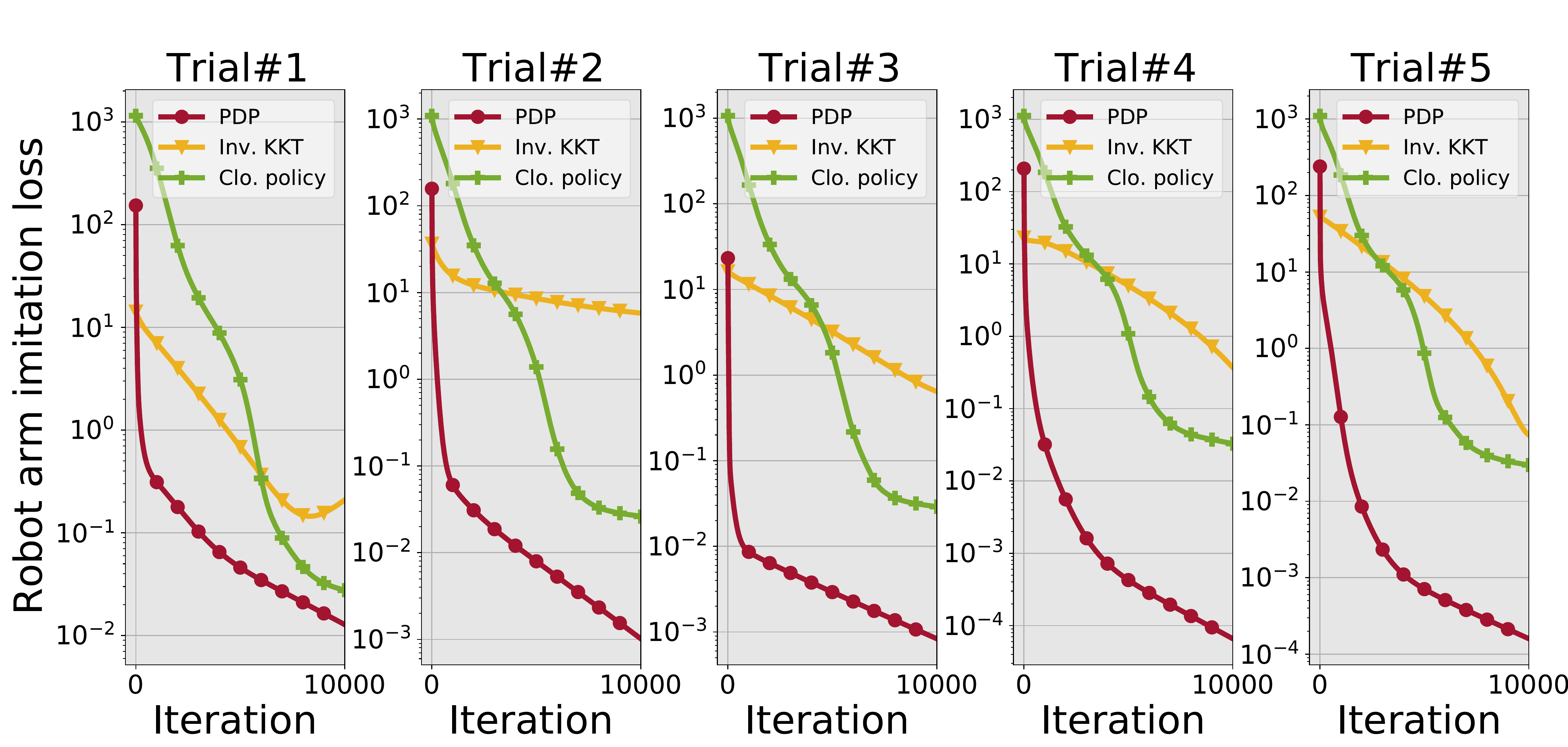}
		\label{appendix-ioc.2}
	\end{subfigure}
	\begin{subfigure}{1\textwidth}
		\centering
		\includegraphics[width=\linewidth]{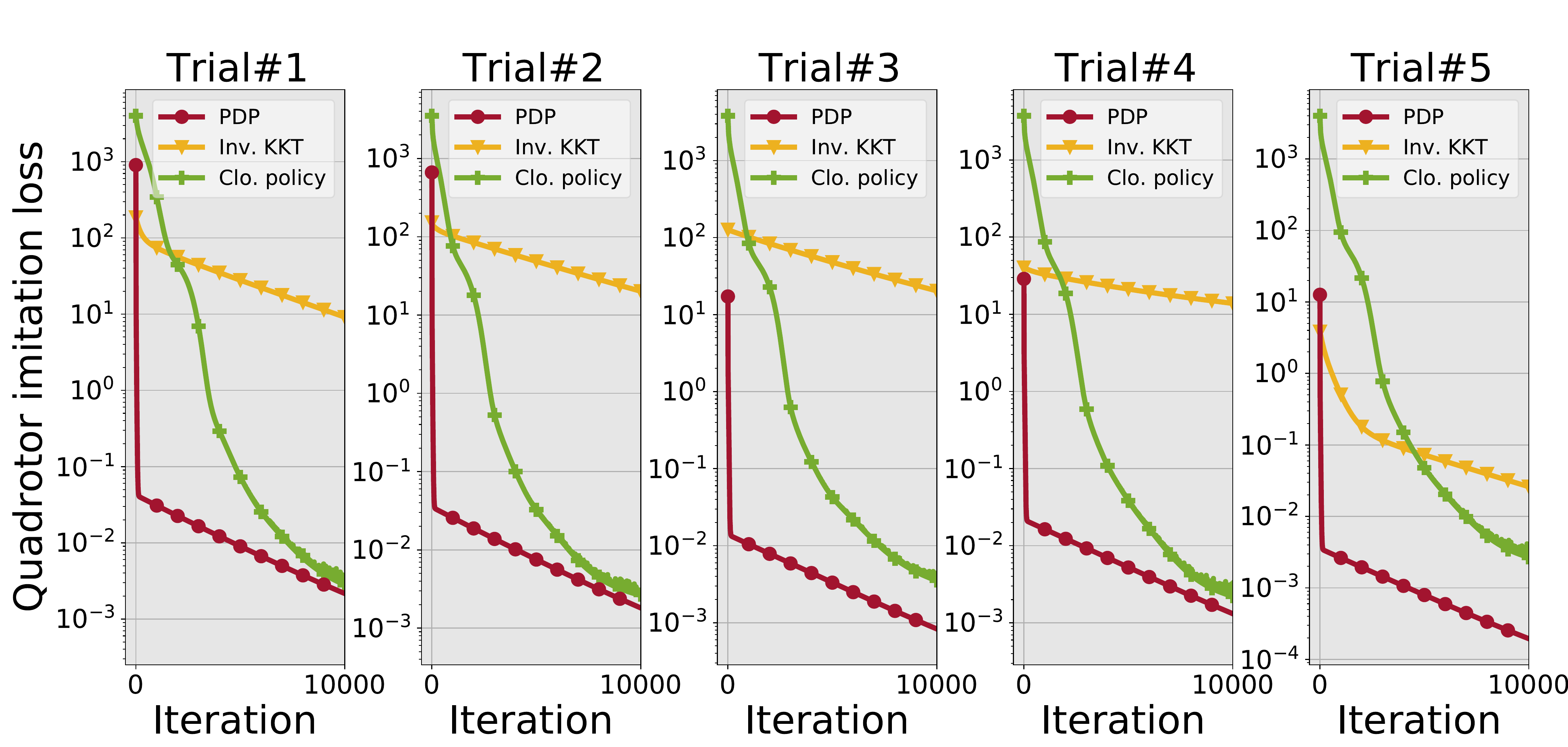}
		\label{appendix-ioc.3}
	\end{subfigure}
	\caption{Experiments for PDP IRL/IOC Mode: imitation loss versus iteration. For each system, we run five trials starting with random initial guess $\boldsymbol{\theta}_0$, and the learning rate is $\eta=10^{-4}$ for all methods. The results show a significant advantage of the PDP over the neural policy cloning and inverse-KKT \cite{englert2017inverse} in terms of lower training loss and faster convergence speed. Please see Appendix Fig. \ref{appendix-ioc-test} for validation. Please find the video demo at  \url{https://youtu.be/awVNiCIJCfs}.}
	\label{appendix-ioc}
\end{figure}

\begin{figure}[h]
	\centering
	\begin{subfigure}{.32\textwidth}
		\centering
		\includegraphics[width=\linewidth]{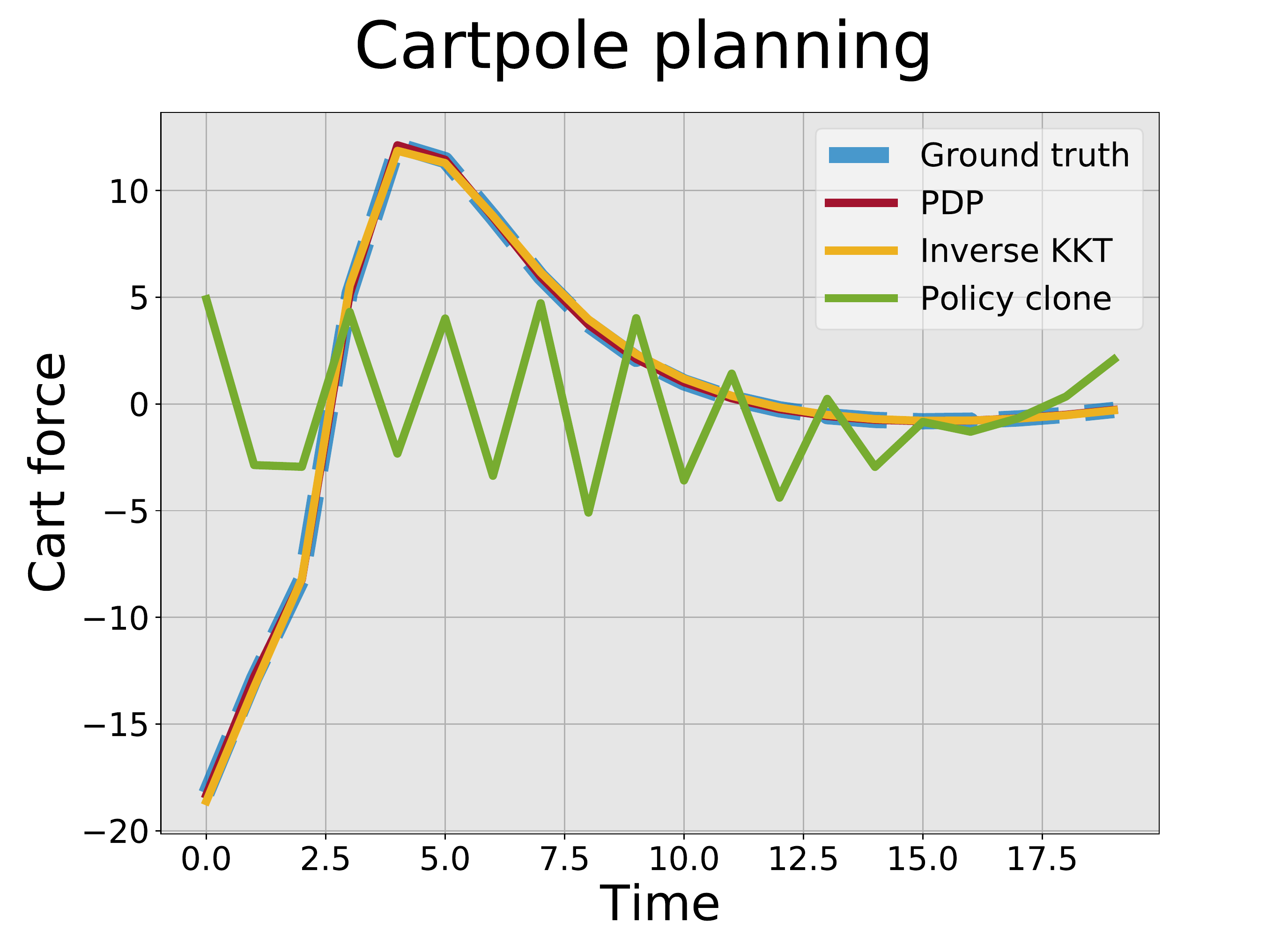}
		\caption{Cart-pole}
		\label{appendix-ioc-test.1}
	\end{subfigure}
	\begin{subfigure}{.32\textwidth}
		\centering
		\includegraphics[width=\linewidth]{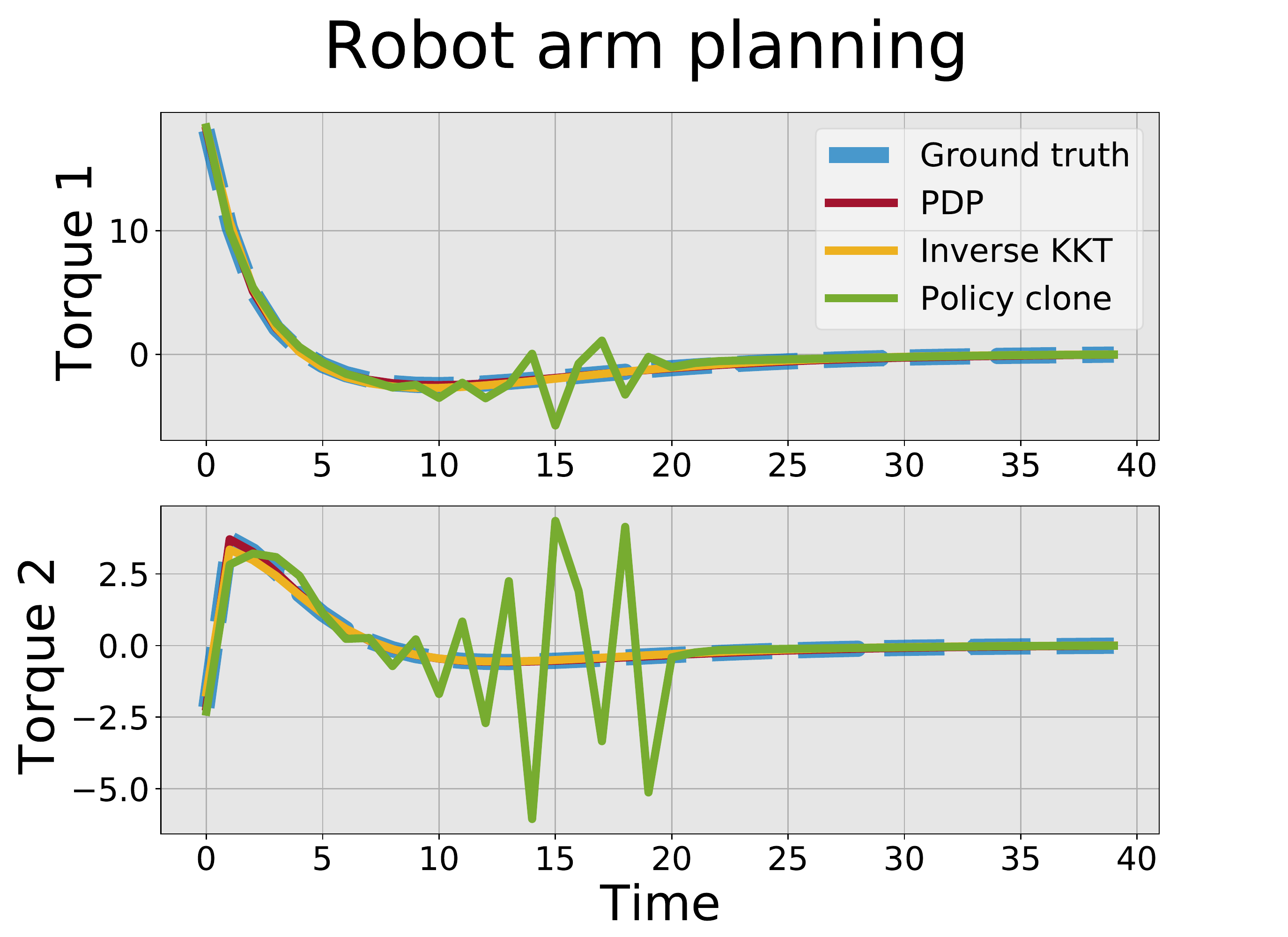}
		\caption{Robot arm}
		\label{appendix-ioc-test.2}
	\end{subfigure}%
	\begin{subfigure}{.32\textwidth}
		\centering
		\includegraphics[width=\linewidth]{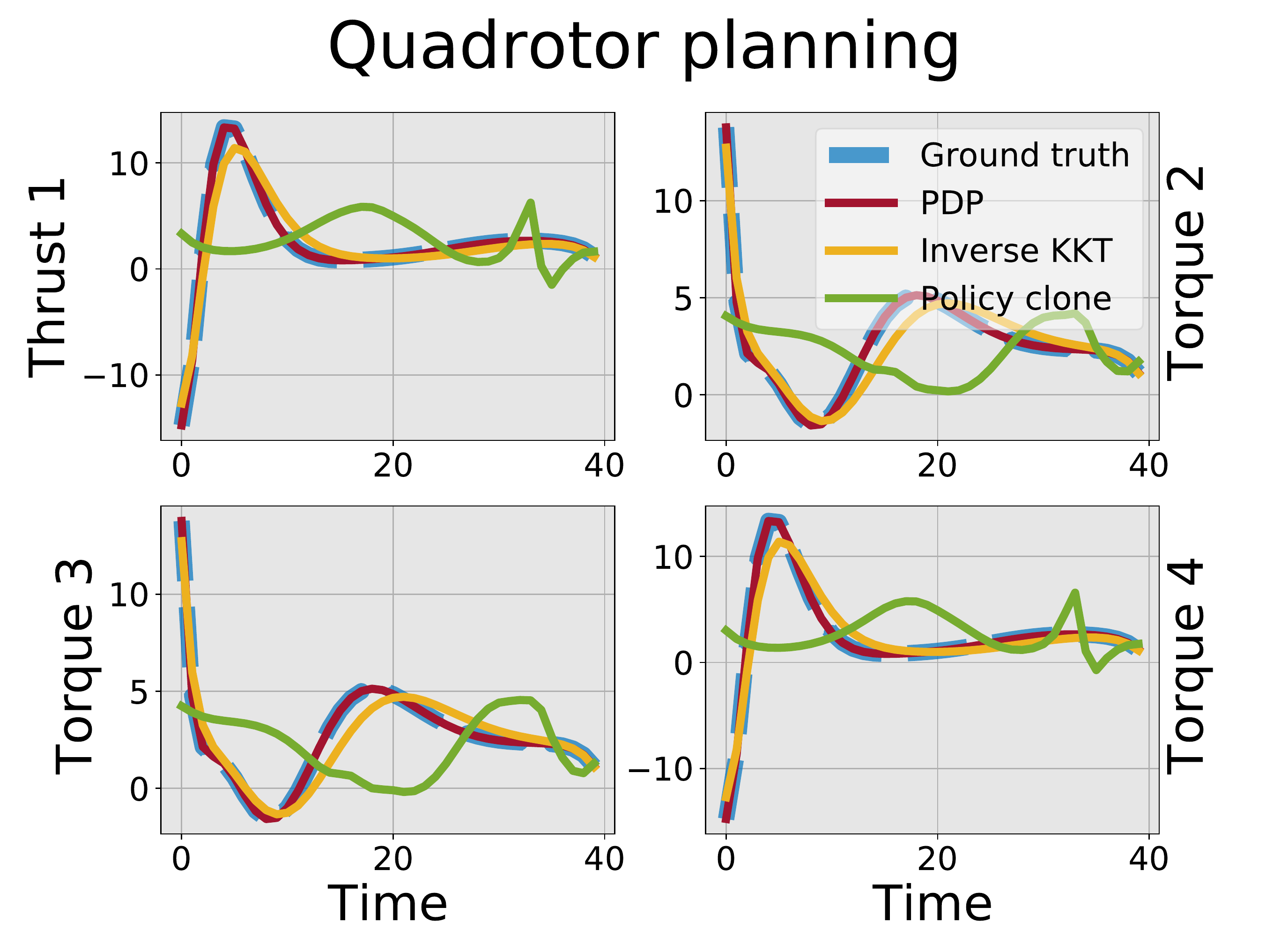}
		\caption{Quadrotor}
		\label{appendix-ioc-test.3}
	\end{subfigure}
	\caption{Validation for the imitation learning experiment in  Fig. \ref{appendix-ioc}. We preform motion planing for each system in unseen conditions (new initial condition and new time horizon) using the learned models. Results show that compared to the neural policy cloning  and inverse KKT \cite{englert2017inverse},  PDP result can accurately plan the expert's trajectory in unseen settings. This indicates PDP can accurately learn the  dynamics and control objective, and has the better generality than the other two. Although  policy imitation  has lower imitation loss than inverse KKT, it has the poorer performance in planing. This is  because with limited data, the cloned policy  can be over-fitting, while the inverse KKT learns a cost function,  a high-level representation of policies, thus has better  generality to unseen conditions.}
	\label{appendix-ioc-test}
\end{figure}

\textbf{Results and validation.} \quad
In Fig. \ref{appendix-ioc}, we show more detailed results of imitation loss versus iteration for  three systems (cart-pole, robot arm, and quadrotor). On each system, we run five trials for all methods with random initial guess, and the learning rate for all methods is set as $\eta=10^{-4}$. In  Fig. \ref{appendix-ioc-test}, we validate the learned models (i.e., learned dynamics and learned control objective) by performing motion planning of each system in unseen settings. Specifically, we set each system with new initial state $\boldsymbol{x}_{0}$ and horizon $T$ and plan the control trajectory using the learned models, and we also show the corresponding true  trajectory of the expert.

\subsection{Experiment of System Identification} \label{appendix-exp-sysid}

\textbf{Data acquisition.} \quad
In the  system identification experiment, we collect a total number of five trajectories from  systems (in Table \ref{experimenttable}) with dynamics known, wherein different trajectories $\boldsymbol{\xi}^{\text{o}}=\{\boldsymbol{{x}}_{0:T}^{\text{o}},\boldsymbol{{u}}_{0:T-1}\}$  have different initial conditions $\boldsymbol{x}_0$ and horizons $T$ ($T$ ranges from $10$ to $20$), with random inputs $\boldsymbol{{u}}_{0:T-1}$ drawn from uniform distribution.

\textbf{DMDc method.} \quad
The DMDc method \cite{proctor2016dynamic}, which can be viewed as a variant of Koopman theory \cite{koopman1931hamiltonian},  estimates a linear dynamics model $\boldsymbol{x}_{t+1}=A\boldsymbol{x}_t+B\boldsymbol{u}_t$, using the following least square regression
\begin{equation}
\min_{A,B}\sum\nolimits_{t=0}^{T-1}\norm{\boldsymbol{x}_{t+1}^{\text{o}}-A\boldsymbol{x}_t^{\text{o}}-B\boldsymbol{u}_t}^2.
\end{equation}

\textbf{Neural network baseline.} \quad
For the neural network baseline, we use a neural network $\boldsymbol{f}_{\boldsymbol{\theta}}(\boldsymbol{x},\boldsymbol{u})$ to represent the system dynamics, where the input of the network is state and control vectors, and output is the state of next step. We train the neural network by minimizing the following residual
\begin{equation}
\min_{\boldsymbol{\theta}}\sum\nolimits_{t=0}^{T-1}\norm{\boldsymbol{x}_{t+1}^{\text{o}}-\boldsymbol{f}_{\boldsymbol{\theta}}(\boldsymbol{x}_t^{\text{o}},\boldsymbol{u}_t)}^2.
\end{equation}

\medskip
\textbf{Learning neural dynamics model.} \quad
In Fig. \ref{figid.4}, we compare the performance of  PDP with  Adam \cite{kingma2015adam} for learning the same neural dynamics model for the robot arm system. Here, the neural dynamics model is a fully-connected feed-forward neural network with   \texttt{(m+n)-(2m+2n)-n} layers and $\tanh$ activation functions, that is, an input layer with \texttt{(m+n)} neurons equal to the dimension of state, $n$, plus the dimension of control $m$, one hidden layer with \texttt{(2m+2n)} neurons and one output layer with \texttt{(n)} neurons. The learning rate for the PDP and the PyTorch Adam is both set as $\eta=10^{-5}$.

\medskip

\begin{figure}
	\begin{subfigure}{1\textwidth}
		\centering
		\includegraphics[width=\linewidth]{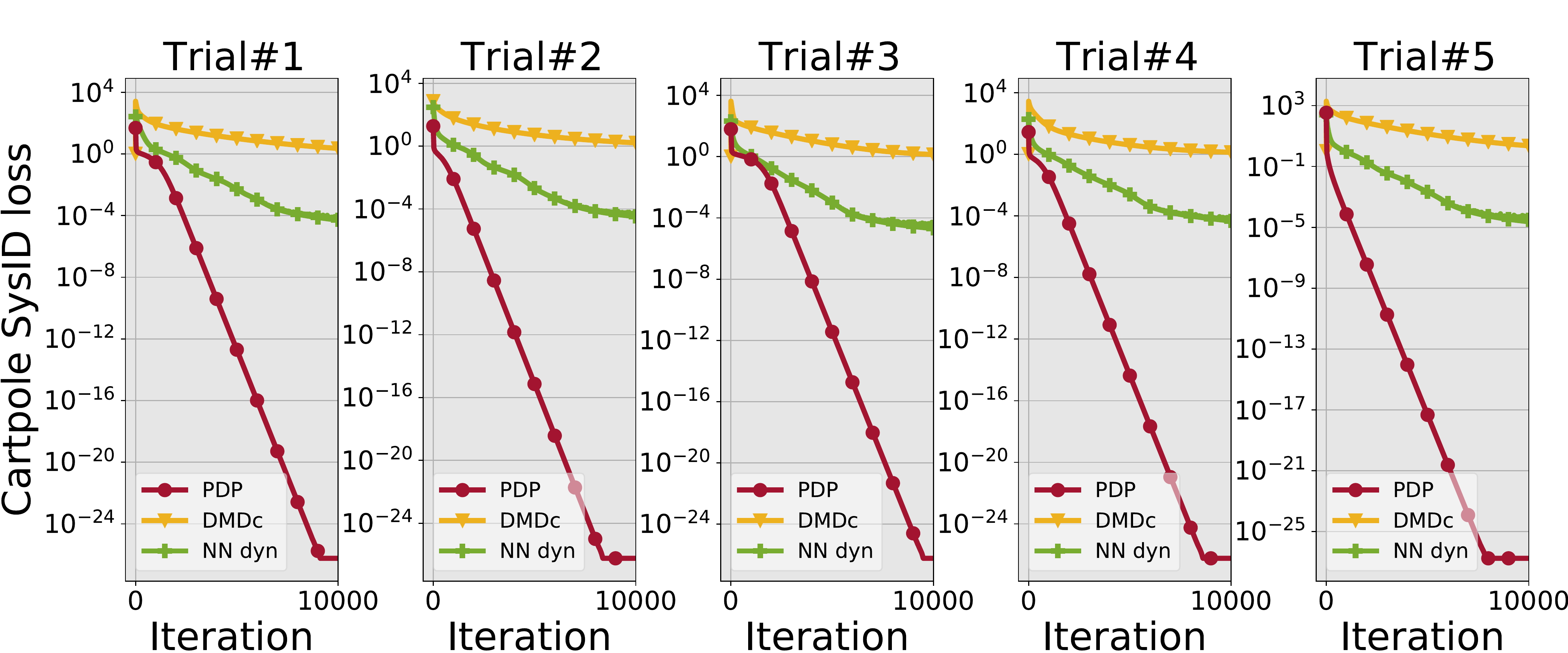}
		\label{appendix-id.1}
	\end{subfigure}
	\begin{subfigure}{1\textwidth}
		\centering
		\includegraphics[width=\linewidth]{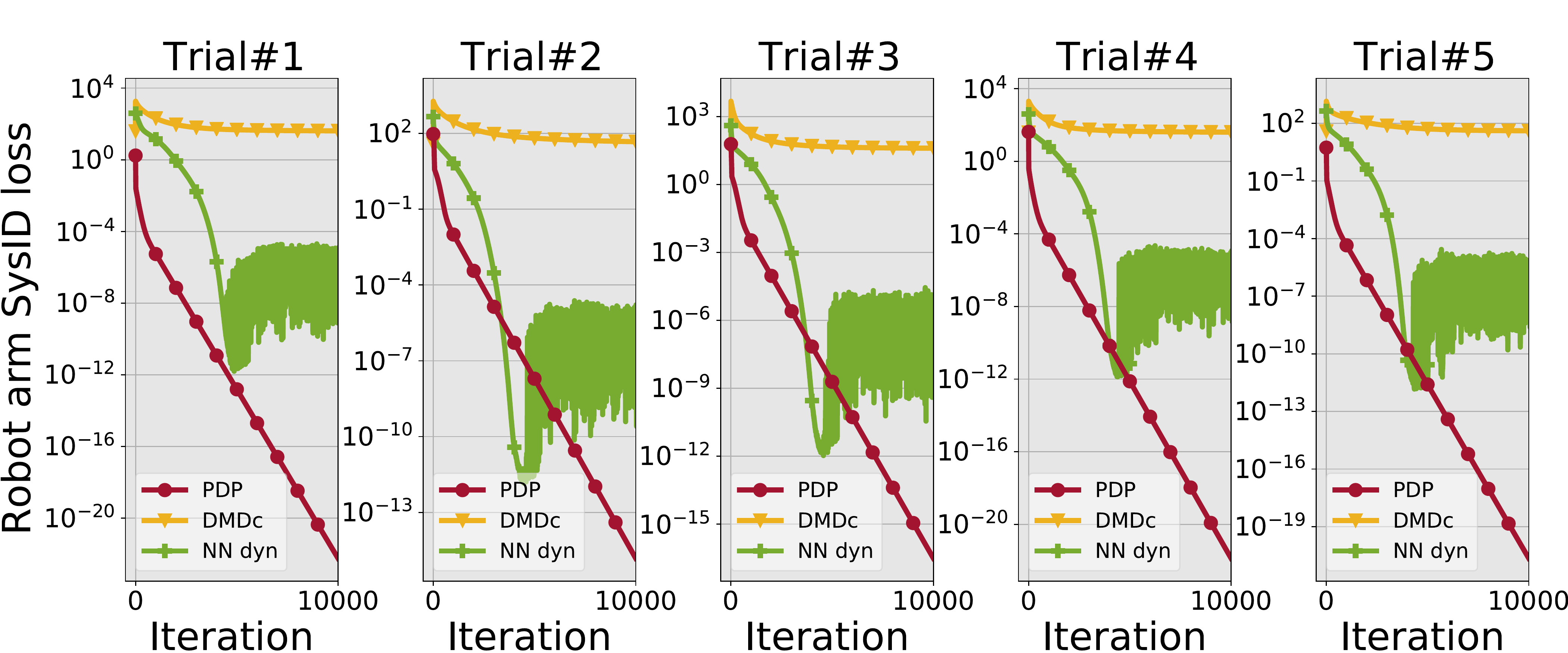}
		\label{appendix-id.2}
	\end{subfigure}
	\begin{subfigure}{1\textwidth}
		\centering
		\includegraphics[width=\linewidth]{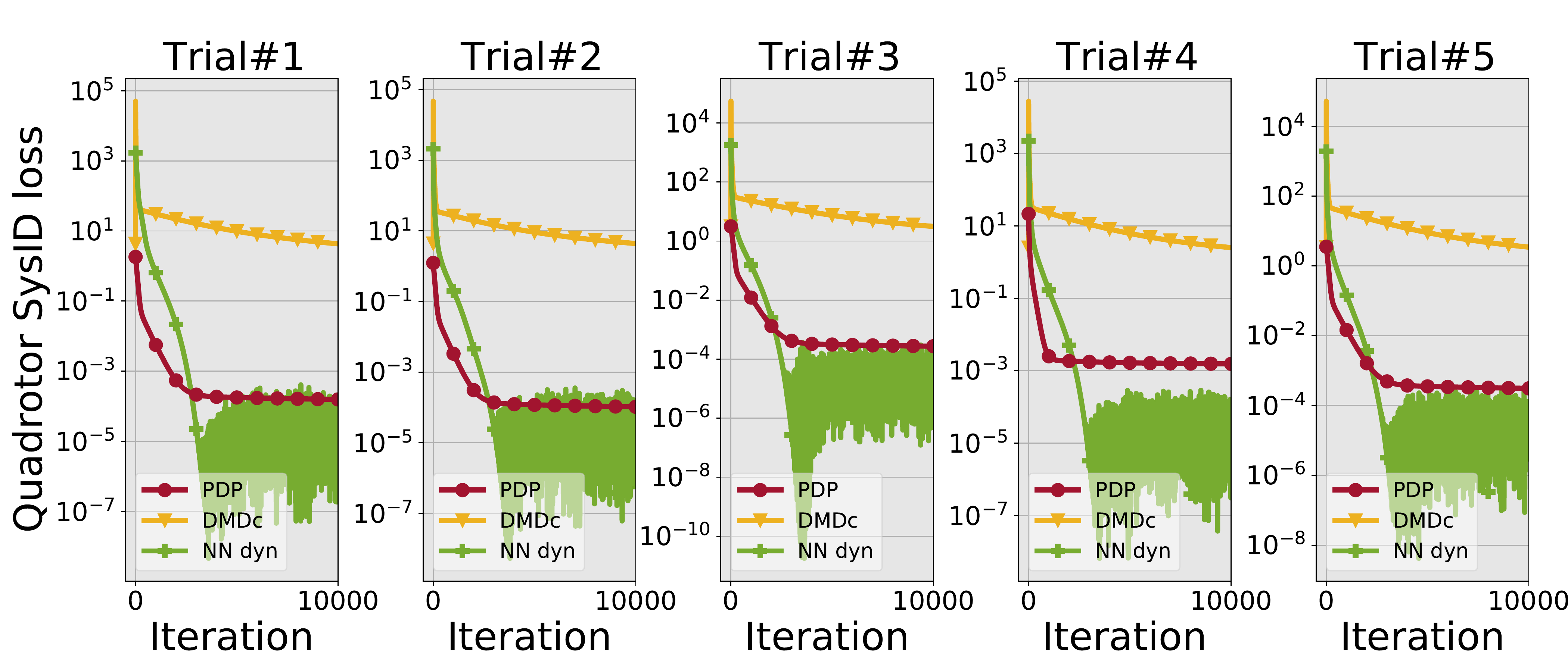}
		\label{appendix-id.3}
	\end{subfigure}
	\caption{Experiments for PDP SysID Mode: SysID loss versus iteration. For each system, we run five trials with random initial guess $\boldsymbol{\theta}_0$, and set the learning rate $\eta=10^{-4}$ for all methods. The results show a significant advantage of the PDP over neural-network dynamics and DMDc in terms of lower training loss and faster convergence speed. Please see  Fig. \ref{appendix-id-test} for validation. Please find the video demo at  \url{https://youtu.be/PAyBZjDD6OY}.}
	\label{appendix-id}
\end{figure}

\begin{figure}[h]
	\centering
	\begin{subfigure}{.32\textwidth}
		\centering
		\includegraphics[width=\linewidth]{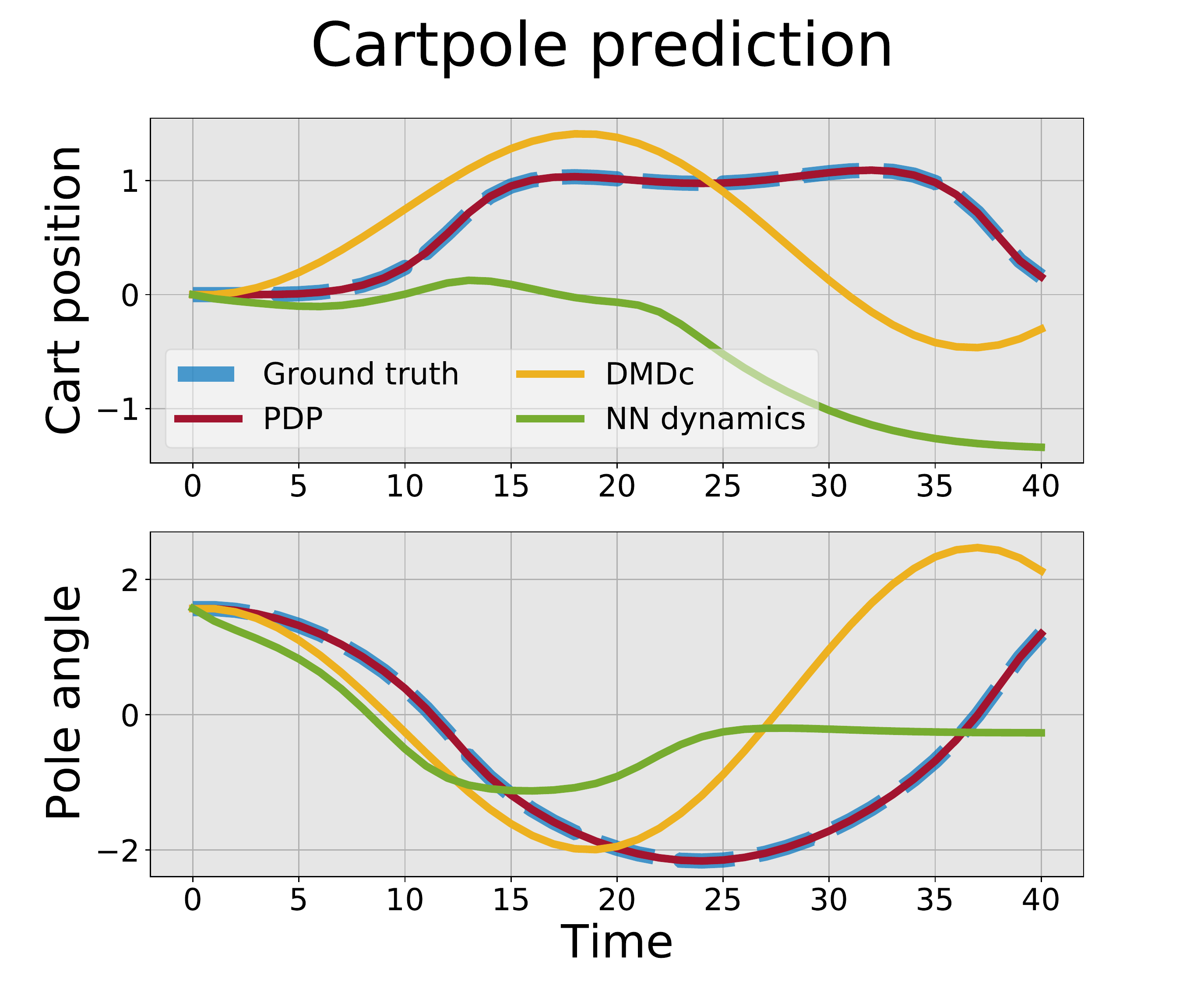}
		\caption{Cart-pole}
		\label{appendix-id-test.1}
	\end{subfigure}
	\begin{subfigure}{.32\textwidth}
		\centering
		\includegraphics[width=\linewidth]{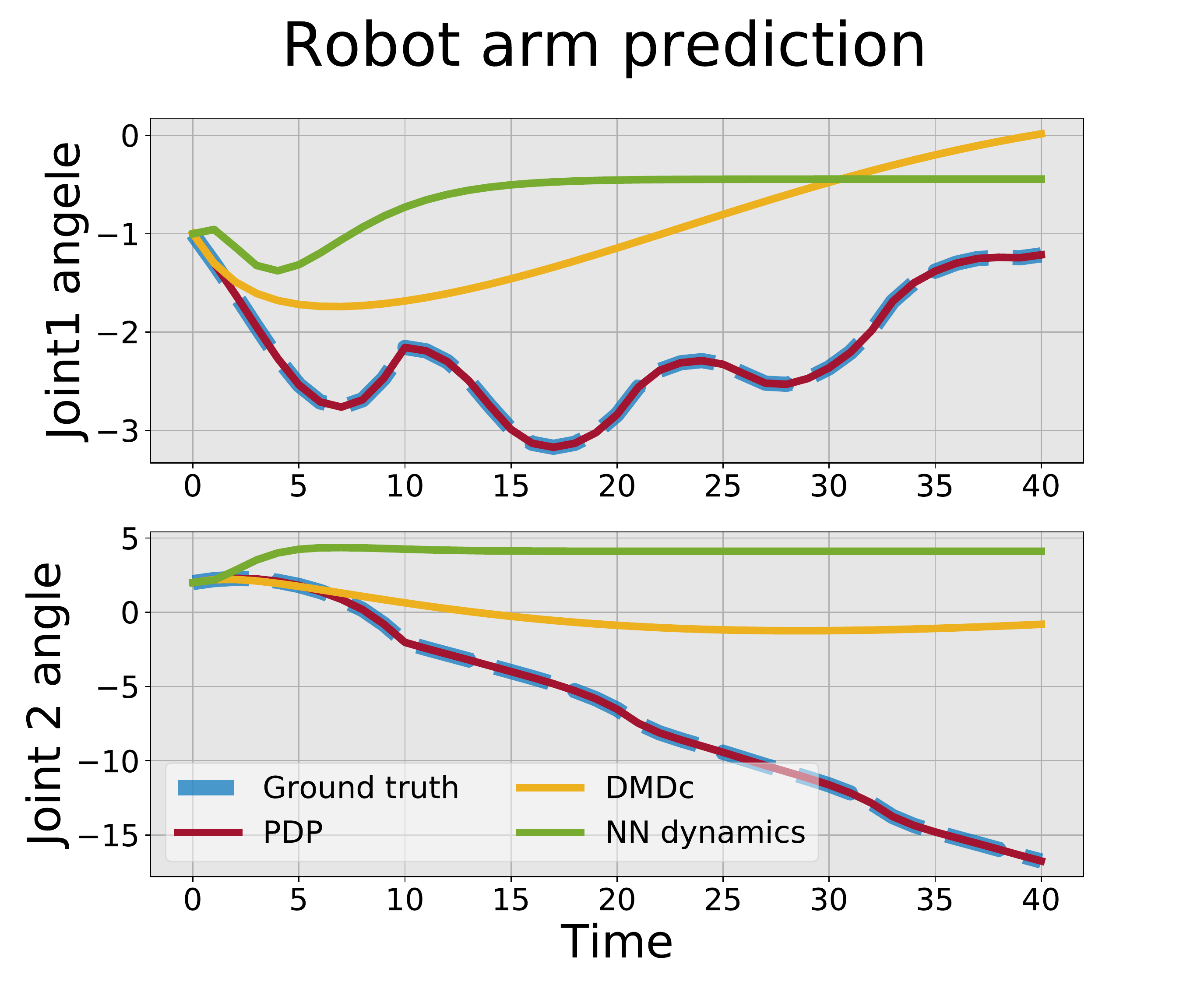}
		\caption{Robot arm}
		\label{appendix-id-test.2}
	\end{subfigure}%
	\begin{subfigure}{.32\textwidth}
		\centering
		\includegraphics[width=\linewidth]{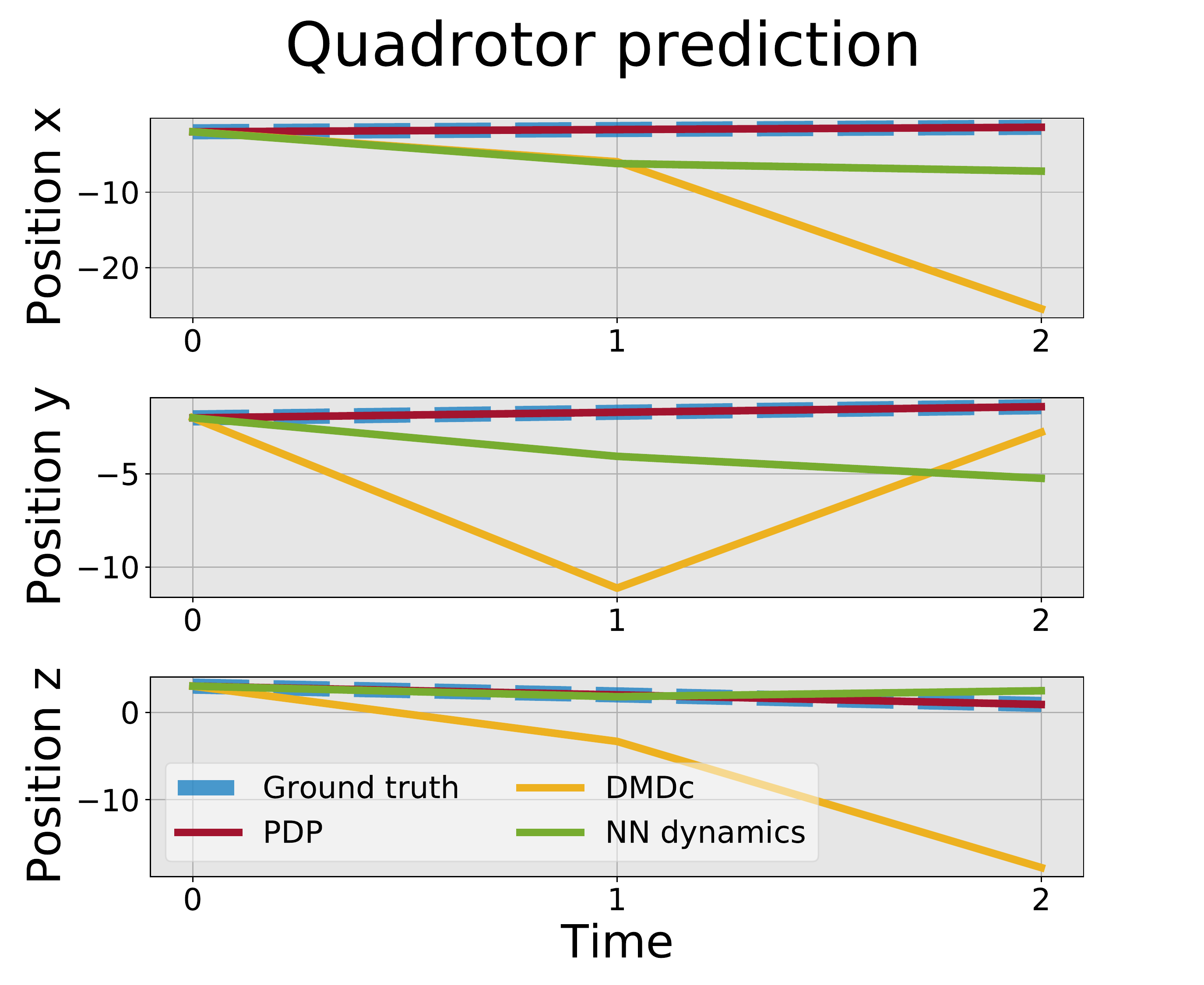}
		\caption{Quadrotor}
		\label{appendix-id-test.3}
	\end{subfigure}
	\caption{Validation for the system identification experiment in  Fig. \ref{appendix-id}. We perform motion prediction  in unactuated conditions ($\boldsymbol{u}=0$) using the learned dynamics. Results show that compared to neural-network dynamics training and DMDc,  PDP can accurately predict the motion trajectory of each systems. This indicates the effectiveness of the PDP in identifying dynamics models.
	}
	\label{appendix-id-test}
\end{figure}

\textbf{Results and validation.} \quad
In  Fig. \ref{appendix-id}, we show more detailed results of SysID loss versus iteration for the three systems (cart-pole, robot arm, and quadrotor).  On each system, we run five trials  with random initial guess, and we set the learning rate as $\eta=10^{-4}$ for all methods. 
In  Fig. \ref{appendix-id-test}, we use the learned dynamics model  to perform motion prediction of each system in unactuated conditions (i.e., $\boldsymbol{u}_t=\boldsymbol{0}$), in order to validate the effectiveness/correctness of the learned dynamics models.

\subsection{Experiment of  Control/Planning} \label{appendix_exp_pdp_oc}

We use the dynamics identified in the system ID part, and the specified control objective function is set as weighted distance to the goal, as given in Table \ref{experimenttable} ($\boldsymbol{\theta}_{\text{obj}}$ is given). Throughout the optimal control/planning experiments, we  use the time horizons $T$ ranging from $20$ to $40$.

\textbf{Learning  neural network policies.} \quad
On the cart-pole and robot-arm systems (in Fig. \ref{figoc.1} and Fig. \ref{figoc.2}), we learn a feedback policy by minimizing  given control objective functions. For both systems, we parameterize the policy using a neural network. Specifically, we use a fully-connected feed-forward neural network which has a layer structure of   \texttt{n-n-m} with $\tanh$ activation functions, i.e., there is an input layer with \texttt{n} neurons equal to the dimension of state,  one hidden layer with \texttt{n} neurons and one output layer with \texttt{m} neurons. The policy parameter $\boldsymbol{\theta}$ is the neural network parameter. We apply the PDP Control/Planning mode in Algorithm \ref{agpdpplan} and set the learning rate $\eta=10^{-4}$. For comparison, we  apply the  guided policy search (GPS) method \cite{levine2013guided} (its deterministic version) to learn the same neural policy with the learning rate $\eta=10^{-6}$ ($\eta$ in GPS is used to update the Lagrange multipliers for the policy constraint and we choose $\eta=10^{-6}$ because it achieves the most stable  results).

\textbf{Motion planning with Lagrange polynomial policies.} On the 6-DoF quadrotor, we use  PDP  to perform motion planning, that is, to find a control sequence to minimize the given control cost (loss) function. Here, we parameterize the  policy $\boldsymbol{u}_t=\boldsymbol{u}(t,\boldsymbol{\theta})$ as $N$-degree Lagrange polynomial \citep{abramowitz1948handbook} with $N+1$ pivot points evenly populated over the time  horizon, that is, $\{(t_0,\boldsymbol{u}_0), (t_1,\boldsymbol{u}_1), \cdots, (t_{N},\boldsymbol{u}_{N})\}$ with $t_i=iT/N$, $i=0,\cdots,N$. The analytical form of the parameterized  policy  is
\begin{equation}\label{polypolicy1}
\boldsymbol{u}(t, \boldsymbol{\theta})=\sum_{i=0}^{N}\boldsymbol{u}_{i}b_{i}(t) \quad \quad \text{with} \quad\quad b_{i}(t)=\prod_{\substack{0\leq j \leq N,  j \neq i}}\frac{t-t_j}{t_i-t_j}.
\end{equation}
Here, $b_{i}(t)$ is called  Lagrange basis, and the policy parameter $ \boldsymbol{\theta} $ is defined as
\begin{equation}
\boldsymbol{\theta}=[\boldsymbol{u}_0,\cdots,\boldsymbol{u}_N]^\prime\in\mathbb{R}^{m(N+1)}.
\end{equation}
The above Lagrange polynomial parameterization has been normally used in some trajectory optimization method such as \cite{elnagar1995pseudospectral,patterson2014gpops}.  
In this planning experiment, we have used different degrees of  Lagrange polynomials, i.e., $N=5$ and $N=35$, respectively, to show how policy expressiveness can influence the final control loss (cost). The learning rate in PDP is set as $\eta=10^{-4}$. For comparison, we also apply iLQR  \cite{li2004iterative} to solve for the optimal control sequence.

\begin{figure}
	\begin{subfigure}{1\textwidth}
		\centering
		\includegraphics[width=\linewidth]{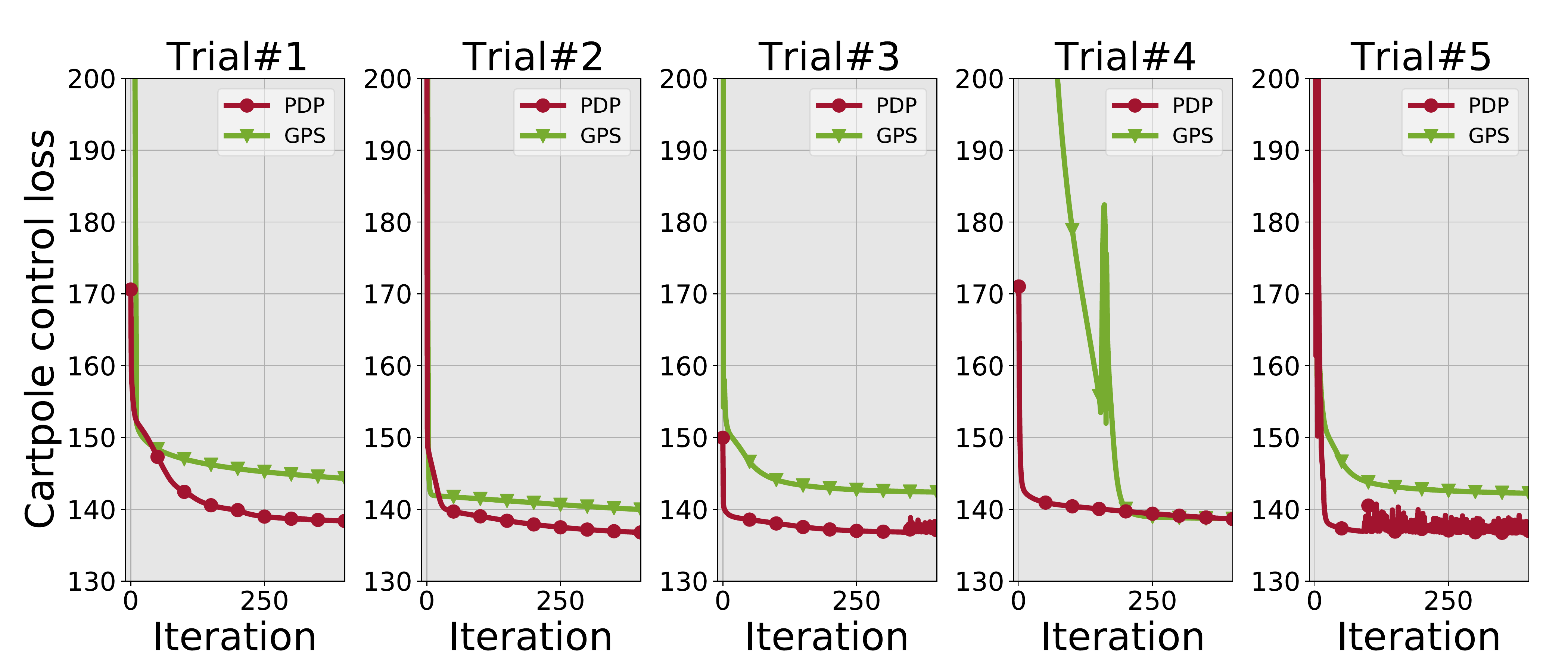}
		\label{appendix-oc.1}
	\end{subfigure}
	\begin{subfigure}{1\textwidth}
		\centering
		\includegraphics[width=\linewidth]{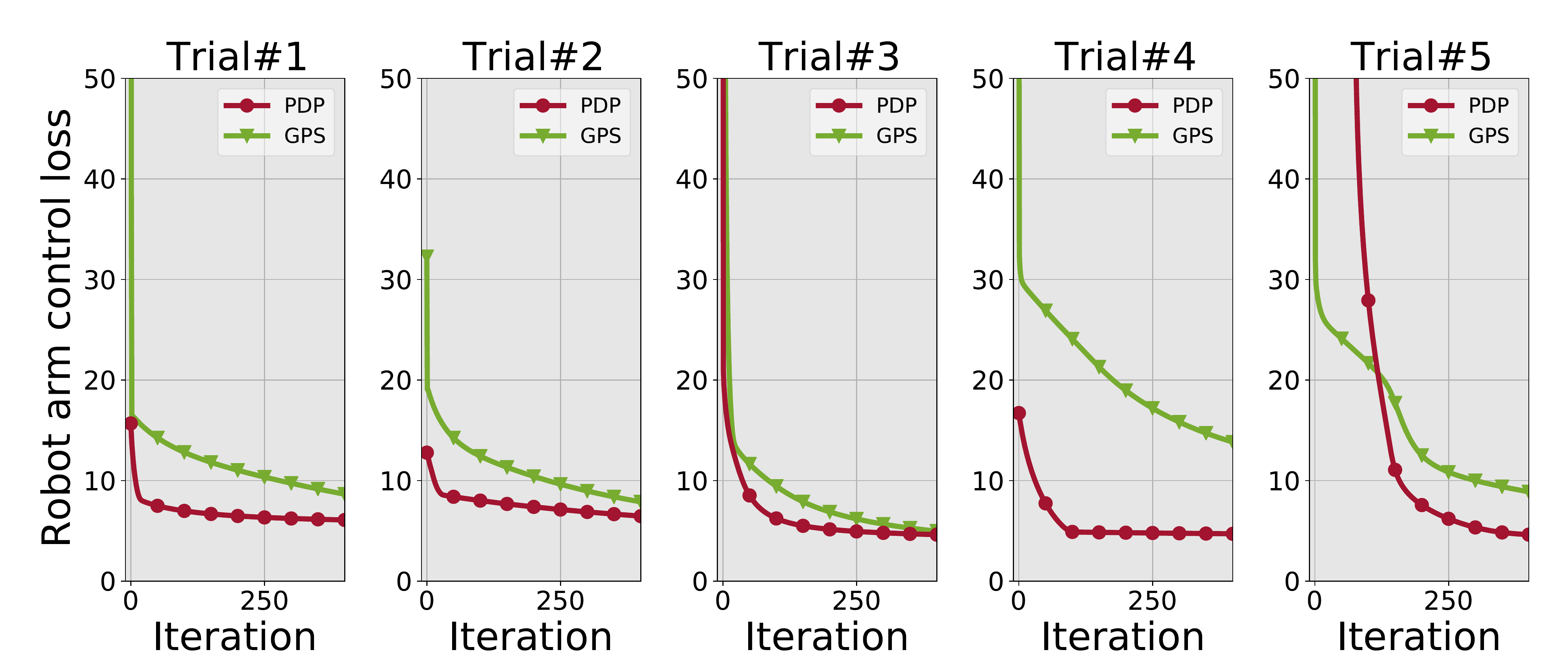}
		\label{appendix-oc.2}
	\end{subfigure}
	\begin{subfigure}{1\textwidth}
		\centering
		\includegraphics[width=\linewidth]{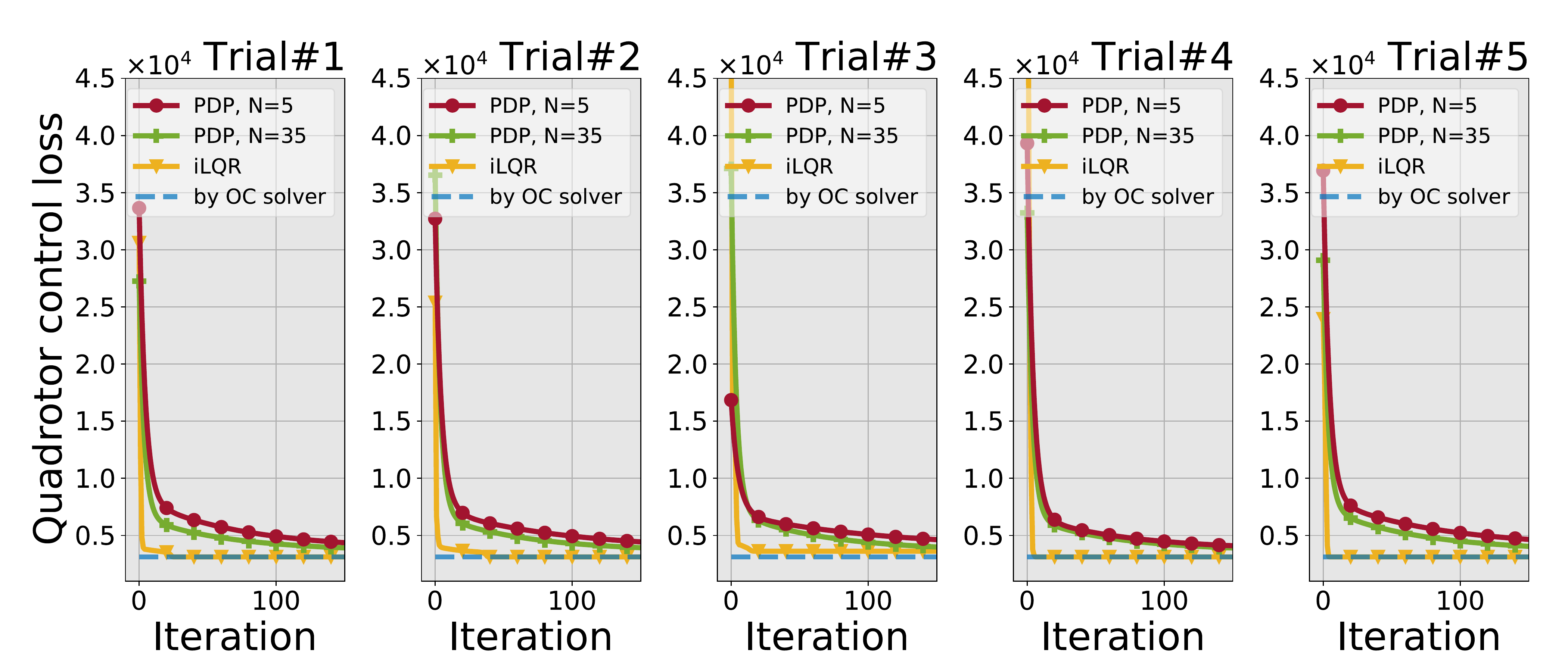}
		\label{appendix-oc.3}
	\end{subfigure}
	\caption{Experiments for PDP Control/Planning Mode: control loss (i.e., objective function value) versus iteration. For the cart-pole (top panel) and robot arm (middle panel) systems, we learn  neural feedback policies, and compare with the GPS method \cite{levine2013guided}. For the quadrotor system, we perform motion planning with a Lagrange polynomial  policy (we use different degree $N$), and compare with iLQR and an OC solver \cite{Andersson2019}.	
   The results show that for learning feedback control policies,  PDP outperforms  GPS in terms of having lower control loss (cost); and for motion planning, iLQR has  faster convergence speed than PDP.
   Please find the video demo at  \url{https://youtu.be/KTw6TAigfPY}.}

	\label{appendix-oc}
\end{figure}

\begin{figure}[H]
	\centering
	\begin{subfigure}{.32\textwidth}
		\centering
		\includegraphics[width=\linewidth]{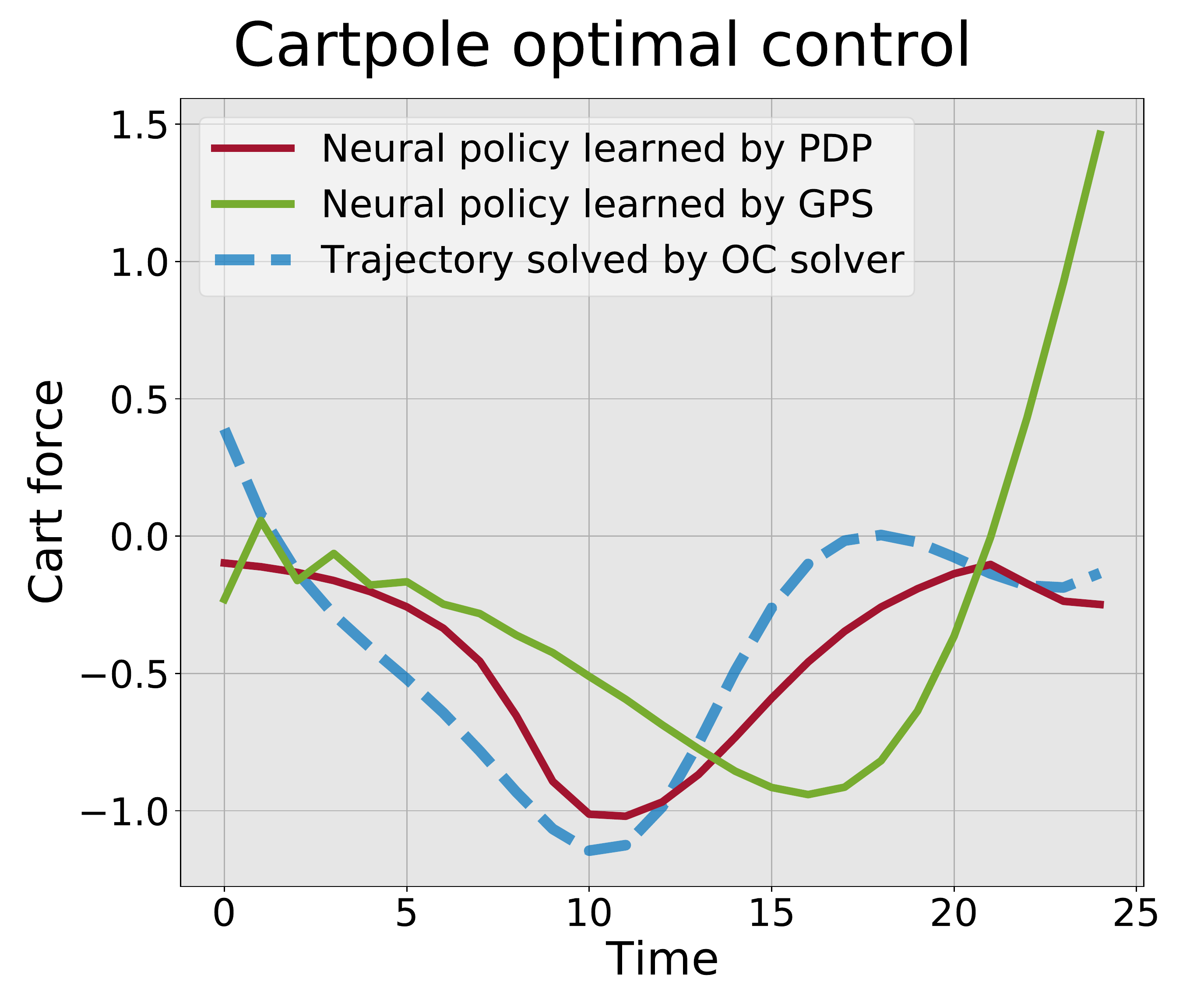}
		\caption{Cart-pole}
		\label{appendix-oc-test.1}
	\end{subfigure}
	\begin{subfigure}{.32\textwidth}
		\centering
		\includegraphics[width=\linewidth]{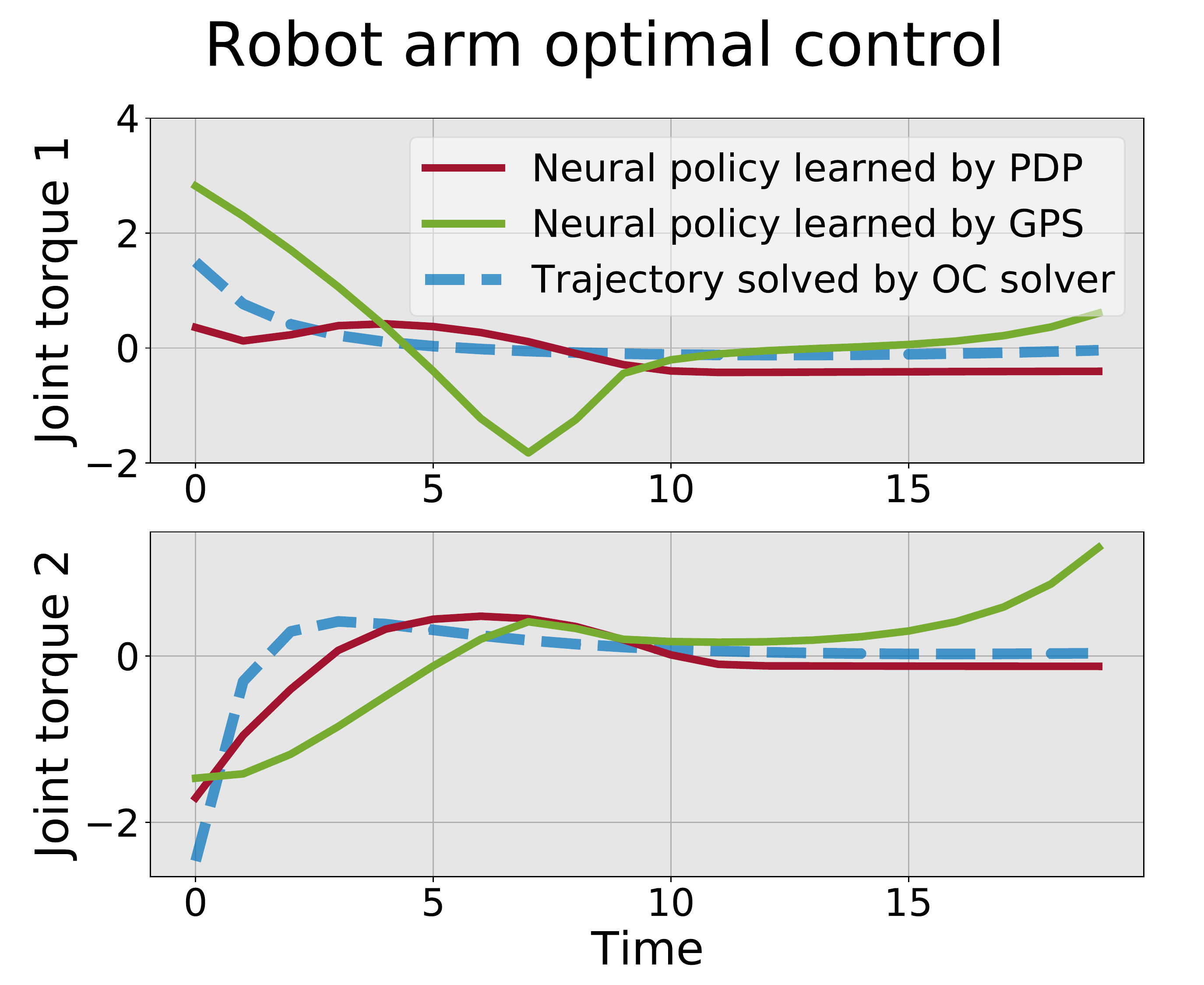}
		\caption{Robot arm}
		\label{appendix-oc-test.2}
	\end{subfigure}%
	\begin{subfigure}{.32\textwidth}
		\centering
		\includegraphics[width=\linewidth]{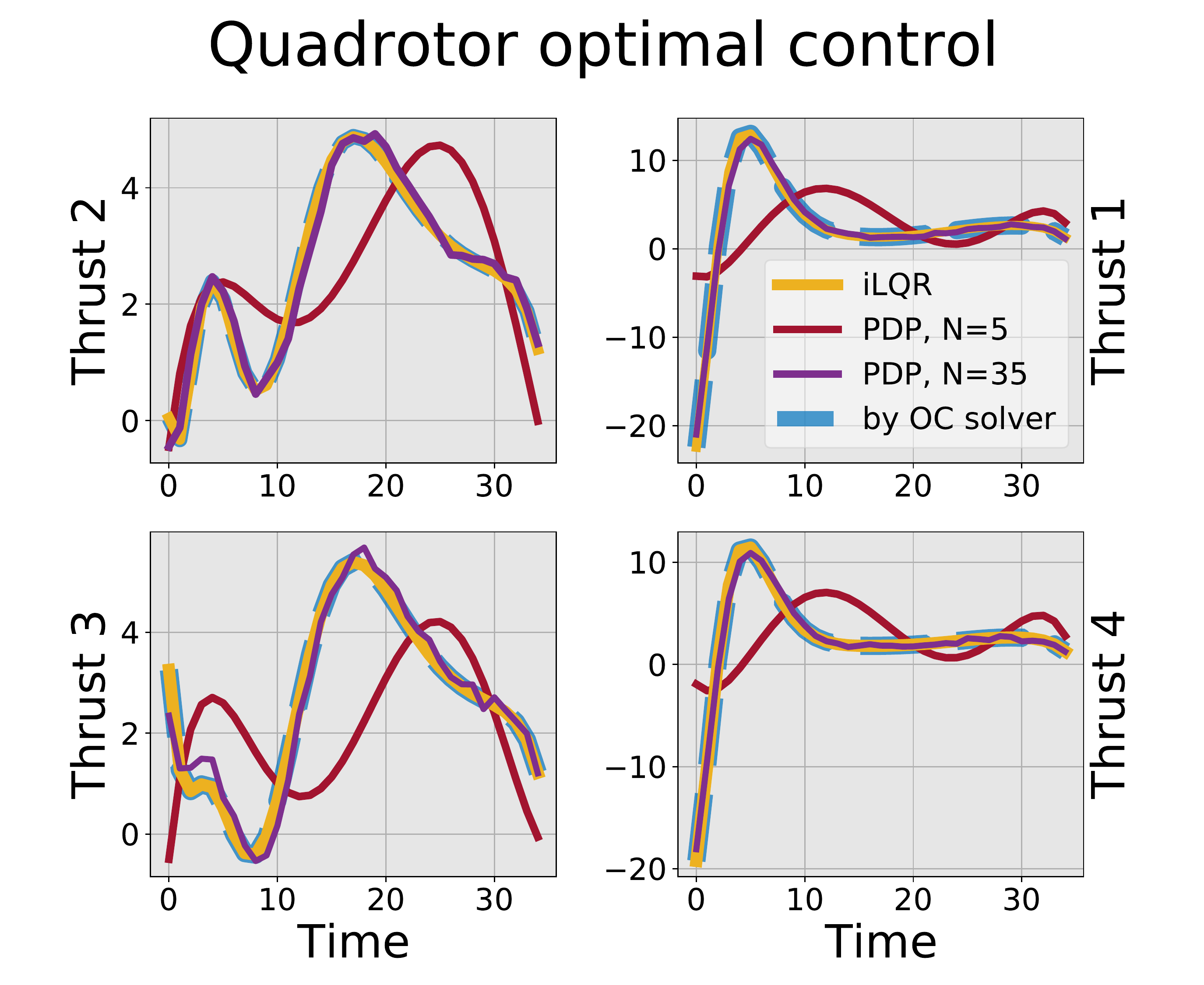}
		\caption{Quadrotor}
		\label{appendix-oc-test.3}
	\end{subfigure}
	\caption{Simulation of the learned policies in the  control and planning experiment in Fig. \ref{appendix-oc}. Fig. \ref{appendix-oc-test.1}-\ref{appendix-oc-test.2} are the simulations of the learned neural feedback policies on the cart-pole and robot arm systems, respectively, where we also plot the optimal trajectory  solved by an OC solver \cite{Andersson2019} for reference. From Fig. \ref{appendix-oc-test.1}-\ref{appendix-oc-test.2}, we observe that  PDP results in a  trajectory that is much closer to the optimal one than that of GPS; this implies that PDP has lower control loss (please check our analysis on this in Appendix \ref{appendix_exp_pdp_oc}) than GPS.  Fig. \ref{appendix-oc-test.3} is the planning results for the quadrotor system using PDP,  iLQR, and an OC solver \cite{Andersson2019}, where we have used different degrees of Lagrange polynomial policies in PDP.  The results show that PDP can successfully plan a trajectory very close to the ground truth optimal trajectory. We also observe that the accuracy of the resulting trajectory depends on choice of the policy parameterization (i.e., expressive power): for example, the use of polynomial policy of a higher degree $N$ results in a trajectory closer to the optimal one (the one using the OC solver) than the use of a lower degree. iLQR is generally able to achieve high-accuracy solutions because it directly optimizes the loss function with respect to  individual control inputs (instead of a parameterized policy), but this comes at the cost of high computation expense, as shown in Fig. \ref{figoc.4}.}
	\label{appendix-oc-test}
\end{figure}

\textbf{Results} \quad In Fig. \ref{appendix-oc}, we show the detailed results of control loss (i.e. the value of control objective function) versus iteration  for three systems (cart-pole, robot arm, and quadrotor). For each system, we run five trials  with random initial parameter $\boldsymbol{\theta}_0$. In Fig. \ref{appendix-oc-test}, we apply the learned neural network policies (for cart-pole and robot arm systems) and the Lagrange polynomial policy (for quadrotor system) to simulate the corresponding system.  For reference, we also plot the optimal  trajectory solved by  an OC solver \cite{Andersson2019} (which corresponds to the minimal control cost).

\textbf{Comments on the result comparison  between GPS \cite{levine2013guided} and PDP.} \quad In learning feedback  policies, 
 comparing the results obtained by the guided policy search (GPS) \cite{levine2013guided} and PDP in  Fig. \ref{appendix-oc} and in Fig. \ref{appendix-oc-test}, we have the following remarks.

(1)  PDP outperforms  GPS in terms of having lower control loss (cost). This can be seen in Fig. \ref{appendix-oc} and Fig. \ref{appendix-oc-test} (in Fig. \ref{appendix-oc-test}, PDP results in a simulated trajectory which is closer to the optimal one than that of GPS). This can be understood from the fact that GPS considers the policy as constraint and updates it in a supervised learning step during the learning process.  Although  GPS aims to \emph{simultaneously} minimize the control cost and the degree to which the policy is violated, it does not necessarily mean that before the learning researches convergence, when \emph{strictly following} a pre-convergence control policy, the system will have a  cost as minimal as it can possibly achieve.
	
(2) Instead,  PDP  adopts a different way to synchronize the fulfillment of policy constraints and the minimization of the control cost. In fact, throughout the entire learning process, PDP always guarantees that the policy constraint is perfectly respected (as the forward pass strictly follows the policy). Therefore, the core difference between PDP and GPS is that PDP does not simultaneously  minimize two aspects---the policy violation and control cost, instead, it enforces that one aspect---policy---is always respected and only focuses on minimizing the other---control cost. The benefit of doing so is that at each  learning step, the control cost  for PDP is always as minimal as it can possibly achieve. This explains why PDP outperforms  GPS in terms of having lower control cost (loss).

\newpage

\section{Related End-to-End Learning  Frameworks}\label{comparewithotherend2endframeworks}

As discussed in Section  \ref{secton-comaprison-end-to-end-framework}, two categories are related to this work. Here, we only detail the difference of PDP from the second category, i.e., the methods that learn an implicit planner within a RL policy.

\textbf{Differentiable MPC.} \cite{amos2018differentiable} develops an end-to-end differentiable MPC framework to jointly learn the system dynamics model and control objective function of an optimal control system. In the forward pass, it first uses iLQR \cite{li2004iterative} to solve the optimal control system and find a fixed point, and then approximate the optimal control system  by a LQR  at the fixed point. In the backward pass,  the gradient is obtain by differentiating  the LQR approximation. This process, however, may have two drawbacks: first, 
since the differentiation in the backward pass is conducted on the LQR approximation instead of on the original system, the obtained gradient thus may not be accurate due to   discrepancy of  approximation; and second, computing the gradient of the LQR approximation requires the inverse of a coefficient matrix, whose size is $(2n+m)T\times(2n+m)T$ with $n$ and $m$  state and action dimensions, respectively, $T$
 the time horizon of the OC system, thus this will cause huge computational cost when handling the system of longer time horizon $T$.

\begin{figure}[H]
	\centering
	\includegraphics[width=0.70\columnwidth]{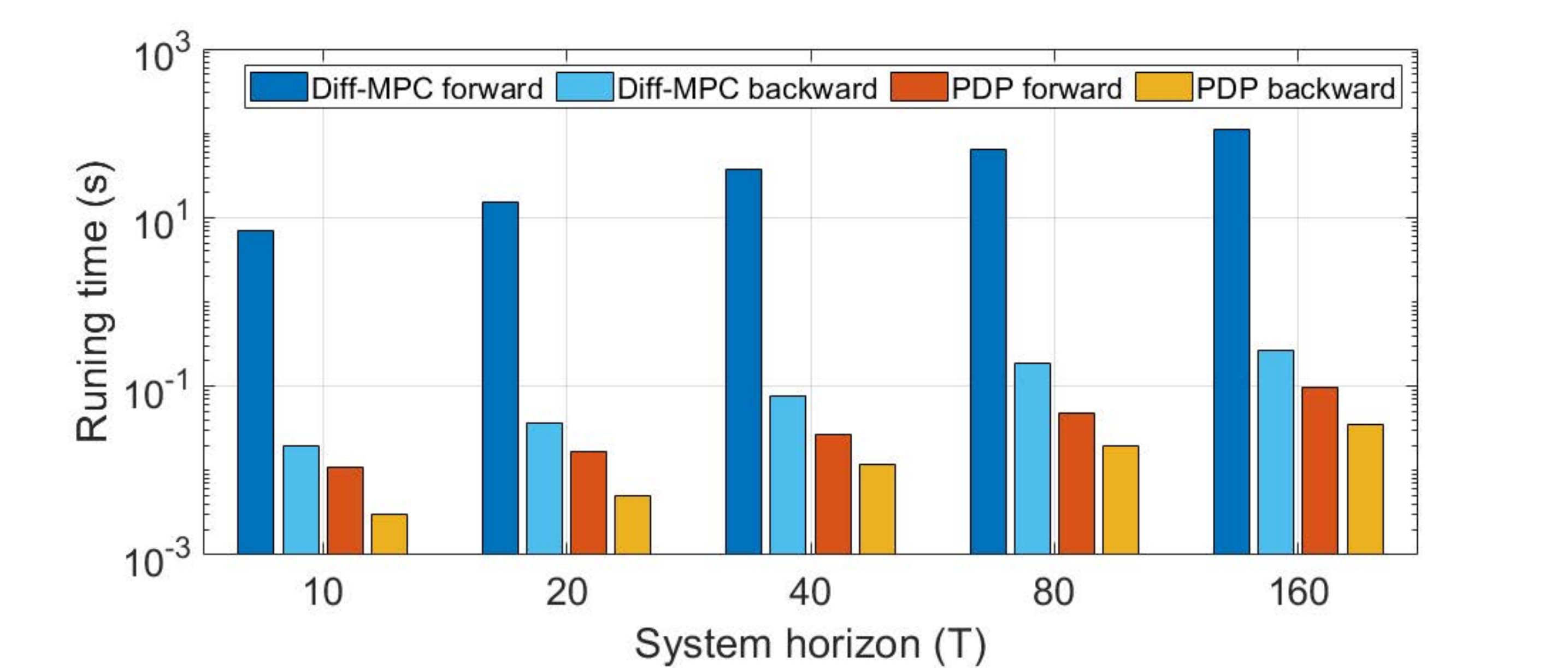}
	\caption{Runtime (per iteration) comparison between the  PDP and differentiable MPC \cite{amos2018differentiable} for different time horizons of a pendulum  system. Note that y-axis is log-scale, and the runtime is  averaged  over 100 iterations. Both  methods are implemented in Python and run on the same machine using CPUs. The results show that the PDP runs 1000x faster than differentiable MPC.}
	\label{comare}
\end{figure}

Compared to differentiable MPC, the first advantage of the PDP framework is that the differentiation in the  backward pass  is directly performed on the parameterized optimal control system (by differentiating through PMP). Second, we develop the auxiliary control system in the backward pass of PDP, whose trajectory  is exactly the gradient of the system trajectory in the forward pass. The gradient then is iteratively solved using the auxiliary control system by  Lemma \ref{theorem1} (Algorithm \ref{algsolvauxsys}). Those proposed techniques enables the PDP  to have significant advantage in computational efficiency over differentiable MPC. To illustrate this,  we have compare the algorithm complexity for both PDP and differentiable MPC in Table \ref{complexitytable} and provide an  experiment in Fig. \ref{comare}.

\textbf{Path Integral Network.}
 \cite{okada2017path} and \cite{pereira2018mpc} develop a differentiable end-to-end framework to learn path-integral optimal control systems.  Path-integral optimal control systems \cite{kappen2005path} however are a limited category of optimal control systems, where the dynamics is affine in control input and the control objective function  is quadratic in control input.  More differently, this path integral network is essentially an `unrolling' method, which means that the forward pass of solving optimal control is extended as a graph of multiple steps of applying gradient descent, and the solution of the optimal control system is considered as the output of the final step of the gradient descent operations. Although the advantage of this unrolling (gradient descent) computational graph is that it can immediately take advantage of  automatic differentiation techniques such as TensorFlow \cite{abadi2016tensorflow} to obtain the gradient in backpropagation, its drawback is however obvious: the framework  is both memory- and computationally- expensive because it needs to store and traverse all intermediate results of the gradient descent process along the graph; furthermore, there is a conflict between computational complexity and accuracy in the forward pass. We have provided its complexity analysis in Table \ref{complexitytable}.

\textbf{Universal Planning Network}.
In  \cite{srinivas2018universal}, the authors develop an end-to-end imitation learning framework  consisting of two layers: the inner layer is a planner, which is formulated as an optimal control system in a latent space and is solved by gradient descent, and an outer layer to minimize the imitation loss between the output of inner layer and  expert demonstrations. However, this framework is also based on the `unrolling' strategy. Specifically, the inner planning layer using gradient descent is considered as a large computation graph, which chains together the sub-graphs of each step of gradient descent. In the backward pass, the gradient derived
from the outer layer back-propagates through the entire computation graph.  Again, this unrolled learning strategy will incur huge memory and computation costs in implementation. Please find its complexity analysis in Table \ref{complexitytable}.

Different from the above `unrolling' learning methods \cite{okada2017path,pereira2018mpc,srinivas2018universal,tamar2017learning}, the proposed PDP method handles the learning of optimal control systems in a `direct and compact' manner. Specifically, in forward pass, PDP only obtains and stores the final solution of the optimal control system and  does not care about the (intermediate) process of how such solution is obtained. Thus, the forward pass of the PDP accepts any external optimal control solver such as  CasADi \cite{Andersson2019}. Using the solution in the  forward pass, the PDP then automatically builds the auxiliary control system, based on which, the exact \emph{analytical} gradient is   solved efficiently in backward pass. Such features guarantee that the complexity of the PDP framework is only linearly scaled up to the time horizon of the  system, which is significantly efficient than the above `unrolling' learning methods (please find the comparison in Table \ref{complexitytable}). In Appendix \ref{pdpcomplexity}, we will present the detailed complexity analysis.

\begin{table}[H]
	\centering
	\caption{Complexity comparison for different end-to-end learning frameworks}
	\label{complexitytable}
	\resizebox{\textwidth}{!}{%
		{	\renewcommand{\arraystretch}{1.1}	
			\begin{tabular}{|l|l|l|l|l|}
				\hline
				\multicolumn{1}{|c|}{\multirow{3}{*}{\thead{Learning\\frameworks}}} & \multicolumn{2}{c|}{Forward pass}   & \multicolumn{2}{c|}{Backward pass}                                                                                                                                                        \\ \cline{2-5} 
				\multicolumn{1}{|c|}{}                            & Method and accuracy                                                                                                  & \begin{tabular}[c]{@{}l@{}}Complexity\\ (linear to)\end{tabular}           & Method                                                                                               & \begin{tabular}[c]{@{}l@{}}Complexity\\ (linear to)\end{tabular}           \\ \hline
				PI-Net \cite{okada2017path}                                           & \begin{tabular}[c]{@{}l@{}}$N$-step unrolled graph \\ using gradient descent;\\ accuracy depends on $N$\end{tabular} & \begin{tabular}[c]{@{}l@{}}computation: $NT$ \\ memory: $NT$\end{tabular} & \begin{tabular}[c]{@{}l@{}}Back-propagation over \\ the unrolled graph\end{tabular}                 & \begin{tabular}[c]{@{}l@{}}computation: $NT$ \\ memory: $NT$\end{tabular} \\ \hline
				UPN  \cite{srinivas2018universal}                                             & \begin{tabular}[c]{@{}l@{}}$N$-step unrolled graph \\ using gradient descent;\\ accuracy depends on $N$\end{tabular} & \begin{tabular}[c]{@{}l@{}}computation: $NT$ \\ memory: $NT$\end{tabular} & \begin{tabular}[c]{@{}l@{}}Back-propagation over \\ the  unrolled graph\end{tabular}                 & \begin{tabular}[c]{@{}l@{}}computation: $NT$ \\ memory: $NT$\end{tabular} \\ \hline
				Diff-MPC \cite{amos2018differentiable}                                         & \begin{tabular}[c]{@{}l@{}}iLQR finds fixed points;\\ can achieve any accuracy\end{tabular}              & \begin{tabular}[c]{@{}l@{}}computation: --- \\ memory: $T$\end{tabular}  & \begin{tabular}[c]{@{}l@{}}Differentiate the LQR \\ approximation and\\ solve linear equations \end{tabular} & \begin{tabular}[c]{@{}l@{}}computation: $T^2$ \\ memory: $T^2$\end{tabular} \\ \hline
				PDP                                               & \begin{tabular}[c]{@{}l@{}}Accept any OC solver;\\ can achieve any accuracy\end{tabular}           & \begin{tabular}[c]{@{}l@{}}computation: ---, \\ memory: $T$\end{tabular}  & \begin{tabular}[c]{@{}l@{}} Auxiliary control system\end{tabular}             & \begin{tabular}[c]{@{}l@{}}computation: $T$, \\ memory: $T$\end{tabular}   \\ \hline
			\end{tabular}%	
	}}
	
	\vspace{3pt}
	\parnotes{\raggedright \small{\quad *Here $T$ denotes the time horizon of the system};  \par}
\end{table}

\section{Complexity  of  PDP} \label{pdpcomplexity}
We consider the algorithm complexity of different learning modes of  PDP (see Appendix \ref{appendixalgorithem}), and suppose that the time horizon of the parameterized system $\boldsymbol{\Sigma}(\boldsymbol{\theta})$ is $T$.

IRL/IOC Mode (Algorithm \ref{agpdpiocmode}):  in forward pass,  PDP needs to obtain and store the optimal trajectory  $\boldsymbol{\xi}_{\boldsymbol{{\theta}}}$ of the optimal control system $\boldsymbol{\Sigma}(\boldsymbol{\theta})$ in (\ref{oc}), and this optimal trajectory can be solved by any (external) optimal control solver. In backward pass,  PDP first uses $\boldsymbol{\xi}_{\boldsymbol{{\theta}}}$ to build the auxiliary control system  $\boldsymbol{\overline\Sigma}(\boldsymbol{{\xi}}_{\boldsymbol{\theta}})$ in (\ref{backoc}) and then  computes $\frac{\partial  \boldsymbol{{\xi}}_{\boldsymbol{{\theta}}}}{\partial \boldsymbol{{\theta}}}$ by Lemma \ref{theorem1}, which takes $2T$ steps.

SysID Mode (Algorithm \ref{agpdpid}): in forward pass,  PDP needs to obtain and store the trajectory  $\boldsymbol{\xi}_{\theta}$ of the original dynamics system $\boldsymbol{\Sigma}(\boldsymbol{\theta})$ in (\ref{ocmodeid}). Such trajectory is simply a result of iterative integration of (\ref{ocmodeid}), which takes $T$ steps. In backward pass,   PDP first uses $\boldsymbol{\xi}_{\theta}$ to build the auxiliary control system  $\boldsymbol{\overline\Sigma}(\boldsymbol{{\xi}}_{\boldsymbol{\theta}})$ in (\ref{iddiff}) and then computes $\frac{\partial  \boldsymbol{{\xi}}_{\boldsymbol{{\theta}}}}{\partial \boldsymbol{{\theta}}}$ by iterative integration of (\ref{iddiff}), which takes $T$ steps.

Control/Planning Mode (Algorithm \ref{agpdpplan}):  in forward pass,  PDP needs to obtain and store the trajectory  $\boldsymbol{\xi}_{\theta}$ of the  controlled system $\boldsymbol{\Sigma}(\boldsymbol{\theta})$ in (\ref{ocmodeplan}). Such trajectory is simply a result of iterative integration of (\ref{ocmodeplan}), which takes $T$ steps. In backward pass,  PDP first uses $\boldsymbol{\xi}_{\theta}$ to build an auxiliary control system  $\boldsymbol{\overline\Sigma}(\boldsymbol{{\xi}}_{\boldsymbol{\theta}})$ in (\ref{mode3controlback}) and then computes $\frac{\partial  \boldsymbol{{\xi}}_{\boldsymbol{{\theta}}}}{\partial \boldsymbol{{\theta}}}$ by  integration of (\ref{mode3controlback}), which takes $T$ steps.

Therefore, we can summarize that the memory- and computational- complexity for the PDP framework is only linear to the time horizon $T$ of the parameterized system $\boldsymbol{\Sigma}(\boldsymbol{\theta})$. This is significantly advantageous over existing end-to-end learning frameworks, as summarized in  Table \ref{complexitytable}.

\section{Limitation of  PDP}\label{pdplimitation}

\textbf{PDP is a first-order algorithm.} We observe that (i)  all  gradient quantities in PDP are analytical and exact;  (ii)  the  development of PDP does not involve any second-order derivative/approximation of functions or models (note that PMP is  a first-order optimality condition for optimal control); and (iii) PDP  minimizes a  loss function directly with respect to  unknown parameters in a  system using gradient descent. Thus, we conclude that {PDP is a first-order gradient-descent based optimization framework}. Specifically for the  SysID and Control/Planning modes of PDP, they  are  also  first-order algorithms. When using these modes to solve optimal control problems, this first-order nature may bring  disadvantages of PDP  compared to high-order methods, such as  iLQR which can be  considered as 1.5-order because it uses  second-order derivative of a value function and  first-order derivative of  dynamics, or DDP which is a second-order method as it uses the second-order derivatives of both value  function and dynamics. The disadvantages of PDP have already been empirically shown in Fig. \ref{figoc.3} and Fig. \ref{appendix-oc}, where the converging speed of  PDP in its planning mode is  slower than that of iLQR. 
For   empirical comparisons  between first- and second-order  techniques, we refer the reader to  \cite{xu2020second}.

\textbf{Convergence to local minima.}
\emph{Since PDP is a first-order gradient-descent based algorithm, PDP can only achieve local minima for  general non-convex optimization problems in (\ref{prob})}. Furthermore, we  observe that the general problem  in (\ref{prob}) belongs to a bi-level optimization framework. As explored in \cite{ghadimi2018approximation}, under certain assumptions such as convexity and smoothness on  models  (e.g., dynamics model, policy,   loss function and control objective function),  global convergence of the  bi-level optimization can be established. But we think such conditions are too restrictive in the context of  dynamical control systems. As a future direction, we will investigate mild  conditions for good convergence by resorting to  dynamical system and control theory, such as Lyapunov theory.

\textbf{Parameterization matters for global convergence.} Although PDP only achieves local convergence, these still exists a question of how likely  PDP can obtain the global convergence.  In our empirical experiments, we find that how  models are parameterized matters for  good convergence performance. For example, in IOC/IRL mode, we observe that using a neural network control objective function (in Fig. \ref{figioc.4}) is more likely to get trapped in local minima than using the parameterization of weighted distance objective functions (in Fig. \ref{figioc.1}-\ref{figioc.3}). In control/planning mode, using a deeper neural network policy (in Fig. \ref{figoc.1}-\ref{figoc.2}) is more like to result in local minima than using a simpler one. Also in the motion planning experiment, we use the Lagrange polynomial to parameterize a policy instead of using standard polynomials, because the latter can lead to poor conditioning and sensitivity issues  (a small change of polynomial parameter results in large change in performance) and thus more easily get stuck in local minima. One high-level explanation is that more complex parameterization will bring extreme non-convexity to the  optimization problem, making  the algorithm more easily trapped in local minima. Again, how to theoretically justify those empirical experience and find the mild conditions for global convergence guarantee still needs to be investigated in  future research.

\section{PDP to Solve 6-DoF Rocket Powered Landing Problems}\label{rocketexperiment}

As a final part in this supplementary,  
we will demonstrate the capability of  PDP to solve the more challenging  6-DoF rocket powered landing problems. 

We here omit the  description of mechanics modeling for the 6-DoF powered rocket system,  and refer the reader to Page 5  in \cite{szmuk2018successive} for the rigid body dynamics model of a rocket system (the notations and coordinates used below follows the ones in \cite{szmuk2018successive}). The state vector of the rocket system is defined as 
\begin{equation}\label{rocketstate}
\boldsymbol{x}=\begin{bmatrix}
m&
\boldsymbol{r}_{\mathcal{I}}^\prime & \boldsymbol{v}_{\mathcal{I}}^\prime  &\boldsymbol{q}_{\mathcal{B}/\mathcal{I}}^\prime  & \boldsymbol{\omega}_{\mathcal{B}}^\prime 
\end{bmatrix}^\prime \in\mathbb{R}^{14}, 
\end{equation}
where $m\in\mathbb{R}$ is the mass of the rocket; $\boldsymbol{r}_{\mathcal{I}}\in \mathbb{R}^{3}$ and  $\boldsymbol{v}_{\mathcal{I}}\in \mathbb{R}^{3}$ are the position and velocity of the rocket (center of mass) in the  inertially-fixed Up-East-North coordinate frame; $\boldsymbol{q}_{\mathcal{B}/\mathcal{I}}\in \mathbb{R}^{4}$ is the unit quaternion denoting the attitude of rocket body frame with respect to the inertial frame (also see the description in the quadrotor dynamics in Appendix \ref{appendix_experiment_setup}); and $\boldsymbol{\omega}_{\mathcal{B}}\in \mathbb{R}^{3}$ is the angular velocity of the rocket expressed in the rocket body frame. In our simulation, we only focus on the final descending phase before landing, and thus assume the mass depletion during such a short phase is very slow and thus $\dot{m}\approx0$. We define the control input vector of the rocket, which is the thrust force vector
\begin{equation}
\boldsymbol{u}=\boldsymbol{T}_{\mathcal{B}}=[T_x, T_y, T_z]^\prime \in\mathbb{R}^{3},
\end{equation}
acting on the gimbal point of the engine (situated at the tail of the rocket) and is expressed in the body frame. Note that the relationship between the total torque $\boldsymbol{M}_{\mathcal{B}}$ applied to the rocket and the thrust force vector $\boldsymbol{T}_{\mathcal{B}} $ is $\boldsymbol{M}_{\mathcal{B}}=\boldsymbol{r}_{\mathcal{I},\mathcal{B}}\times \boldsymbol{T}_{\mathcal{B}}$, with $\boldsymbol{r}_{\mathcal{I},\mathcal{B}}\in\mathbb{R}^{3}$ being constant position vector from the center-of-mass of the rocket to the gimbal point of the engine.  The continuous dynamics is discretized using the Euler method: $\boldsymbol{x}_{t+1}=\boldsymbol{x}_{t}+\Delta\cdot \boldsymbol{f}(\boldsymbol{x}_t,\boldsymbol{u}_t)$ with the discretization interval $\Delta=0.1$s.

For the rocket  system, the unknown dynamics parameter, $\boldsymbol{\theta}_{\text{dyn}}$, includes the rocket's initial  mass  ${m}_0$, and the moment of inertia  ${\boldsymbol{J}}_{\mathcal{B}}\in\mathbb{R}^{3\times3}$, and the  rocket length   $\ell$, thus, $\boldsymbol{\theta}_{\text{dyn}}=\{m_0, \boldsymbol{J}_{\mathcal{B}},  \ell\} \in \mathbb{R}^8.$

For the control objective (cost) function, we consider a weighted combination of the following aspects:
\begin{itemize}
	\item distance  of the rocket position from the target position, associated with  weight $w_1$;
	\item distance  of the rocket velocity from the target velocity, associated with weight $w_2$;
	\item penalty of the excessive title angle of the rocket, associated with weight  $w_3$;
	\item penalty of the side effects of the thrust vector, associated with weight  $w_4$;
	\item penalty of the total fuel cost, associated with weighted $w_5$.
\end{itemize}
So the parameter of the control objective function, $\boldsymbol{\theta}_{\text{obj}}=\begin{bmatrix}
w_1,\,\, w_2,\,\, w_3,\,\, w_4,\,\, w_5
\end{bmatrix}^\prime\in \mathbb{R}^5$.
In sum, the overall parameter for the 6-DoF rocket powered landing control system is
\begin{equation}\label{thetarocket}
\boldsymbol{\theta}=\{\boldsymbol{\theta}_{\text{dyn}},\,\, \boldsymbol{\theta}_{\text{obj}}\}\in\mathbb{R}^{13}.
\end{equation}

\begin{figure}
	\centering
	\begin{subfigure}{.35\textwidth}
		\centering
		\includegraphics[width=\linewidth]{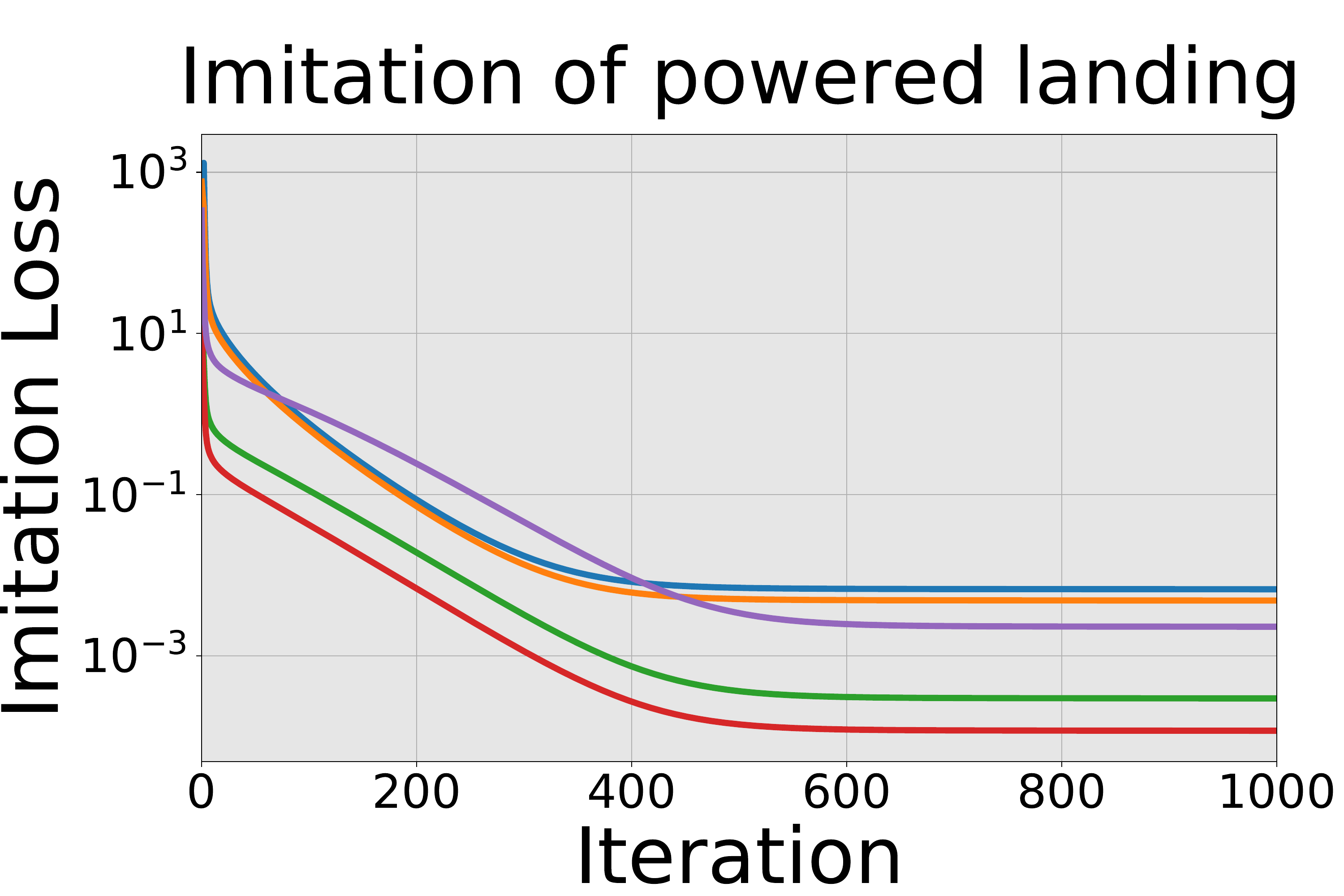}
		\caption{Training}
		\label{appendix-ioc-rocket.1}
	\end{subfigure}%
	\begin{subfigure}{.43\textwidth}
		\centering
		\includegraphics[width=\linewidth]{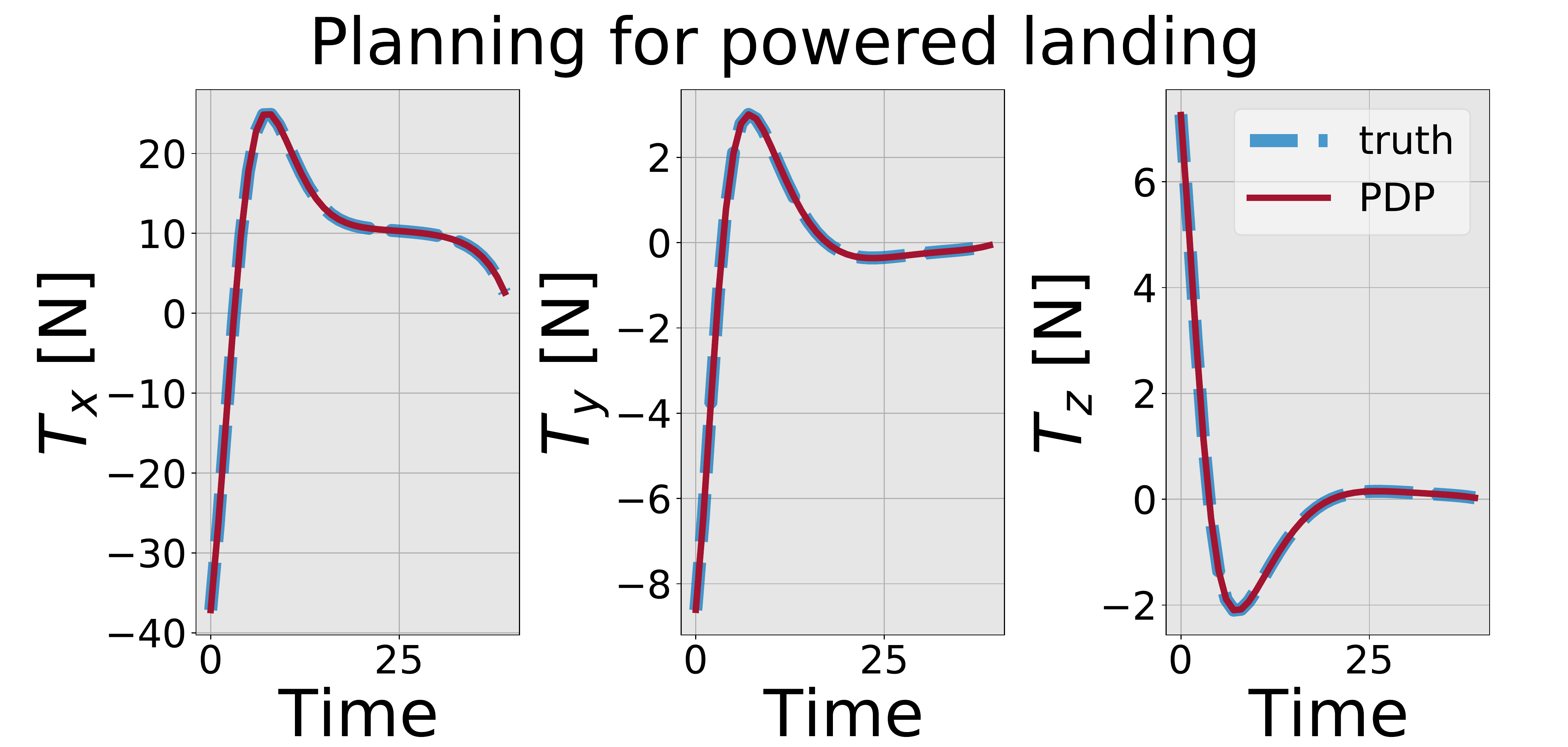}
		\caption{Validation}
		($T_x$ is defined along the rocket  direction)
		\label{appendix-ioc-rocket.2}
	\end{subfigure}%
	\caption{(a) Training process for imitation learning of 6-DoF rocket powered landing: the imitation loss versus iteration; here we have performed five trials  (labeled by different colors) with random initial parameter guess. (b) Validation: we use the learned models (dynamics and control objective function) to perform motion planning of the rocket powered landing in unseen settings (i.e. given new initial state condition and new time horizon requirement); here we also plot the ground-truth motion planning of the expert for reference. The results in (a) and (b) show that the PDP can accurately learn the  dynamics and control objective function from demonstrations, and have good generalizability to novel situations. Please find the video demo at  \url{https://youtu.be/4RxDLxUcMp4}.}
	\label{appendix-ioc-rocket}\vspace{5mm}
\end{figure}

\textbf{Imitation learning.} \quad We apply the IRL/IOC mode of  PDP to perform imitation learning of the 6-DoF rocket powered landing. The experiment process is similar to the experiments in Appendix \ref{apendix-exp-imitation}, where we collect five trajectories from an expert system with dynamics and control objective function both known (different trajectories have different time horizons $T$ ranging from $40$ to $50$ and different initial state conditions). Here we minimize imitation loss $L(\boldsymbol{\xi}_{\boldsymbol{\theta}},\boldsymbol{\theta}){=} {\norm{\boldsymbol{{\xi}}^{\text{d}}-\boldsymbol{{\xi}}_{\boldsymbol{\theta}}}^2}$ over the parameter of dynamics and control objective, $\boldsymbol{\theta}$ in (\ref{thetarocket}). The learning rate is set to $\eta=10^{-4}$, and we run five trials with random initial parameter guess $\boldsymbol{\theta}_0$. The imitation loss $L(\boldsymbol{\xi}_{\boldsymbol{\theta}},\boldsymbol{\theta})$ versus iteration is plotted in  Fig. \ref{appendix-ioc-rocket.1}. To validate the learned models (the learned dynamics and the learned objective function), we use the learned models to perform motion planing of rocket powered landing in unseen settings (here we use new initial condition and new time horizon). The planing results are plotted in Fig. \ref{appendix-ioc-rocket.2}, where we also plot the ground truth  for comparison.

\begin{figure}
	\centering
	\begin{subfigure}{.35\textwidth}
		\centering
		\includegraphics[width=\linewidth]{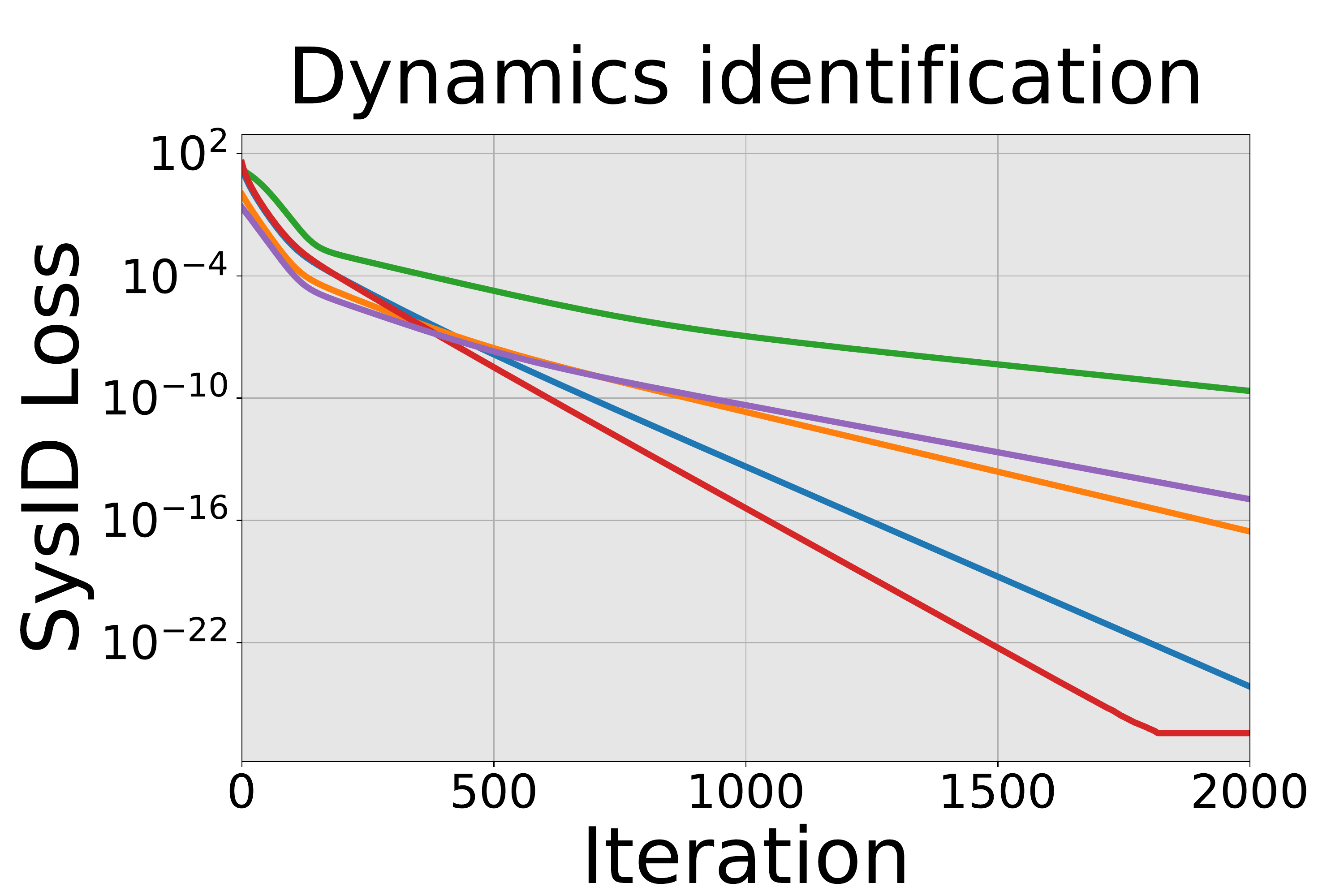}
		\caption{Training}
		\label{appendix-id-rocket.1}
	\end{subfigure}%
	\begin{subfigure}{.43\textwidth}
		\centering
		\includegraphics[width=\linewidth]{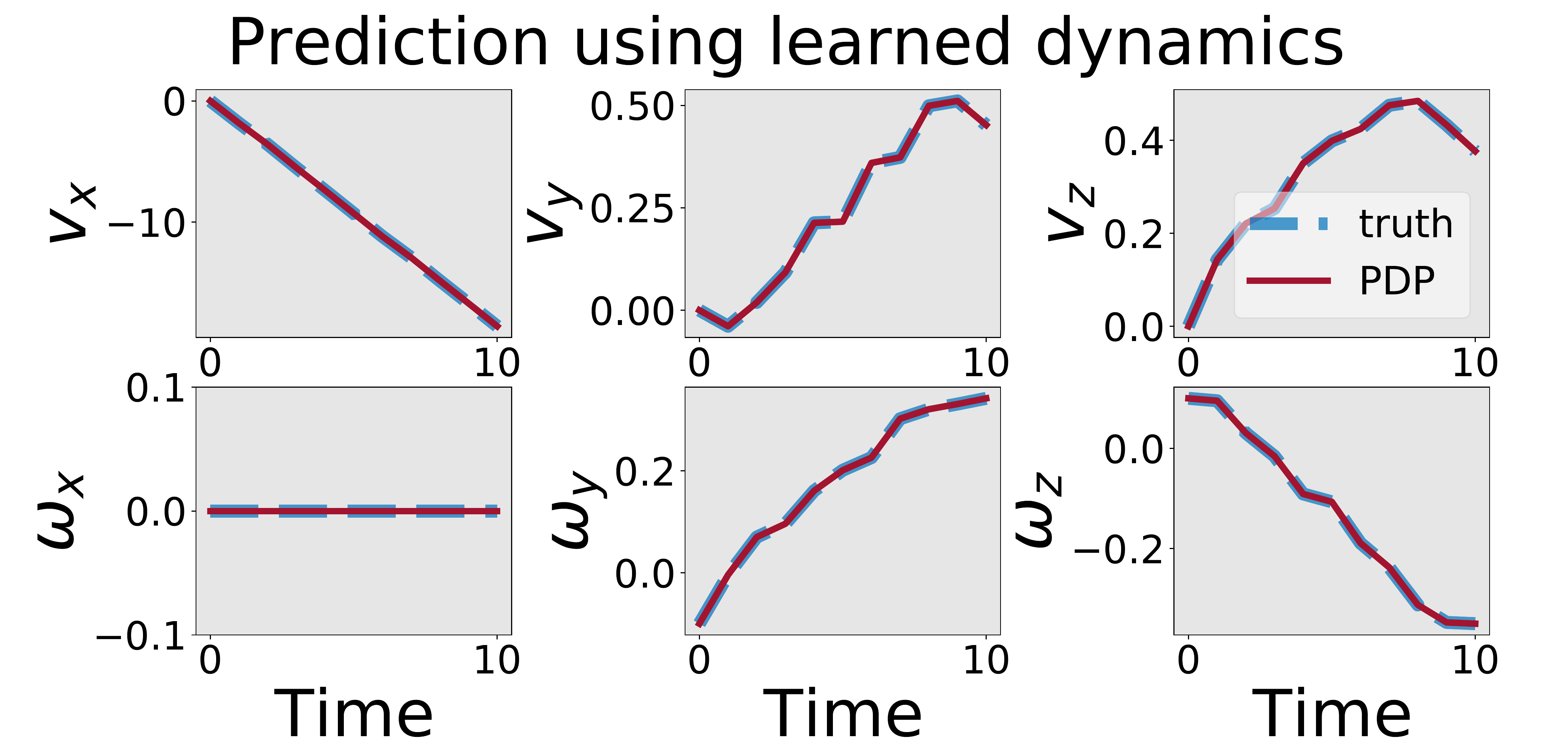}
		\caption{Validation}
		\label{appendix-id-rocket.2}
	\end{subfigure}%
	\caption{(a) Training process for identification of rocket dynamics: SysID loss versus iteration; here we have performed five trials  (labeled by different colors) with random initial parameter guess. (b) Validation: we use the learned dynamics model to perform motion prediction of the rocket given a new control sequence; here we also plot the ground-truth motion (where we know the exact dynamics). The results in (a) and (b) show that the PDP can accurately identify the  dynamics model of the rocket.}
	\label{appendix-id-rocket}\vspace{10mm}
\end{figure}

\textbf{System identification.}\quad We apply the SysID mode of  PDP to identify the dynamics parameter $\boldsymbol{\theta}_{\text{dyn}}$ of the rocket. The experiment process is similar to the experiments in Appendix \ref{appendix-exp-sysid}, where we collect five trajectories with different initial state conditions, time horizons ($T$ ranges from $10$ to $20$), and random control inputs. We minimize the SysID loss  $L(\boldsymbol{\xi}_{\boldsymbol{\theta}},\boldsymbol{\theta})= {\norm{\boldsymbol{{\xi}}^{\text{o}}-\boldsymbol{{\xi}}_{\boldsymbol{\theta}}}^2}$  over  $\boldsymbol{\theta}_{\text{dyn}}$ in (\ref{thetarocket}). The learning rate is set to $\eta=10^{-4}$, and we run five trials with random initial parameter guess for $\boldsymbol{\theta}_{\text{dyn}}$. The SysID loss $L(\boldsymbol{\xi}_{\boldsymbol{\theta}},\boldsymbol{\theta})$ versus iteration is plotted in  Fig. \ref{appendix-id-rocket.1}. To validate the learned dynamics, we use it to predict the motion of  rocket given a new sequence of control inputs. The prediction results are in  Fig. \ref{appendix-id-rocket.2}, where we also plot the ground truth  for reference.

\begin{figure}
	\centering
	\begin{subfigure}{.36\textwidth}
		\centering
		\includegraphics[width=\linewidth]{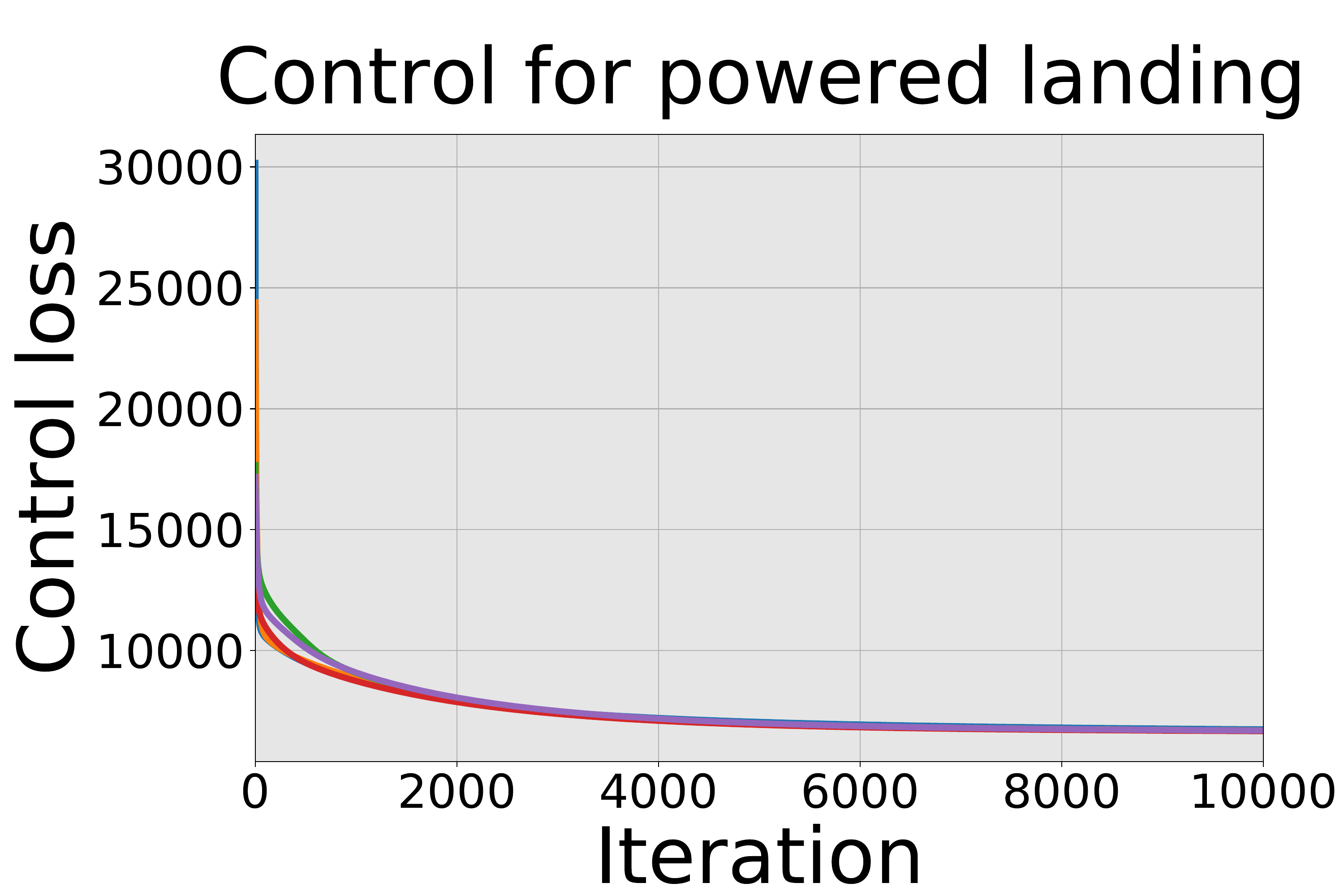}
		\caption{Training}
		\label{appendix-oc-rocket.1}
	\end{subfigure}%
	\begin{subfigure}{.43\textwidth}
		\centering
		\includegraphics[width=\linewidth]{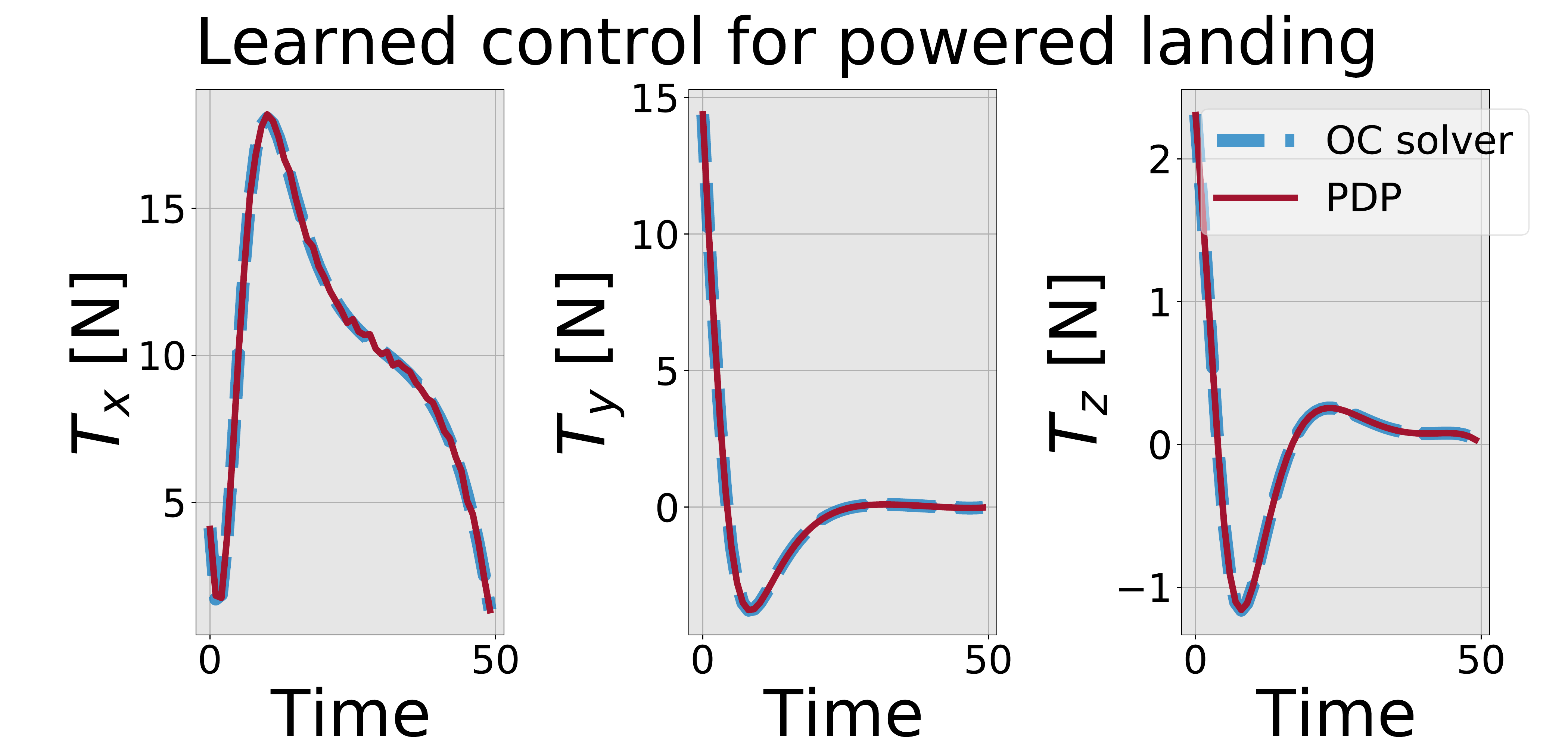}
		\caption{Validation}($T_x$ is defined along the rocket)
		\label{appendix-oc-rocket.2}
	\end{subfigure}%
	\caption{(a) Training process of learning the optimal control policy for rocket powered landing: the control loss versus iteration; here we have performed five trials  (labeled by different colors) with random initial guess of the policy parameter. (b) Validation: we use the learned policy to simulate the rocket control trajectory; here we also plot the ground-truth optimal control solved by an OC solver. The results in (a) and (b) show that the PDP can successfully find the optimal control policy (or optimal control sequence) to successfully perform the rocket powered landing. Please find the video demo  at  \url{https://youtu.be/5Jsu772Sqcg}.}
	\label{appendix-oc-rocket} 
\end{figure}

\textbf{Optimal powered landing control.}\quad We apply the Control/Planning mode of PDP to find an optimal control sequence for the  rocket to perform a successful powered landing. The experiment process is similar to the experiments performed for the quadrotor system in Appendix \ref{appendix_exp_pdp_oc}. We set the time horizon as $T=50$, and randomly choose an initial state condition $\boldsymbol{x}_0$ for the rocket. We minimize the control loss function, which is now  a given control objective function with $\boldsymbol{\theta}_{\text{obj}}$ known. The control policy we use here is  parameterized as the Lagrangian polynomial, as described in (\ref{polypolicy1}) in Appendix \ref{appendix_exp_pdp_oc}, here with degree $N=25$. The control loss is set as the control objective function learned in the previous imitation learning experiment. 
The learning rate is set to $\eta=10^{-4}$, and we run five trials with random initial guess of the policy parameter. The the control loss $L(\boldsymbol{\xi}_{\boldsymbol{\theta}},\boldsymbol{\theta})$ versus iteration is plotted in  Fig. \ref{appendix-oc-rocket.1}. To validate the learned optimal control policy, we use it to simulate the motion (control trajectory) of the rocket landing, and compare with the ground truth optimal trajectory obtained by an OC solver. The validation results are in  Fig. \ref{appendix-oc-rocket.2}.

\end{document}